\documentclass{article}

\usepackage{microtype}
\usepackage{graphicx}
\usepackage{subcaption}
\usepackage{booktabs} 
\usepackage{array}

\usepackage{hyperref}
\usepackage{url}

\usepackage[preprint]{icml2026}

\usepackage{amsmath}
\usepackage{amssymb}
\usepackage{mathtools}
\usepackage{amsthm}
\usepackage{diagbox}
\usepackage{multirow}
\usepackage{makecell}

\usepackage[capitalize,noabbrev]{cleveref}

\theoremstyle{plain}

\theoremstyle{definition}

\theoremstyle{remark}

\usepackage[textsize=tiny]{todonotes}

\icmltitlerunning{LFNO: Bridging Laplace and Fourier via Transient-Steady Decomposition}

\begin{document}

\twocolumn[
  \icmltitle{LFNO: Bridging Laplace and Fourier via Transient-Steady Decomposition}

  \begin{icmlauthorlist}
    \icmlauthor{Jeongun Ha}{yyy}
    \icmlauthor{Sanga Yoon}{yyy}
    \icmlauthor{Donghun Lee$\dagger$}{yyy}
  
  \end{icmlauthorlist}
  
  \icmlaffiliation{yyy}{Department of Mathematics, Korea University, 145 Anam-ro Seongbuk-gu Seoul, Republic of Korea}

  \icmlcorrespondingauthor{Donghun Lee}{holy@korea.ac.kr}

  \icmlkeywords{Machine Learning, ICML}

  \vskip 0.3in
]



\printAffiliationsAndNotice{}  

\begin{abstract}\label{sec: abstract} 

We introduce the Laplace–Fourier Neural Operator (LFNO), a unified framework for modeling dynamical systems across transient and steady-state regimes by integrating the spectral advantages of Laplace and Fourier Neural Operators. LFNO employs a dual-branch architecture that explicitly decomposes system dynamics into transient and steady-state components. We evaluate LFNO on nine benchmarks, including three ODE systems (Duffing, Lorenz, and Pendulum) and six PDE systems (Euler–Bernoulli beam, Heat, Reaction–diffusion, Brusselator, Burgers, and Navier–Stokes). LFNO significantly outperforms existing operators on ODE systems, where transient dynamics dominate, and consistently surpasses LNO while achieving performance competitive with FNO on PDE benchmarks. Furthermore, LFNO offers improved stability and physical interpretability through its component-wise decomposition. These results demonstrate that LFNO provides a robust and unified approach for learning complex dynamical systems across multiple temporal scales.

\end{abstract}

\section{Introduction}\label{sec: introduction}

In recent years, advances through machine learning have emerged as a powerful paradigm for solving differential equations in several applications, such as fluid dynamics \citep{lomax2001fundamentals, brunton2016discovering, ling2016reynolds, tartakovsky2020physics, ranade2021discretizationnet}, electromagnetism \citep{xiong2019data,baldan2021solving,zhang2021maxwell}, and quantum mechanics \citep{brockherde2017bypassing, weinan2018deep, hermann2020deep}. However, many of these approaches still struggle to handle problems with diverse boundary conditions or high-dimensional dynamics efficiently. 

Neural operator frameworks were proposed to overcome these obstacles by directly learning the mapping between input functions and solutions. \citep{ lu2021learning} represented input functions through a branch network and query locations through a trunk network, enabling the approximation of highly nonlinear operators across diverse domains. Based on this concept, neural networks for solving partial differential equations (PDEs), such as those in \citep{li2020neural, li2020multipole}, emerged to generalize resolution-invariant methods.
Fourier Neural Operator (FNO) \citep{li2021fourier} was later introduced as a spectral approach to operator learning. FNO uses Fourier transforms to parameterize integral kernels in the frequency domain, retaining and updating only low-frequency modes for resolution invariance, efficient training, and strong performance on complex PDE benchmarks. \citep{shin2022pseudo} theoretically and experimentally demonstrated that extending the FNO through the refinement of Fourier-based kernels can enhance expressiveness in high-frequency or inhomogeneous scenarios. However, its reliance on periodic boundary conditions and regular grids limits its applicability to general geometries. While some studies have attempted to address these limitations by incorporating additional techniques into existing models \citep{li2024physics,diab2025temporal, eshaghi2025variational}, other works have aimed to solve the problem by expanding the domain.

\citep{fanaskov2023spectral} enhances representation by combining Fourier series with Chebyshev bases, improving the approximation beyond periodic settings. \citep{bonev2023spherical} extends the FNO approach to respect the geometry of the two-dimensional sphere, which applies to forecast atmospheric dynamics, and \citep{li2023fourier, tran2023factorized} generalizes neural operators from uniform rectangular meshes to arbitrary geometries, expanding their applicability to complex spatial domains. More recently, Laplace Neural Operator (LNO) \citep{cao2024laplace} was proposed to extend operator learning to the Laplace domain. By exploiting a pole–residue formulation, LNO naturally captures both transient and steady-state behaviors of dynamical systems, offering improved accuracy for problems with non-periodic or exponentially decaying responses. While LNO captures the representation of steady-state signals, FNO provides a more robust mechanism for learning their complex, non-linear modal interactions in the frequency domain.

Inspired by the spectral decomposition capabilities of LNO and FNO, we propose the Laplace–Fourier Neural Operator (LFNO), a unified framework that synergistically combines their respective strengths. LFNO leverages the Laplace-domain representation to isolate transient dynamics while employing Fourier integral operators to enhance the expressivity of steady-state components in the frequency domain. Our contributions are as follows:

\begin{itemize} 
    \item We develop LFNO, a novel operator that unifies the theoretical foundations of Laplace-domain pole-residue modeling and Fourier spectral analysis. 
    \item We introduce an architecture-level integration that treats transient and steady-state components via dual-branch processing, ensuring high-fidelity approximation of complex temporal evolutions. 
    \item Through comprehensive benchmarks, we demonstrate that LFNO significantly outperforms LNO and FNO on ODEs with transient behavior, while maintaining performance competitive with FNO on complex PDE systems. 
\end{itemize}

\section{Methodology}\label{sec: methodology}

This section details how the complementary strengths of LNO and FNO are combined to develop our proposed method, the Laplace-Fourier Neural Operator (LFNO).

\paragraph{Pole-residue formulation}\label{para: pole_residue_formulation}

Building upon the theoretical foundations of LNO  \citep{cao2023laplace, cao2024laplace}, we model the dynamical system by substituting the kernel integral operator with a representation defined in the Laplace domain. Specifically, we parameterize the kernel $K_{\phi}(s)$ in a pole-residue form as $K_{\phi}(s) = \sum_{n=1}^{N} \frac{\beta_n}{s - \mu_n}$, where $\beta_n$ and $\mu_n$ denote trainable complex poles and residues, respectively. According to the residue theorem \citep{cao2023laplace}, the transformed output $U(s) = K_\phi(s)V(s)$ admits the following pole-residue decomposition:

\begin{equation}\label{eq: pole-residue form}
U(s) = \sum_{n=1}^{N} \frac{\gamma_n}{s - \mu_n} + \sum_{\ell=-\infty}^{\infty} \frac{\lambda_\ell}{s - i\omega_\ell},
\end{equation}

where the response residues $\gamma_n$ and $\lambda_\ell$ are given by:

\begin{equation}\label{eq: residue-theorem}
    \gamma_n = \beta_n \sum_{\ell=-\infty}^{\infty} \frac{\alpha_\ell}{\mu_n - i\omega_\ell}, \quad \lambda_\ell = \alpha_\ell \sum_{n=1}^{N} \frac{\beta_n}{i\omega_\ell - \mu_n},
\end{equation}

where $\alpha_\ell$ denotes the complex Fourier coefficients of the input. This formulation allows for a rigorous separation of the response into two distinct components: the transient response, associated with the system poles $\mu_n$, and the steady-state response, associated with the excitation poles $i\omega_\ell$ of the input signal. By evaluating the residues $\gamma_n$ at the system poles $\mu_n$ and $\lambda_\ell$ at the excitation poles $i\omega_\ell$, the time-domain solution $u(t)$ is reconstructed via the inverse Laplace transform as a superposition of decaying and oscillatory exponentials.

While this dual-response characterization extends spectral analysis beyond conventional Fourier-based methods \citep{li2021fourier, fanaskov2023spectral}, enabling the capture of non-stationary decaying dynamics, the discrete treatment of steady-state components in original LNOs may limit expressivity for complex nonlinear dependencies. To address this, we leverage the FNO framework \citep{li2021fourier} to generalize the steady-state response. By applying a learned operator $R$ in the frequency domain, our LFNO effectively captures spectral coupling and nonlinearities that simple linear pole-residue methods might overlook. This hybrid approach ensures that the transient branch resolves decaying dynamics on the $s$-plane, while the steady-state branch achieves high-fidelity convergence in the frequency domain.

\paragraph{Mathematical formulation}\label{para: math_formulation}

Following \citep{cao2024laplace}, we adopt the simple pole assumption for the Laplace domain representation, where the inverse Laplace transform of the learned representation yields the solution:
\begin{equation}\label{eq: inverse_LT_main}
    u(t) = \sum_{n=1}^N \gamma_n e^{\mu_n t} + \sum_{\ell=-L}^{L} \lambda_\ell e^{i\omega_\ell t},
\end{equation}

where the first term corresponds to the transient response $u_{transient}(t)$ and the second to the steady-state response $u_{steady}(t)$. Since $u_{steady}(t)$ is a superposition of discrete frequency modes, we apply the Fourier transform to represent it in the frequency domain:
\begin{equation}\label{eq: steady_state_dirac_main}
    \mathcal{F}(u_{steady})(\omega) = \sum_{\ell=-L}^{L} \lambda_\ell \delta(\omega_\ell - \omega).
\end{equation}

We now apply the Fourier integral operator \citep{li2021fourier}. In the discrete frequency domain, FNO parameterizes the operator $R$ as a complex-valued tensor $R \in \mathbb{C}^{d \times d \times (2L+1)}$, where $d$ is the number of hidden channels. For each frequency mode $\ell \in \{-L, \dots, L\}$, the updated steady-state coefficients $\hat{\lambda}_\ell$ are computed as:

\begin{equation}\label{eq: steady_state_FNO_revised}
    \hat{\lambda}_\ell = R_\ell \cdot \lambda_\ell, \quad \text{for } -L \leq \ell \leq L
\end{equation}

where $R_\ell \in \mathbb{C}^{d \times d}$ is the learnable weight matrix for the $\ell$-th mode. This transformation allows the model to learn a neural-approximated transfer function that enhances the expressivity of the steady-state component. Applying the inverse transform, the enhanced steady-state response $\tilde{u}_{steady}$ is obtained by

\begin{equation}\label{eq: inverse_FT_steady_revised}
    \tilde{u}_{steady}(t) = \sum_{\ell=-L}^{L} \hat{\lambda}_\ell e^{i\omega_\ell t} = \sum_{\ell=-L}^{L} (R_\ell \lambda_\ell) e^{i \omega_\ell t}.
\end{equation}

The final solution is formulated as the superposition of the preserved transient dynamics and the filtered steady-state response:

\begin{equation}\label{eq: full_solution_revised}
    u(t) = \sum_{n=1}^N \gamma_n e^{\mu_n t} + \sum_{\ell=-L}^{L} (R_\ell \lambda_\ell) e^{i \omega_\ell t}.
\end{equation}

The additive combination in \cref{eq: full_solution_revised} follows the superposition principle of linear differential equations, while the non-linear interaction between components is further captured by the local linear transformations and projection layers in the subsequent architecture.

\paragraph{Architecture}

\begin{figure*}[ht]
    \centering
    \subfloat[Full Architecture of LFNO]{\includegraphics[width=0.7\textwidth]{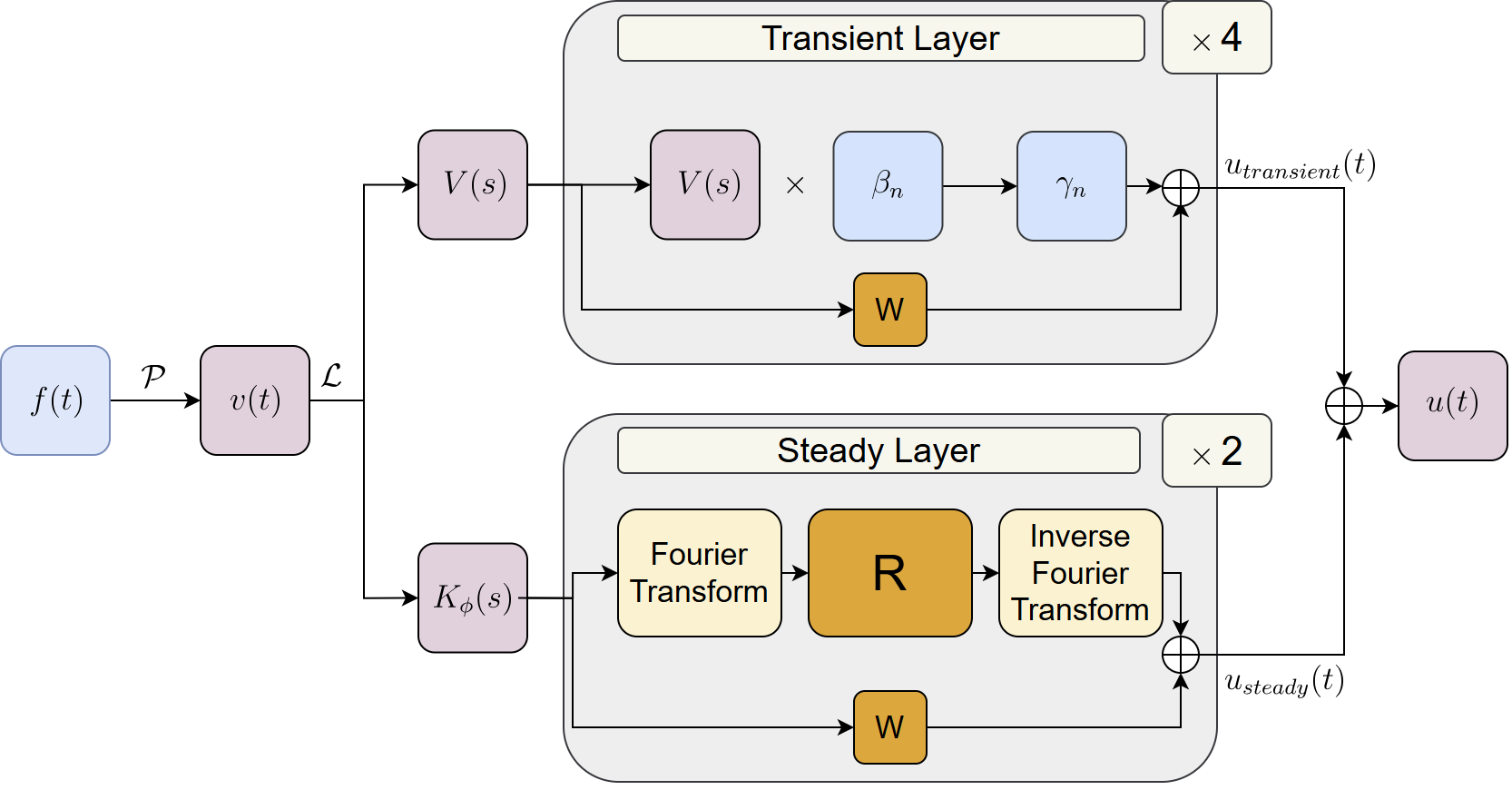}}

    \vskip 0.5em

    \subfloat[Description of Applying the Fourier integral operator in Steady Layer]{\includegraphics[width=0.7\textwidth]{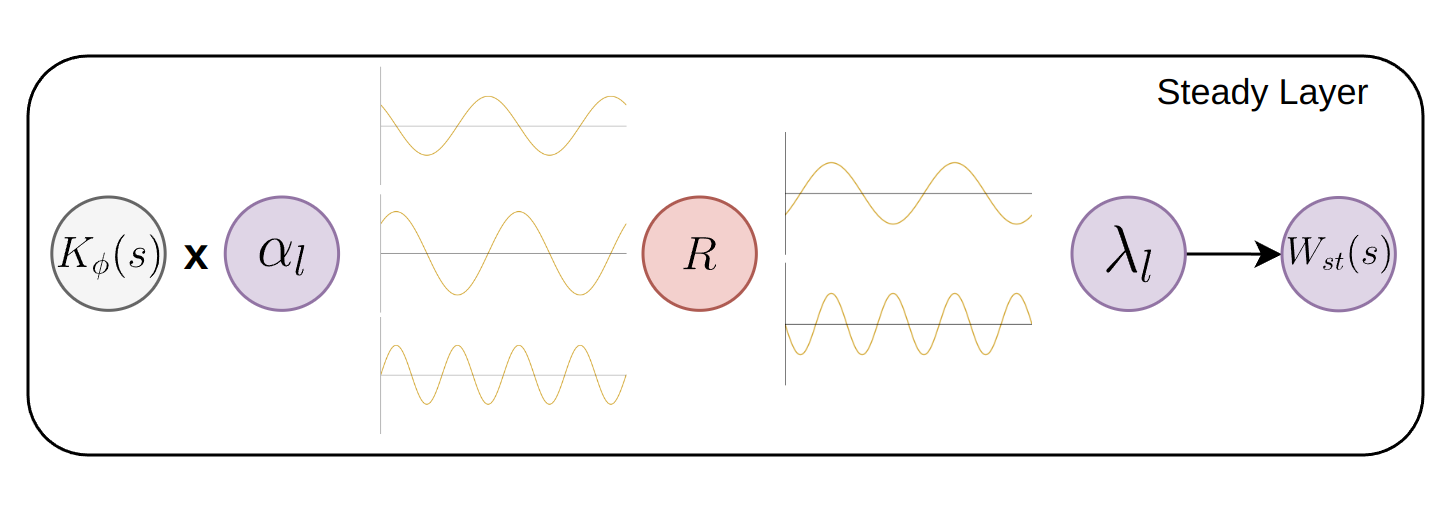}}
    \caption{Model architecture of LFNO. (a) Overall process: given input function $\mathbf{f}(t)$, 1. embed the input function $\mathbf{f}(t)$ into a higher-dimensional latent space $\mathbf{v}(t)$ via a shallow neural network $\mathcal{P}$, 2. apply Laplace transform $\mathcal{L}$, 3. take the Transient and the Steady layers for each component, and 3. Project the integrated result back to the target space to obtain the output $u(t)$. (b) Steady layer structure: Hybrid spectral processing that applies the pole-residue method followed by a learnable Fourier weight matrix $R$ on the lower Fourier modes, and filters out the higher modes.}
    \label{fig: LFNO_architecture}
\end{figure*}

The overall architecture is illustrated in \cref{fig: LFNO_architecture}. Starting from an input function $\mathbf{f}(t)$, we first lift it to a higher-dimensional representation using a neural network $\mathcal{P}$. We then apply the Laplace transform to obtain $v(t)$ and decompose it into transient and steady-state components for each channel. In the steady layers, we apply the operator $R_\ell$ to the lower Fourier modes while filtering out higher modes to achieve regularization and computational efficiency \citep{li2021fourier}. The transient layers operate in the time-domain, modeling decaying exponential dynamics. In this study, we allocate more layers to the transient module than to the steady-state module. This design choice is motivated by the empirical observation that transient phases—characterized by rapid transitions and complex spatiotemporal gradients—typically require a higher model capacity to achieve convergence compared to the periodic and structured nature of the steady-state response. Finally, a local linear transformation $W$ integrated with non-linear activation functions is applied, and the result is projected back to the target dimension using another neural network $\mathcal{Q}$.

\section{Dataset Generation}\label{sec: dataset_generation}

To rigorously evaluate the generalization and data efficiency of LFNO, we synthesize custom datasets across a wide range of physical systems. We adopt three ODEs (Duffing, Lorenz, Pendulum) and six PDEs (Euler-Bernoulli beam, heat, reaction-diffusion, Brusselator, Burgers, and Navier-Stokes). These benchmarks are specifically chosen to represent diverse spectral characteristics, ranging from simple oscillations to highly nonlinear turbulence. The sample visualizations for the dataset are shown in \cref{sec: dataset_samples}.

\subsection{ODEs}\label{subsec: 1d_odes}

We generate 380 samples for each ODE dataset by varying the parameter $A$ in the source term $f(t)=Ae^{-0.05t}\sin(5t)$, as in \citep{cao2024laplace}, where $A$ is sampled from 380 evenly spaced values in the interval $[0.05, 10]$. While the number of independent trajectories is relatively compact ($N=380$), each sample contains a high-density sequence of 2048 time steps with a resolution of $\Delta t = 0.01s$. This configuration is intentionally designed to evaluate the model's sample efficiency in data-scarce scenarios, challenging the framework to resolve complex temporal evolutions from a limited pool of physical realizations. Such a setup reflects real-world scenarios where high-fidelity data acquisition is computationally or experimentally expensive. We employ the \texttt{solve\_ivp} solver with the \texttt{RK45} algorithm (substep of 4) to ensure numerical stability. For all equations and coefficients, details are given in \cref{subsec: 1d_odes_configuration}.

\subsection{PDEs}\label{subsec: 2d_pdes}

The PDE datasets are synthesized using two distinct strategies to evaluate the framework's versatility: (1) applying task-specific forcing functions to simulate externally driven dynamics, and (2) varying initial conditions for unforced, nonlinear systems, such as the Burgers' and Navier-Stokes equations. While the spatio-temporal grid resolution for most PDEs is maintained at $50 \times 50$, the Navier-Stokes equation was simulated on a $32 \times 32$ spatial grid. This adjustment was necessary to mitigate Out-of-Memory (OOM) constraints associated with the LNO architecture during high-resolution 2D processing. We utilize the \texttt{RK45} algorithm, except for the reaction-diffusion (R-D) equation, which requires the \texttt{BDF} method for stiff dynamics. By maintaining a restricted sample size of 380 across all nine benchmarks, we demonstrate that LFNO possesses superior sample efficiency. The details of the equations and coefficients are in \cref{subsec: 2d_pdes_configuration}. All PDE samples are described in \cref{tab: pde_configurations}. 

\section{Results} \label{sec: results}

We evaluate the empirical performance of LFNO by comparing it against LNO and FNO across diverse problem settings. Our analysis focuses on these three closely related frameworks—LFNO, LNO, and FNO—to ensure a rigorous comparison. Quantitative performance is assessed using the relative $\mathcal{L}_2$ and $\mathcal{L}_{\infty}$ error on test samples, while qualitative effectiveness is evaluated by visualization of predicted trajectories. In particular, the $\mathcal{L}_{\infty}$ metric provides a rigorous assessment of robustness against worst-case local errors, such as shock-induced oscillations in Burgers' equation, which may be overlooked by average-based metrics like $\mathcal{L}_2$.

Notably, for the Navier-Stokes (NS) equations, we conduct experiments on two distinct regimes to verify the model's scalability: (i) a canonical configuration featuring structured initial conditions to assess spectral energy preservation, and (ii) a high-complexity turbulent regime using the standard benchmarks from the original FNO study to evaluate performance under low-viscosity ($\nu=10^{-3}$) and multi-scale vortex interactions. For all benchmarks, we employ a fixed hyperparameter configuration and a consistent number of training epochs. Detailed results follow in the subsections below, with implementation specifics provided in \cref{sec: implementation_details}.

\subsection{ODE Experiments}\label{subsec: 1d_odes_results}

The detailed $\mathcal{L}_2$ and $\mathcal{L}_{\infty}$ errors are presented in \cref{tab: l2_error_table_ode} and \cref{tab: l_infty_error_table_ode}. The results indicate that LFNO consistently achieves superior performance across diverse dynamical systems. Specifically, in the Duffing $c=0.5$ and Lorenz $\rho=10$ tasks, LFNO significantly reduces the $\mathcal{L}_{\infty}$ error compared to both FNO and LNO, demonstrating its robustness in handling complex nonlinearities.

It should be noted that while FNO shows competitive performance in the purely periodic Duffing $c=0$ case following rigorous hyperparameter optimization, LFNO maintains a dominant advantage in systems with both transient and steady-state characteristics, such as the Lorenz and dissipative Duffing/Pendulum oscillators. For qualitative evaluation, we overlay the inference results of the baseline models as shown in \cref{tab: 1d_comparison}. In the Lorenz experiments, LFNO captures long-term damping and chaotic trajectories with high fidelity, whereas LNO and FNO struggle to reproduce the correct amplitudes. Furthermore, LFNO effectively integrates periodic and transient feature extraction, outperforming LNO in periodic capture for Duffing $c=0.5$ and Pendulum $c=0.5$ experiments.

{
    \newcommand{\imhere}[2]{
    \includegraphics[width=\linewidth]{images/results/ODE_results/#1_LFNO_prediction_singleFigure_LFNO_1D_#2.png}
}
    \newcolumntype{C}{>{\centering\arraybackslash}m{0.4\linewidth}}
    \begin{table}[t]
    \centering
    \begin{tabular}{cC@{\hspace{-2mm}}C}
        \toprule
        Task & \makecell{LFNO vs. FNO} & \makecell{LFNO vs. LNO} \\
        \midrule
        \makecell{Duffing \\ $c = 0$} & 
        \imhere{FNO}{Duffing_c0} &
        \imhere{LNO}{Duffing_c0} \\
        \makecell{Duffing \\ $c = 0.5$} & 
        \imhere{FNO}{Duffing_c05} &
        \imhere{LNO}{Duffing_c05} \\
        \makecell{Lorenz \\ $\rho = 5$} &
        \imhere{FNO}{Lorenz_rho5} &
        \imhere{LNO}{Lorenz_rho5} \\
        \makecell{Lorenz \\ $\rho = 10$} &
        \imhere{FNO}{Lorenz_rho10} &
        \imhere{LNO}{Lorenz_rho10} \\
        \makecell{Pendulum \\ $c= 0.5$} & 
        \imhere{FNO}{Pendulum_c05} &
        \imhere{LNO}{Pendulum_c05} \\
        \bottomrule
    \end{tabular}
    \vskip 0.1in
    \caption{Qualitative results for three models. Comparison of LFNO with FNO (left) and LNO (right). Red: LFNO results; Green: comparison model results.}
    \label{tab: 1d_comparison}
    
\end{table}
}

\begin{table}[ht]
    \begin{center}

    \begin{small}
    \resizebox{\columnwidth}{!}{
    \begin{tabular}{c|ccccc}
        \toprule
        & \multicolumn{5}{c}{\textbf{ODEs}} \\
        \midrule
        Equation & \makecell{Duffing \\ $c=0$}& \makecell{Duffing \\ $c=0.5$}& \makecell{Lorenz \\ $\rho=5$}& \makecell{Lorenz \\ $\rho=10$} & \makecell{Pendulum \\ $c=0.5$} \\
        \midrule
        Epochs & 5000 & 5200 & 8100 & 4000 & 6800 \\
        \midrule
        FNO & 0.0726 & 0.0565 & 0.1225 & 0.4340 & 0.0426 \\ 
        LNO & 0.2014 & 0.0762 & 0.0288 & 0.2571 & 0.0735 \\
        LFNO & \textbf{0.0518} & \textbf{0.0142} & \textbf{0.0075} & \textbf{0.1731} & \textbf{0.0315} \\
        \bottomrule
    \end{tabular}
    }
    \end{small}
    \end{center}
    \caption{$\mathcal L_2$ error table for ODEs tasks.}
    \label{tab: l2_error_table_ode}
\end{table}

\begin{table}[ht]
    \begin{center}

    \begin{small}
    \resizebox{\columnwidth}{!}{
    \begin{tabular}{c|ccccc}
        \toprule
        & \multicolumn{5}{c}{\textbf{ODEs}} \\
        \midrule
        Equation & \makecell{Duffing \\ $c=0$}& \makecell{Duffing \\ $c=0.5$}& \makecell{Lorenz \\ $\rho=5$}& \makecell{Lorenz \\ $\rho=10$} & \makecell{Pendulum \\ $c=0.5$} \\
        \midrule
        Epochs & 5000 & 5000 & 5000 & 5000 & 5000 \\
        \midrule
        FNO & \textbf{0.1338} & 0.0293 & 0.4295 & 0.7137 & 0.0356 \\ 
        LNO & 0.6686 & 0.1600 & \textbf{0.0837} & 0.4087 & 0.1405\\
        LFNO & 0.2842 & \textbf{0.0218} & 0.2245 & \textbf{0.2646} & \textbf{0.0337} \\
        \bottomrule
    \end{tabular}
    }
    \end{small}
    \end{center}
    \caption{$\mathcal L_\infty$ error table for ODEs tasks.}
    \label{tab: l_infty_error_table_ode}
\end{table}

\begin{table*}[ht]
    \begin{center}
    \begin{small}
    \begin{tabular}{c|ccccccc}
        \toprule
        & \multicolumn{7}{c}{\textbf{PDEs}} \\
        \midrule
        Equation & \makecell{Beam}& \makecell{Heat}& \makecell{R-D}& \makecell{Brusselator} & \makecell{Burgers} & \makecell{NS} & \makecell{NS $\nu=10^{-3}$}\\
        \midrule
        Epochs & 5000 & 5000 & 5000 & 5000 & 5000 & 5000 & 5000\\
        \midrule
        FNO & \textbf{0.0037} & \textbf{0.0015} & \textbf{0.0028} & \textbf{0.0023} & 0.0046 & 0.0006 & 0.0286\\ 
        LNO & 0.2509 & 0.0227 & 0.0779 & 0.0137 & 0.0275 & 0.0042 & 0.0458\\
        LFNO & 0.0095 & 0.0031 & 0.0043 & 0.0029 & \textbf{0.0026} & \textbf{0.0004} & \textbf{0.0026} \\
        \bottomrule
    \end{tabular}
    \end{small}
    \end{center}
    \caption{$\mathcal L_2$ error table for PDEs tasks.}
    \label{tab: l2_error_table_pde}
\end{table*}

\begin{table*}[ht]
    \begin{center}
    \begin{small}
    \begin{tabular}{c|ccccccc}
        \toprule
        & \multicolumn{7}{c}{\textbf{PDEs}} \\
        \midrule
        Equation & \makecell{Beam}& \makecell{Heat}& \makecell{R-D}& \makecell{Brusselator} & \makecell{Burgers} & \makecell{NS} & \makecell{NS $\nu=10^{-3}$}\\
        \midrule
        Epochs & 5000 & 5000 & 5000 & 5000 & 5000 & 5000 & 5000 \\
        \midrule
        FNO & \textbf{0.0072} & \textbf{0.0059} & \textbf{0.0078} & \textbf{0.0083} & 0.0471 & \textbf{0.0024} & 0.0411\\ 
        LNO & 0.2798 & 0.0560 & 0.1400 & 0.0567 & 0.2387 & 0.0066 & 0.1268\\
        LFNO & 0.0245 & 0.0124 & 0.0148 & 0.0110 & \textbf{0.0125} & \textbf{0.0024} & \textbf{0.0071} \\
        \bottomrule
    \end{tabular}
    \end{small}
    \end{center}
    \caption{$\mathcal L_\infty$ error table for PDEs tasks.}
    \label{tab: l_infty_error_table_pde}
\end{table*}

\begin{figure*}[t]
    \centering
    \scriptsize
    \setlength{\tabcolsep}{2pt}

    \begin{tabular}{
        >{\centering\arraybackslash}m{0.07\textwidth}
        >{\centering\arraybackslash}m{0.18\textwidth}
        >{\centering\arraybackslash}m{0.18\textwidth}
        >{\centering\arraybackslash}m{0.18\textwidth}
        >{\centering\arraybackslash}m{0.18\textwidth}
    }

        & GT & FNO & LFNO & LNO \\

        Beam &
        \includegraphics[width=0.17\textwidth]{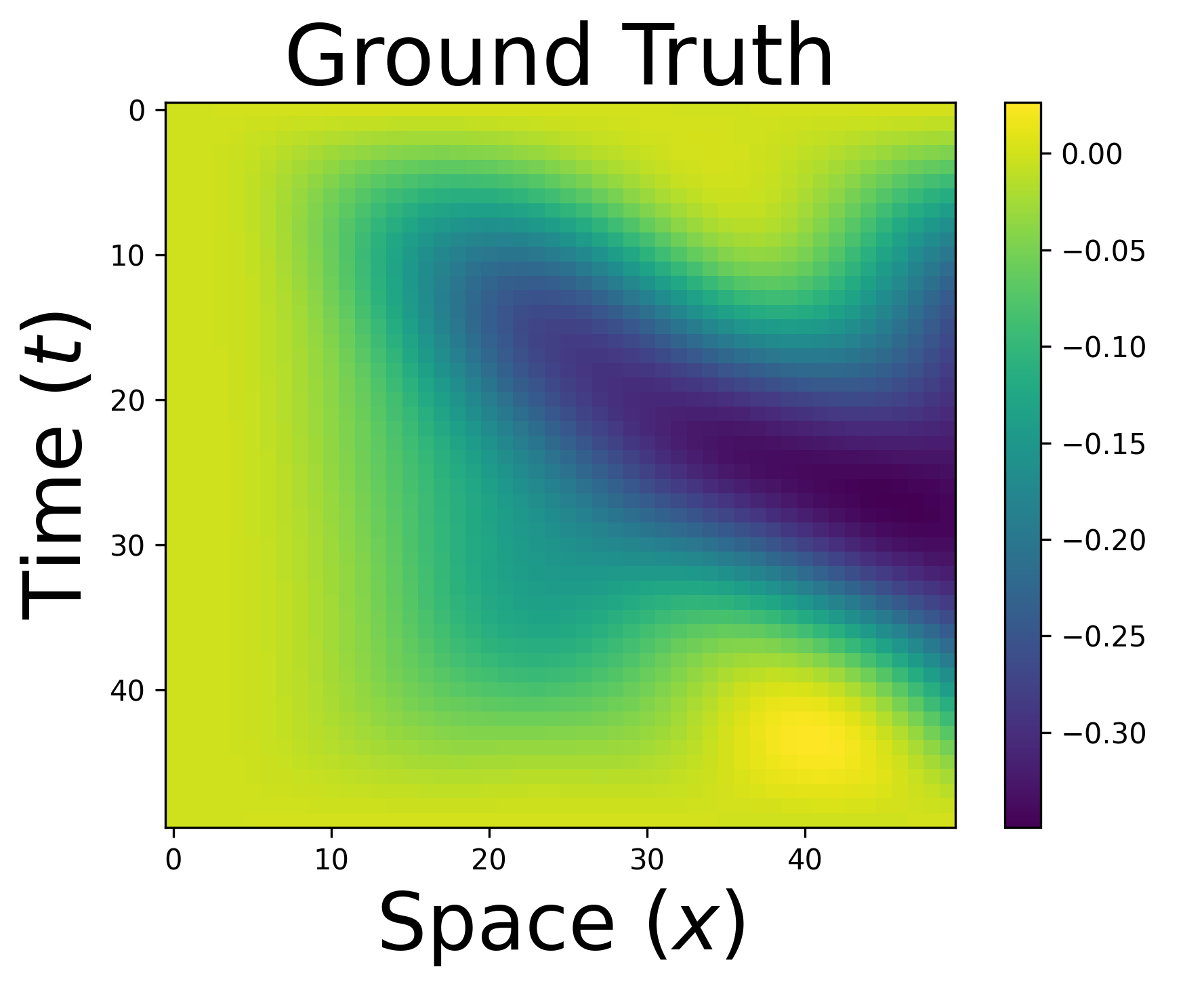} &
        \includegraphics[width=0.17\textwidth]{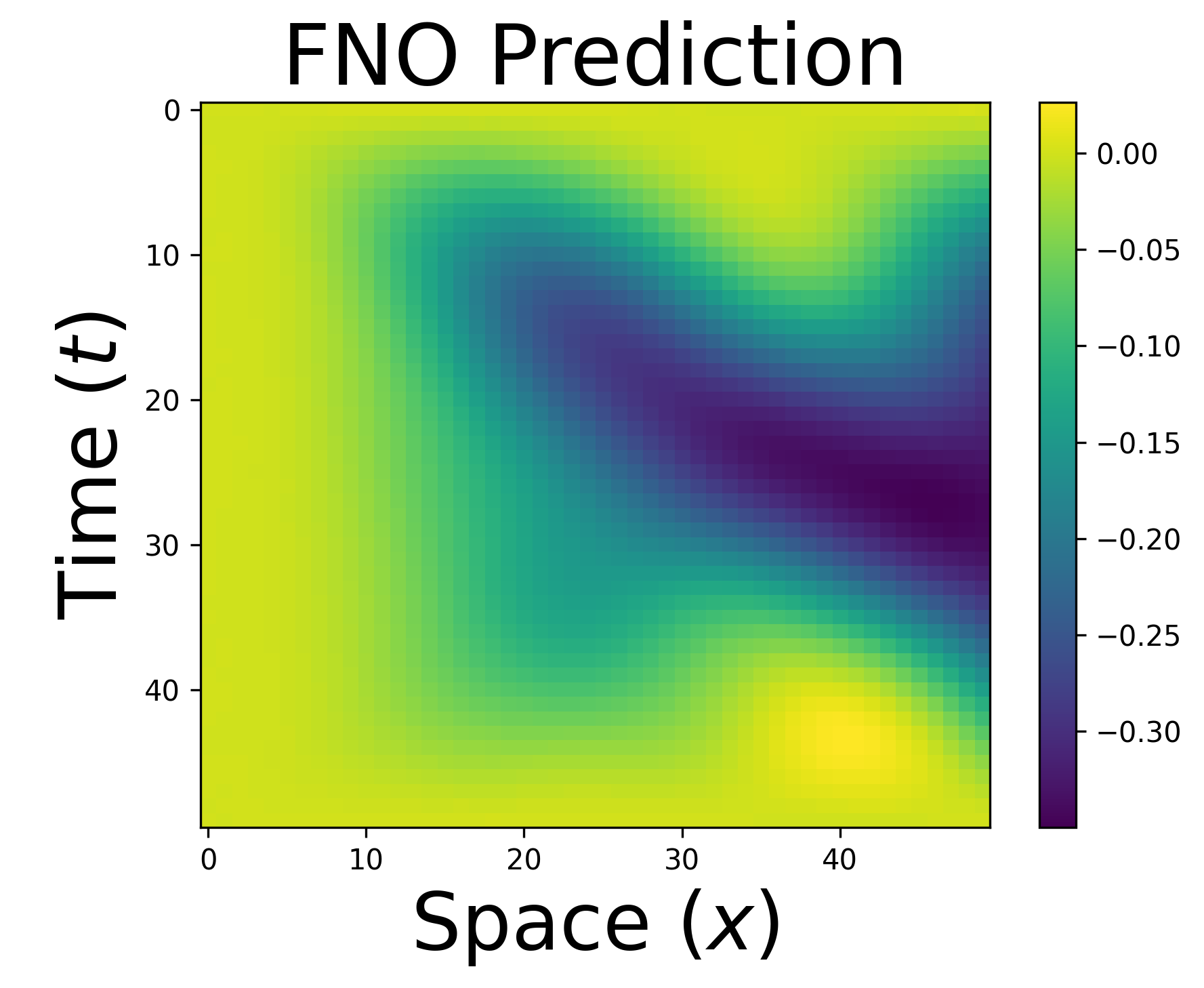} &
        \includegraphics[width=0.17\textwidth]{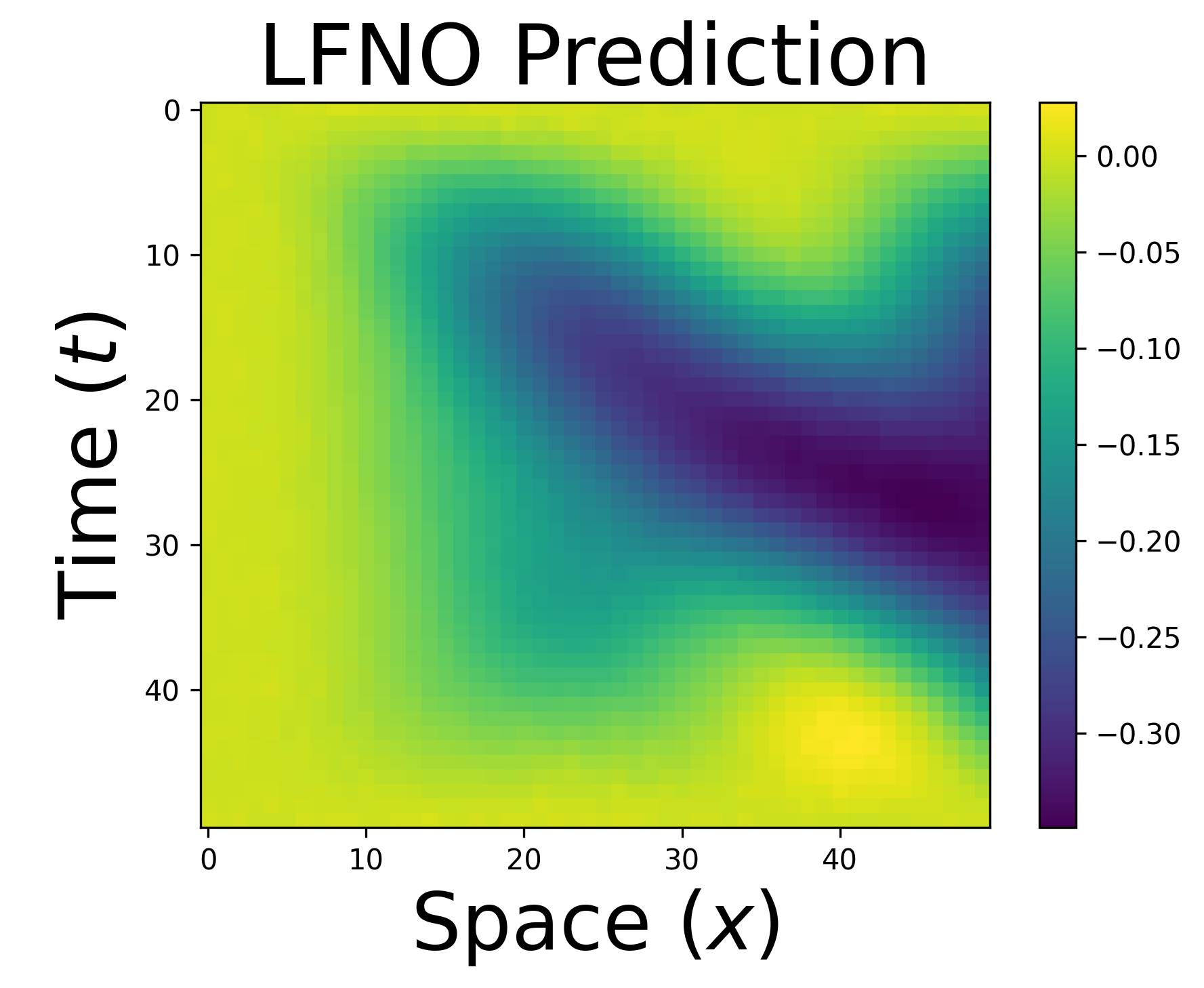} &
        \includegraphics[width=0.17\textwidth]{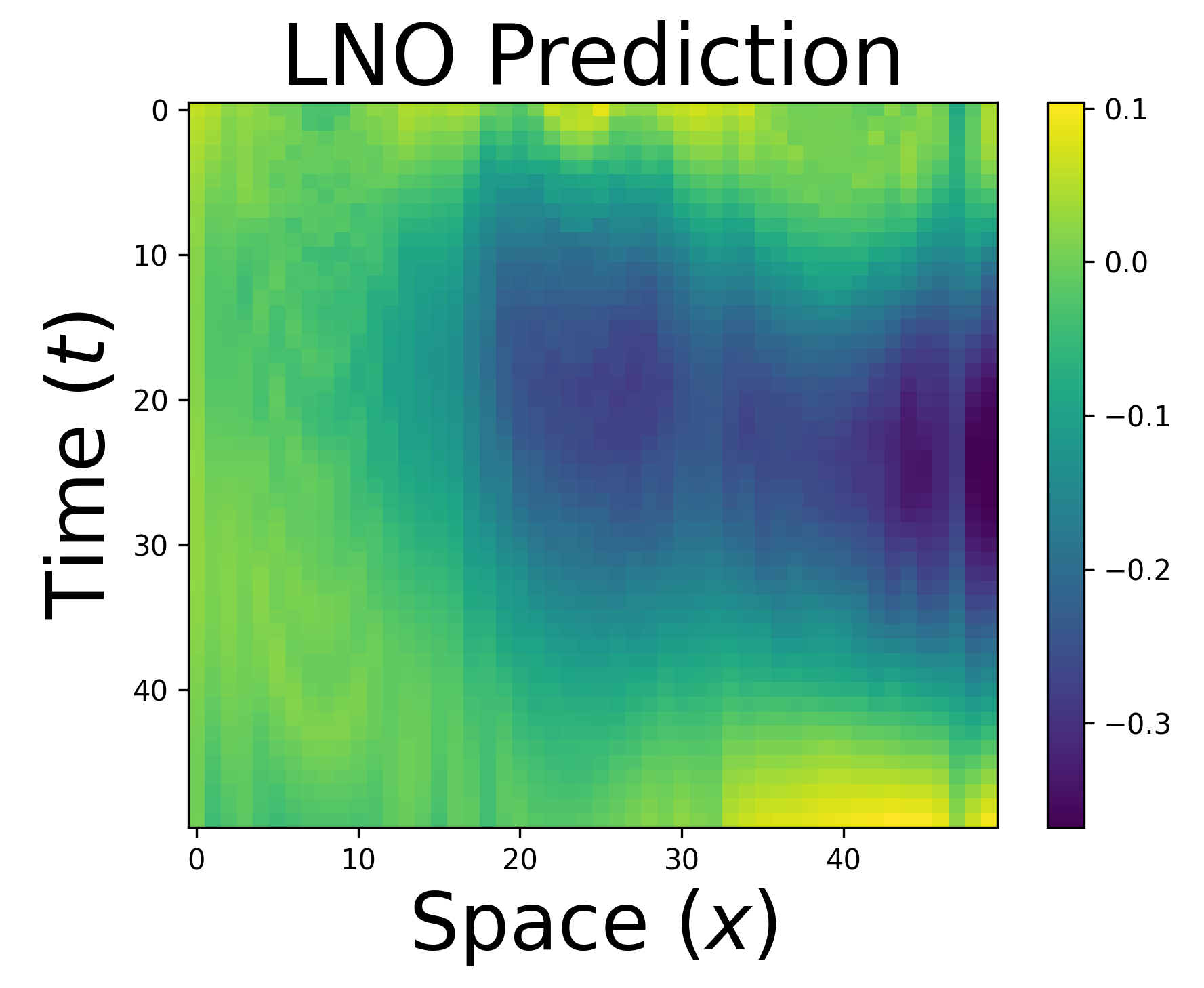} \\[-0.4em]

        Heat &
        \includegraphics[width=0.17\textwidth]{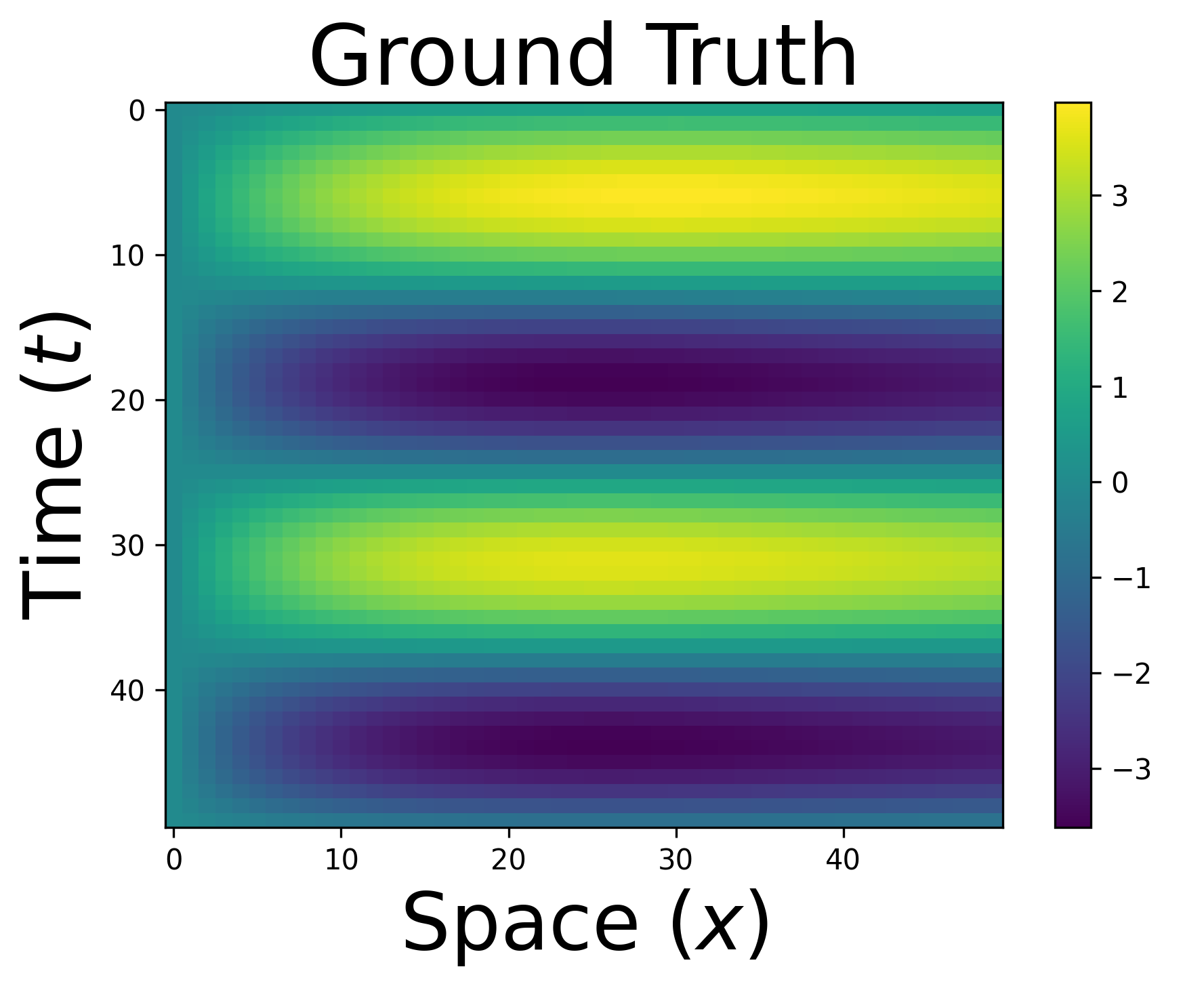} &
        \includegraphics[width=0.17\textwidth]{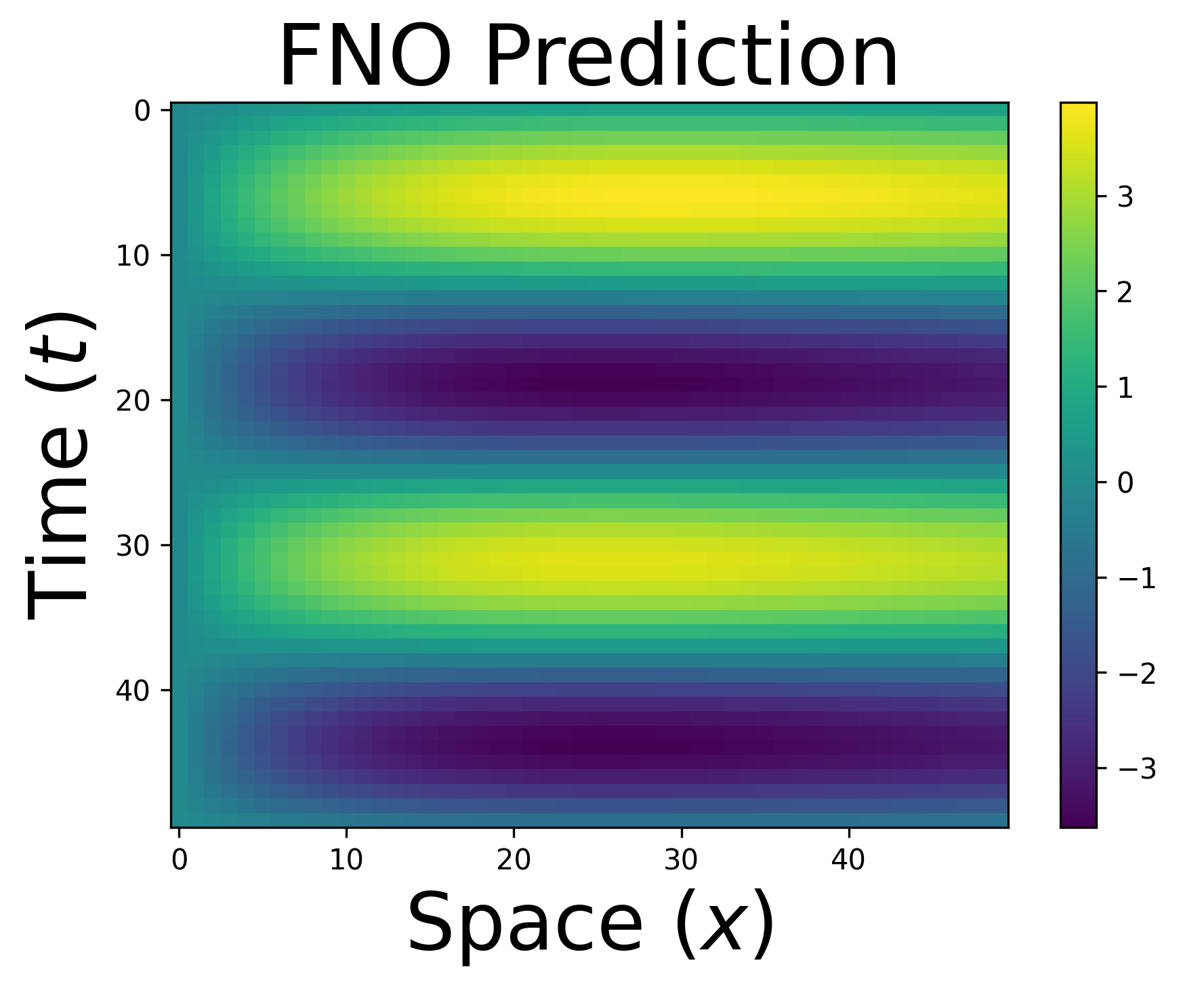} &
        \includegraphics[width=0.17\textwidth]{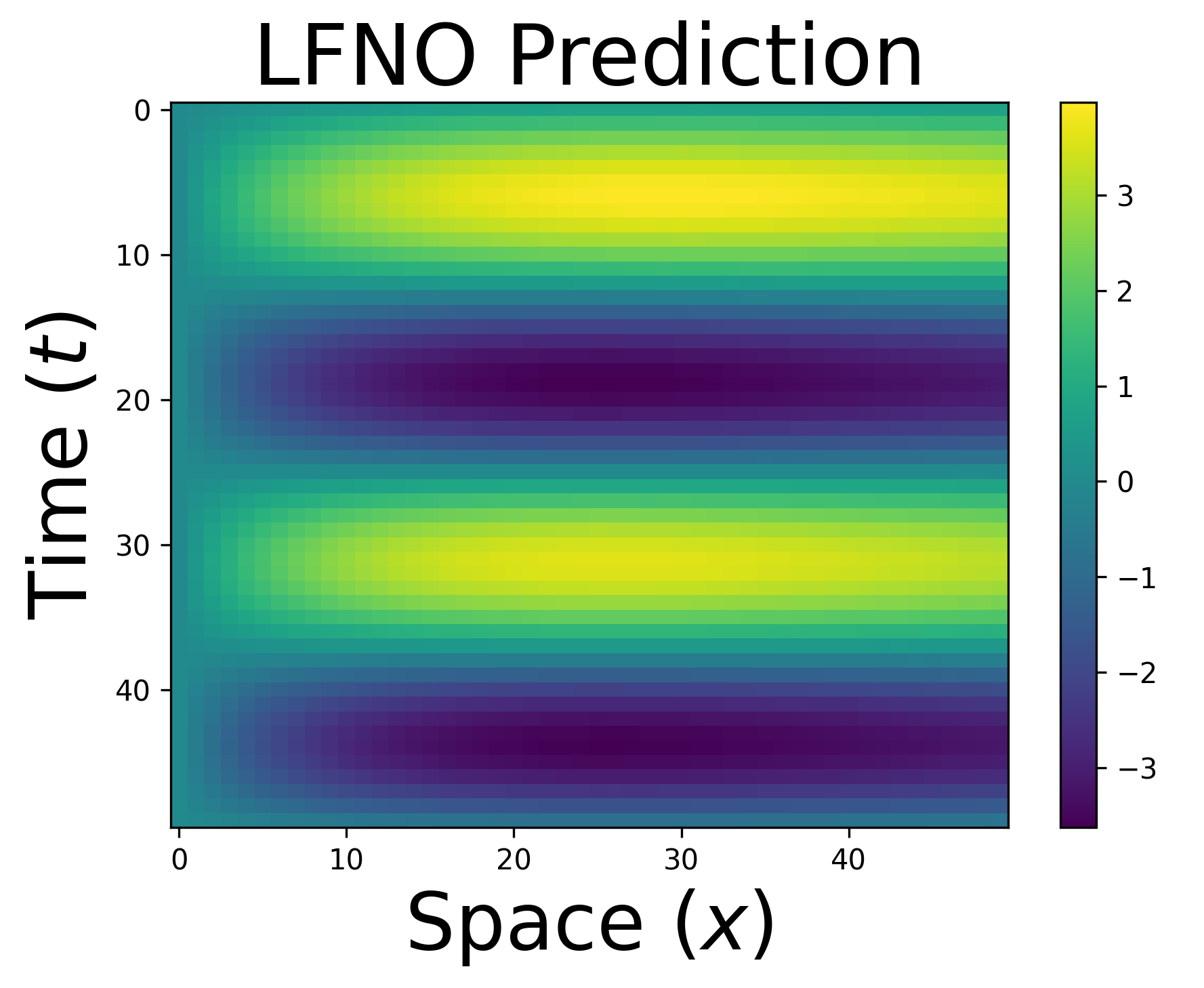} &
        \includegraphics[width=0.17\textwidth]{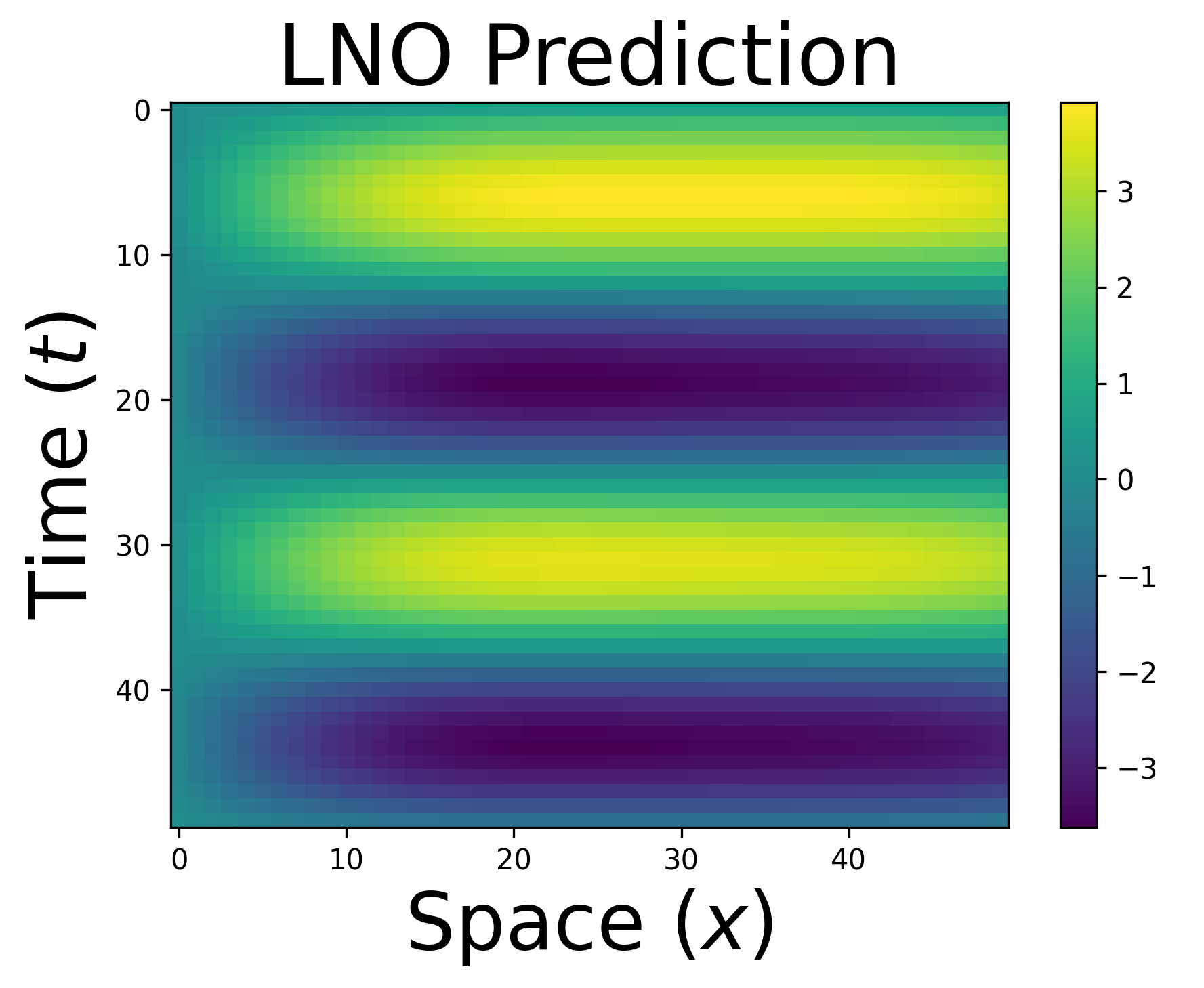} \\[-0.4em]

        R-D &
        \includegraphics[width=0.17\textwidth]{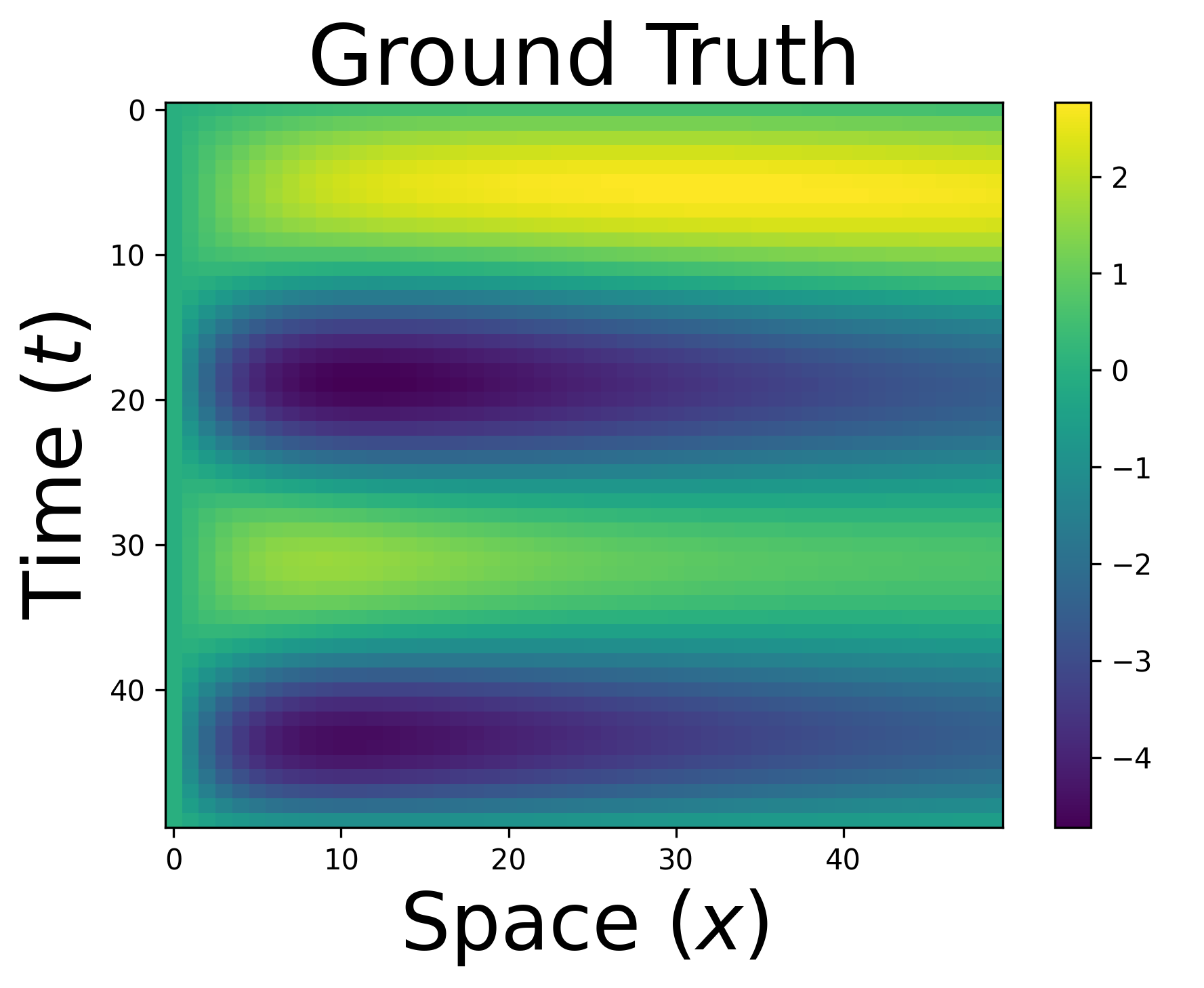} &
        \includegraphics[width=0.17\textwidth]{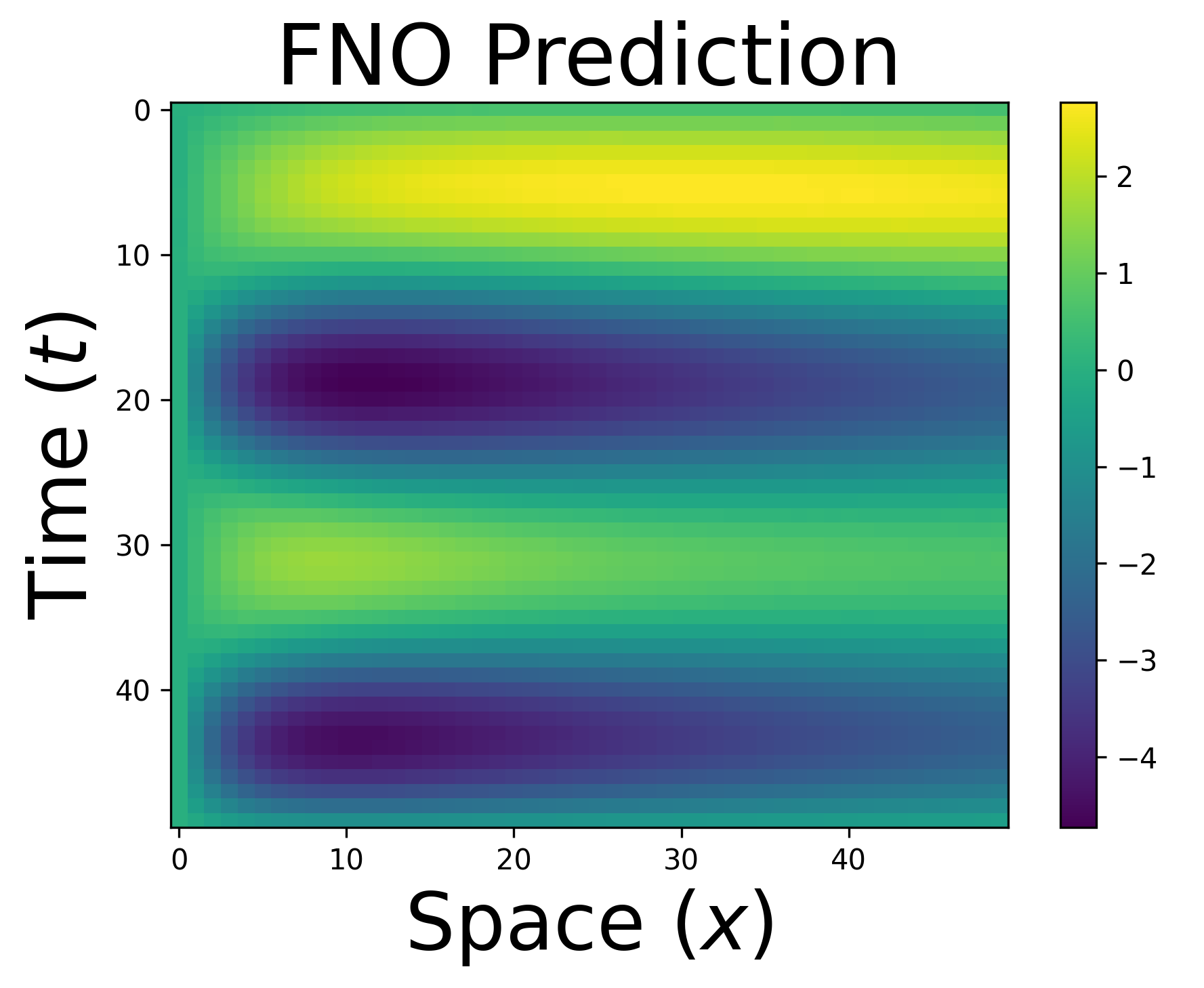} &
        \includegraphics[width=0.17\textwidth]{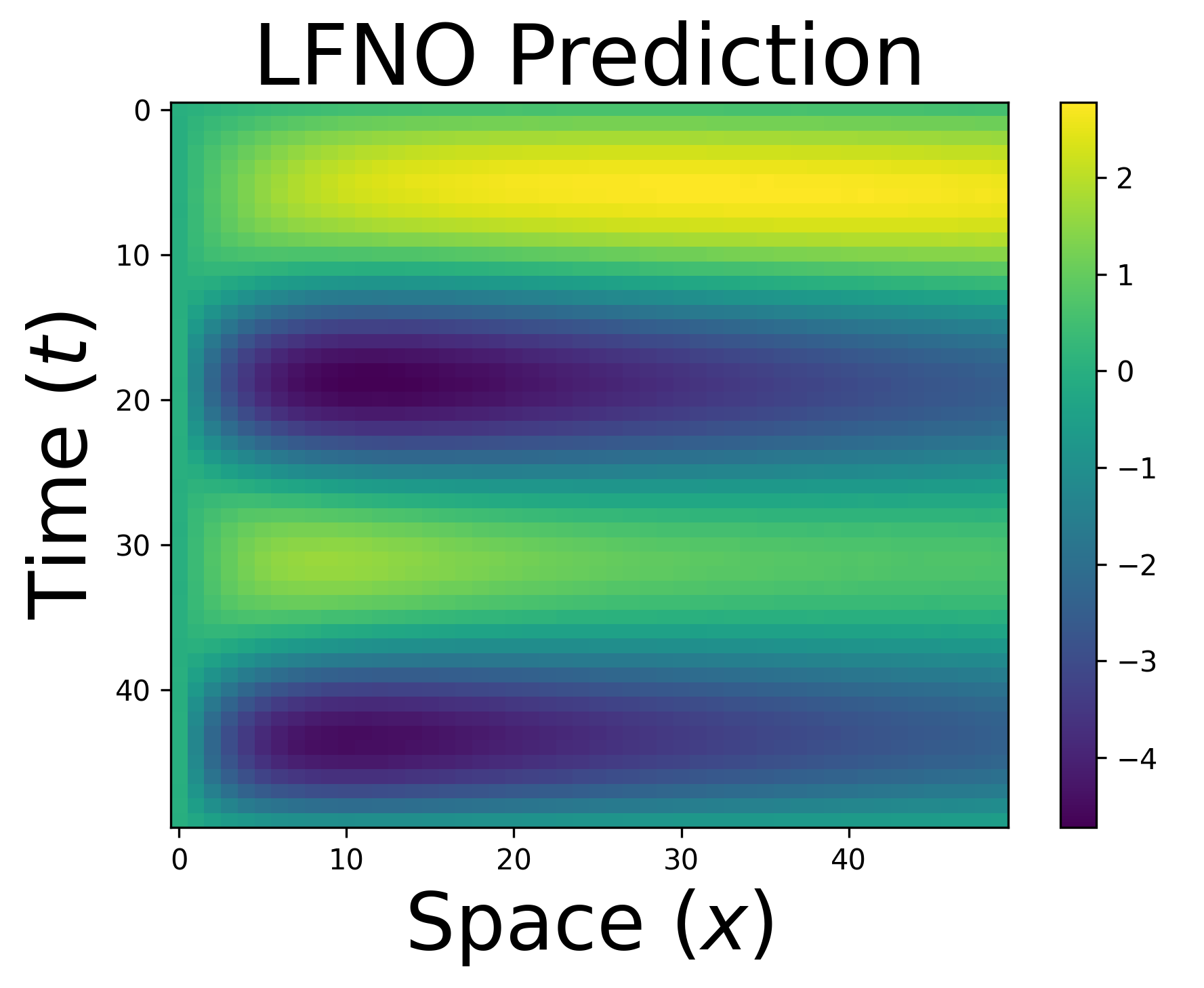} &
        \includegraphics[width=0.17\textwidth]{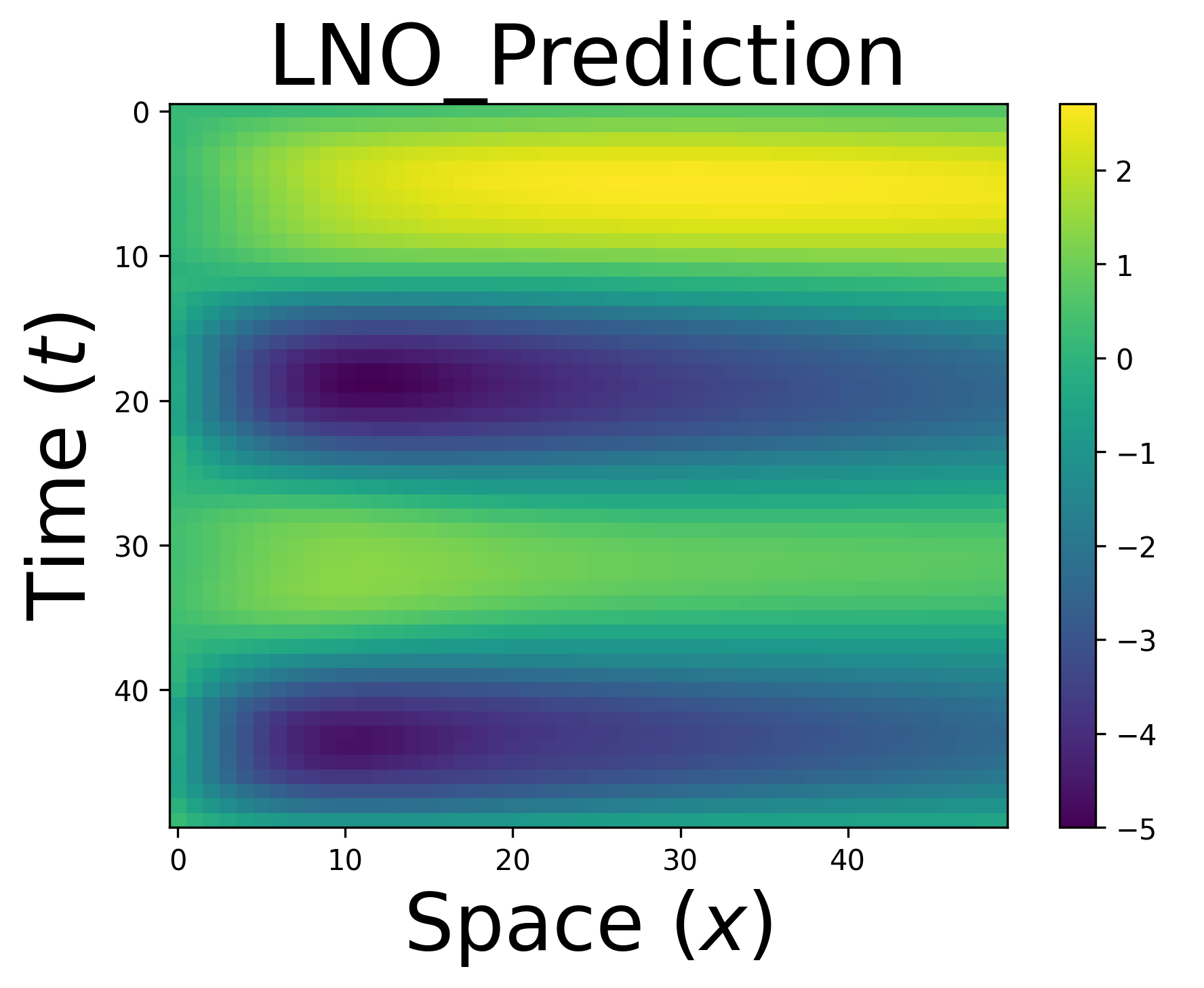} \\[-0.4em]

        Brusselator &
        \includegraphics[width=0.17\textwidth]{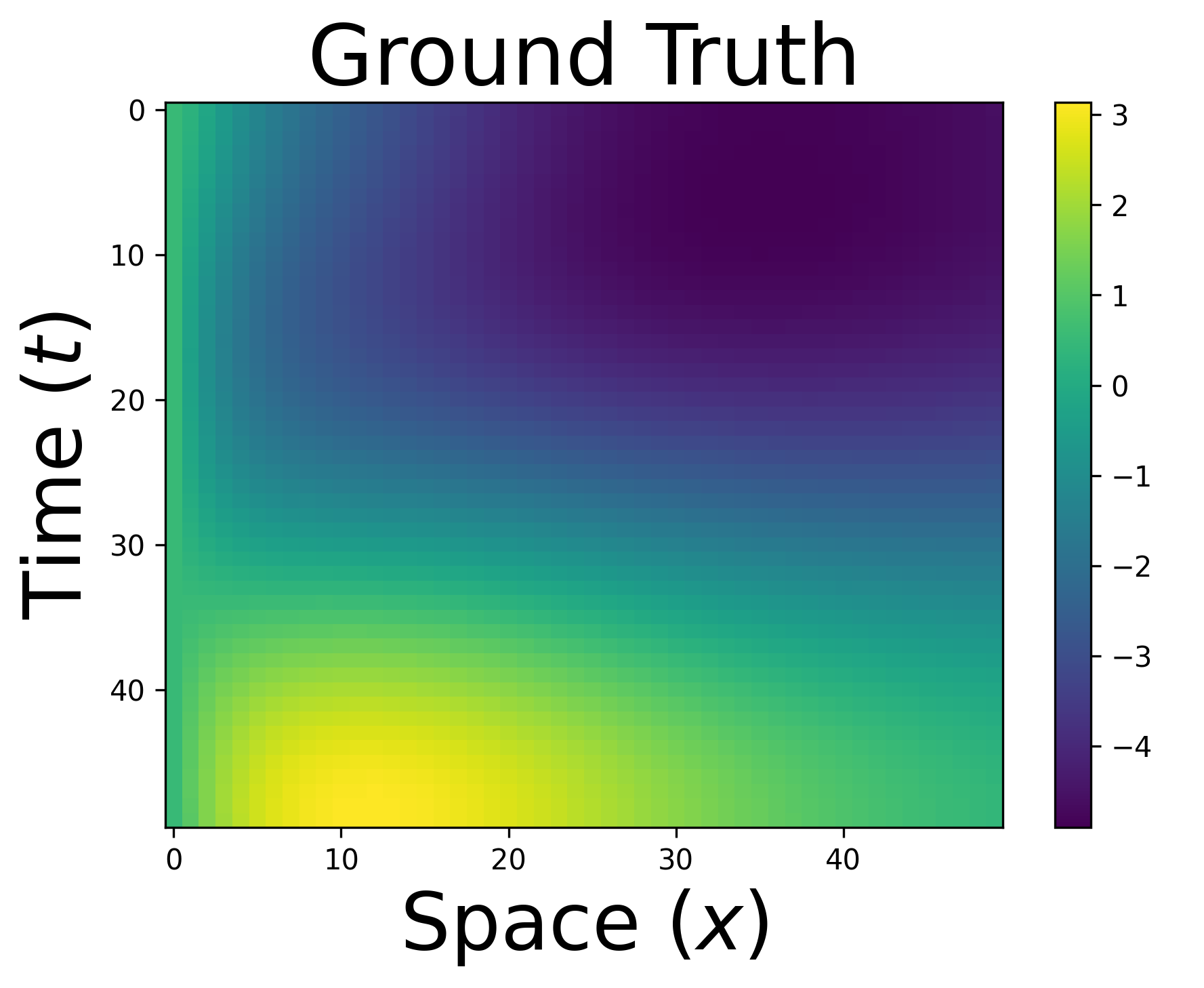} &
        \includegraphics[width=0.17\textwidth]{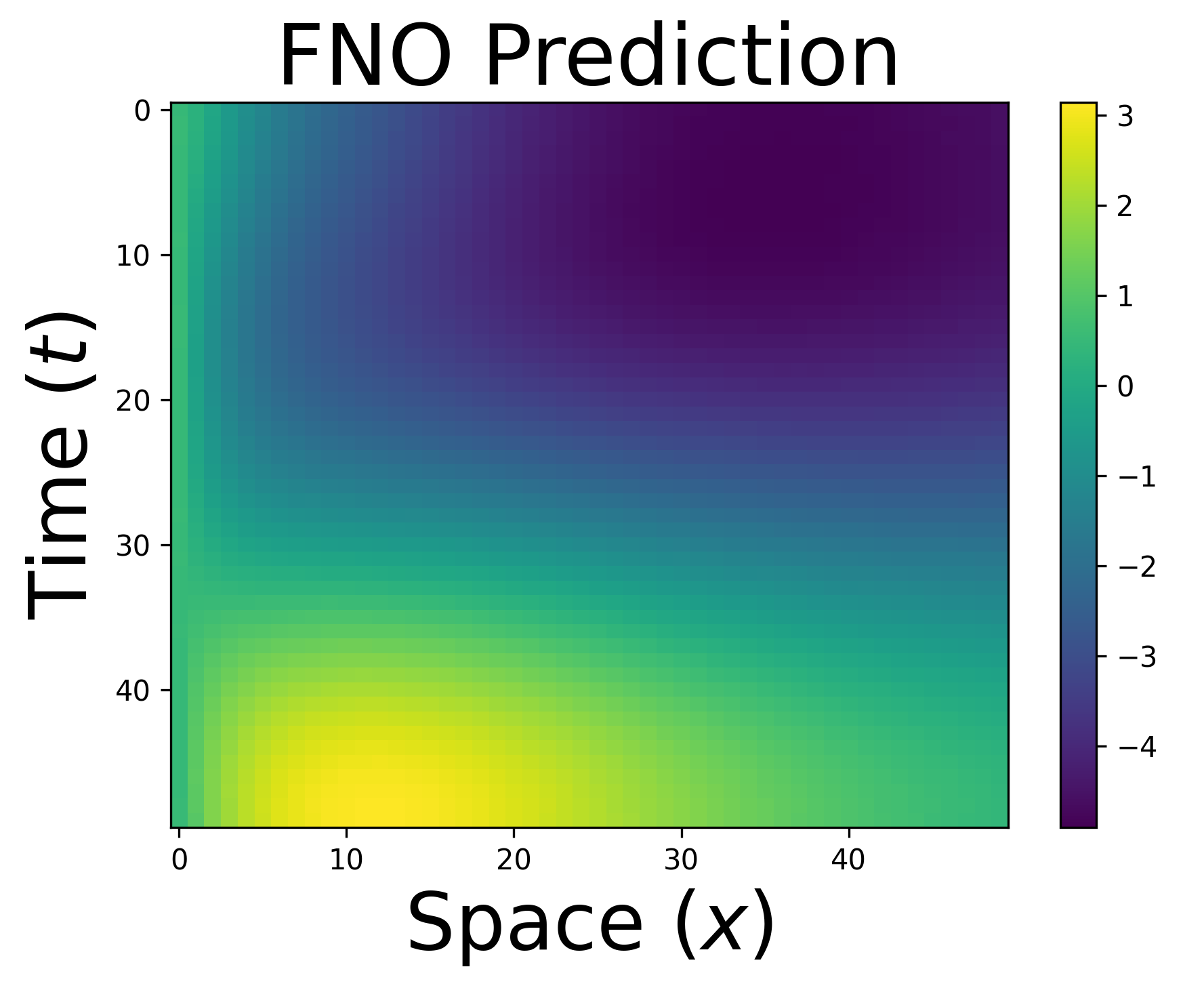} &
        \includegraphics[width=0.17\textwidth]{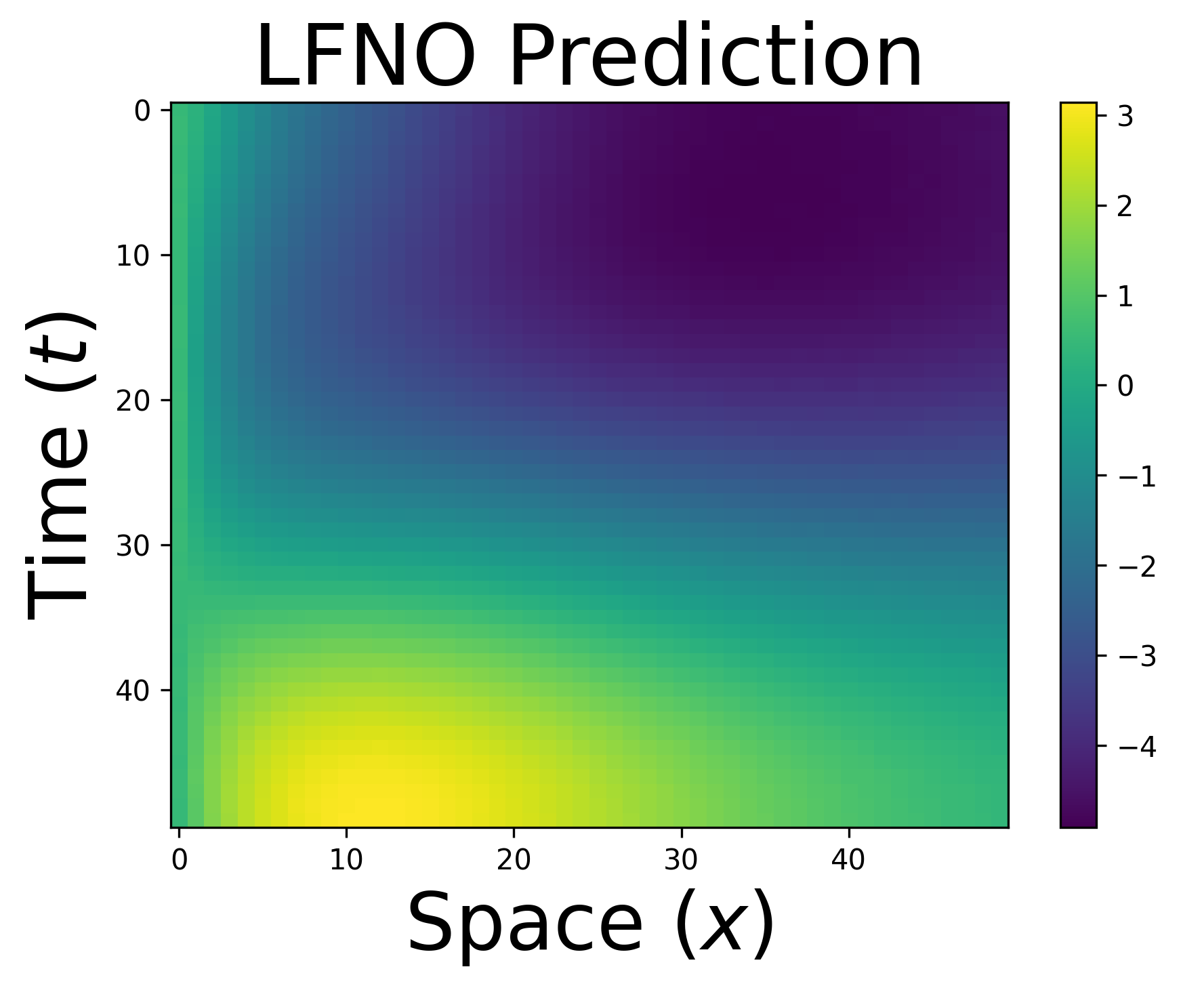} &
        \includegraphics[width=0.17\textwidth]{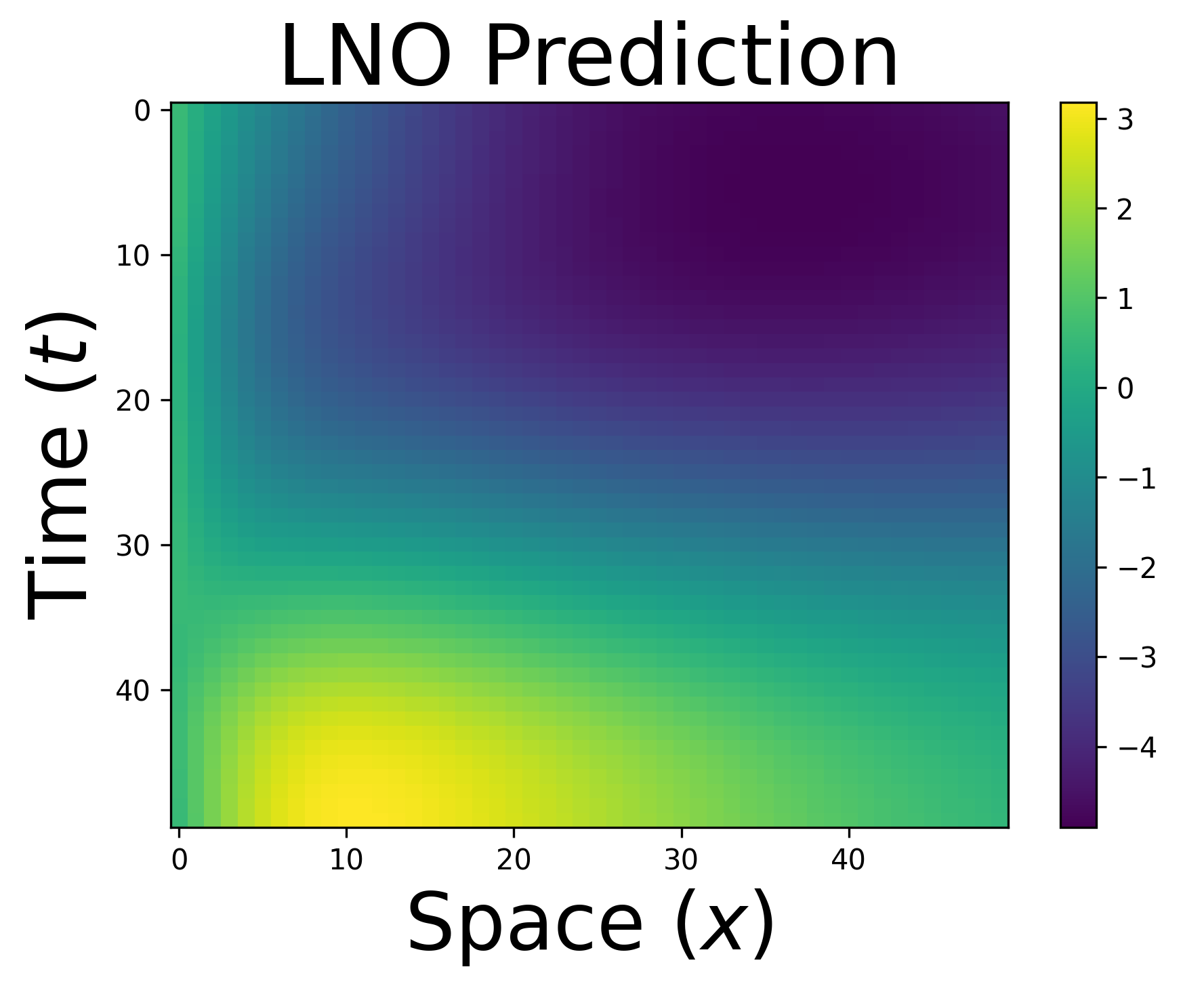} \\[-0.4em]

        Burgers &
        \includegraphics[width=0.17\textwidth]{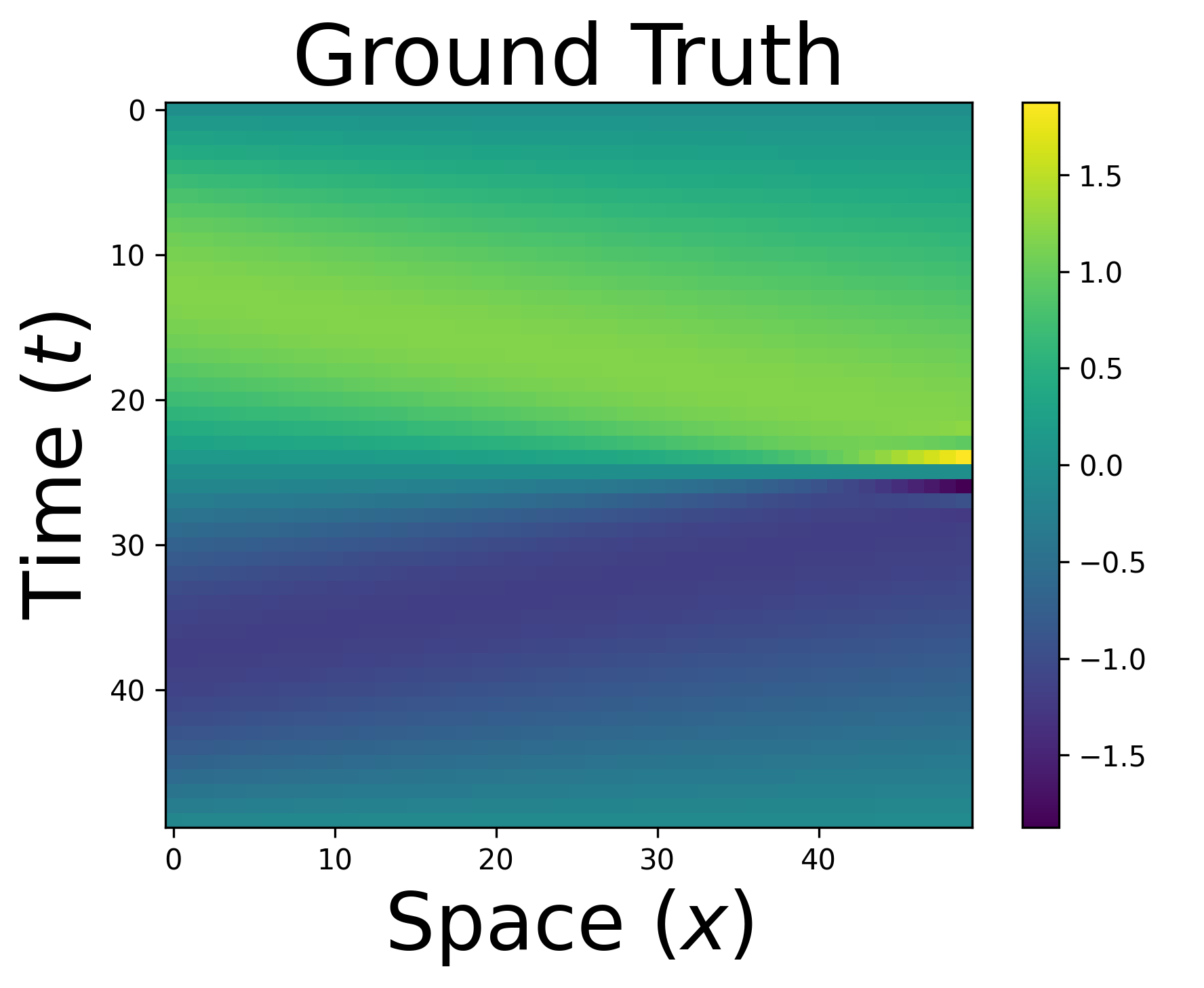} &
        \includegraphics[width=0.17\textwidth]{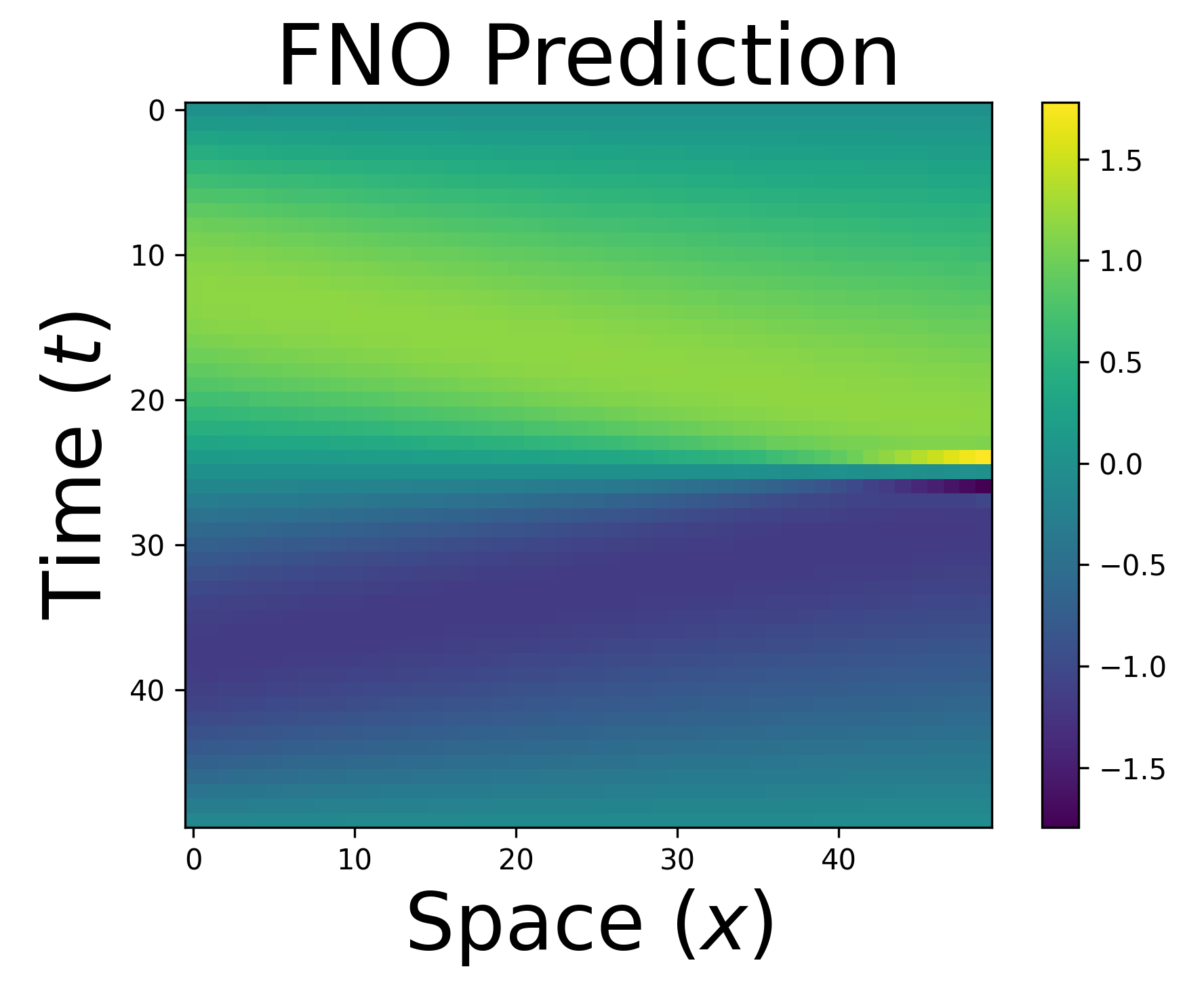} &
        \includegraphics[width=0.17\textwidth]{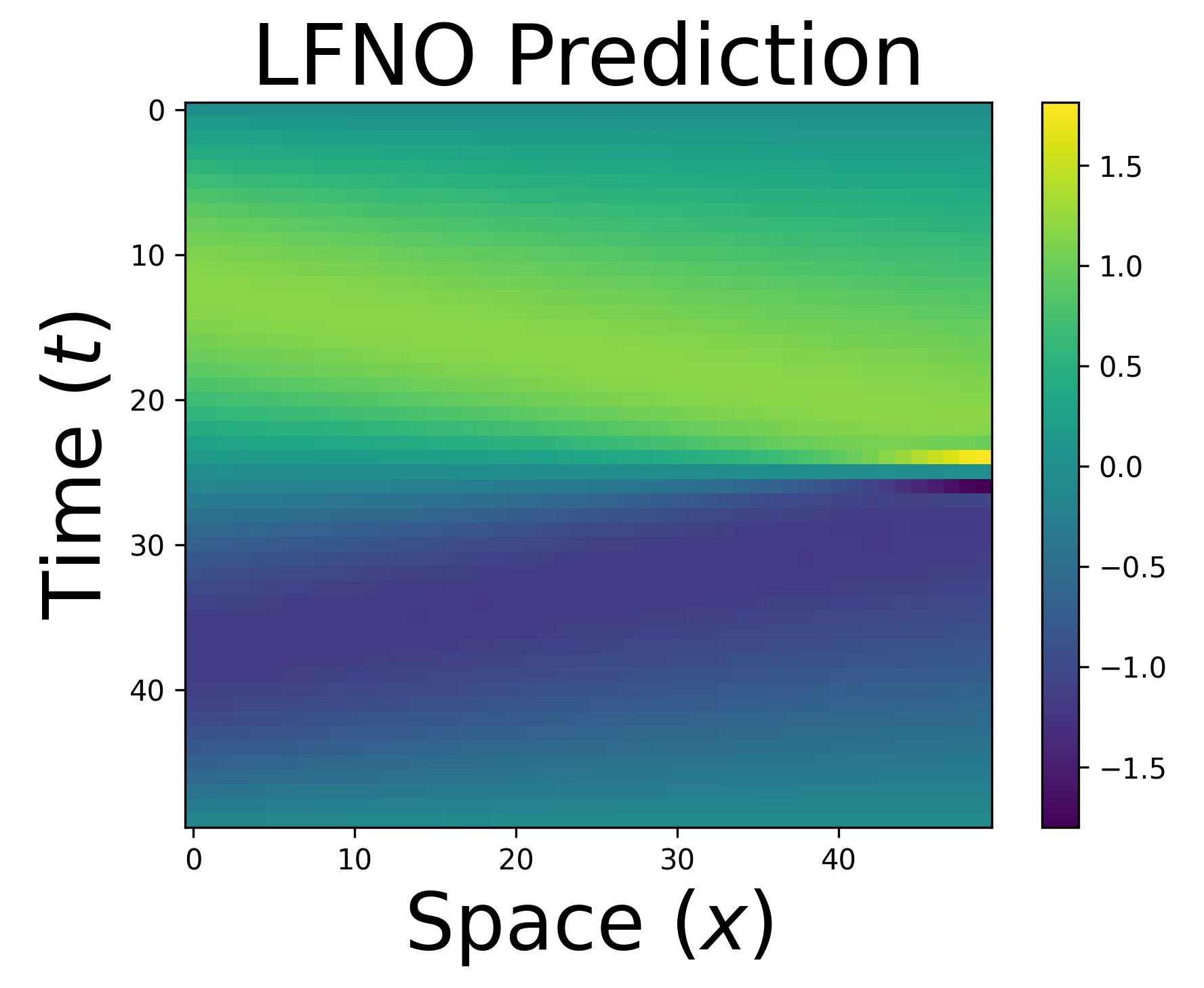} &
        \includegraphics[width=0.17\textwidth]{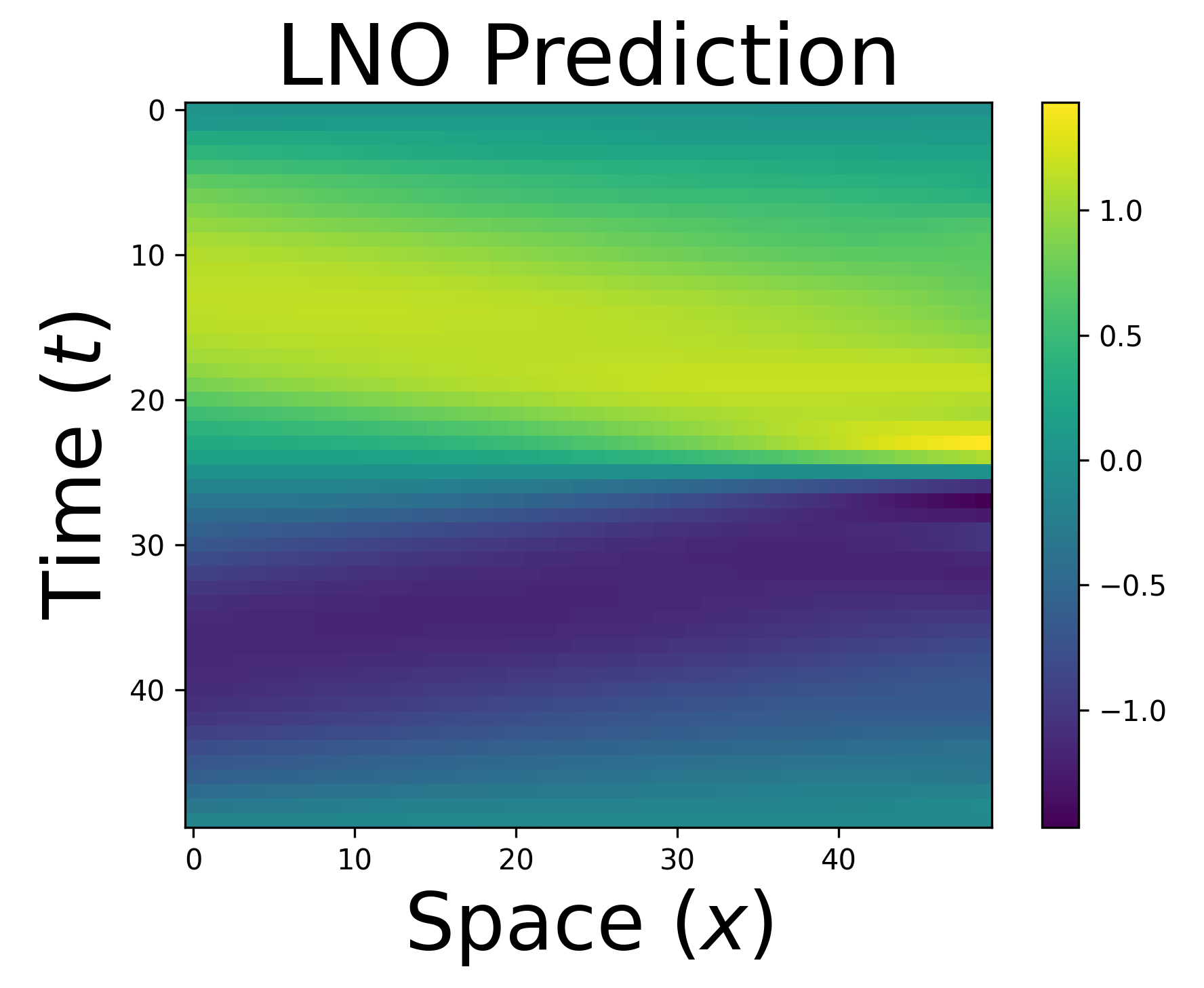} \\[-0.4em]

        NS &
        \includegraphics[width=0.17\textwidth]{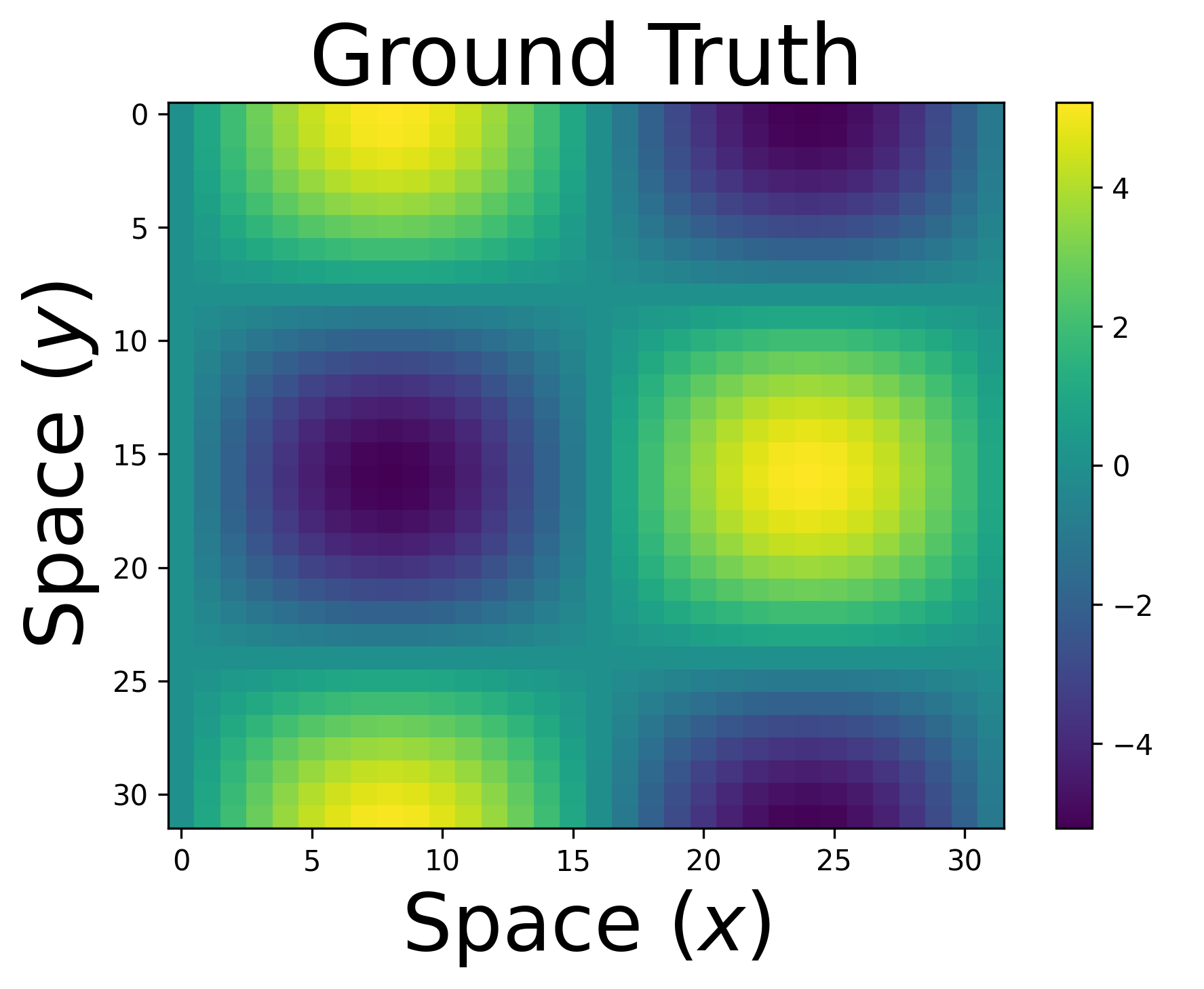} &
        \includegraphics[width=0.17\textwidth]{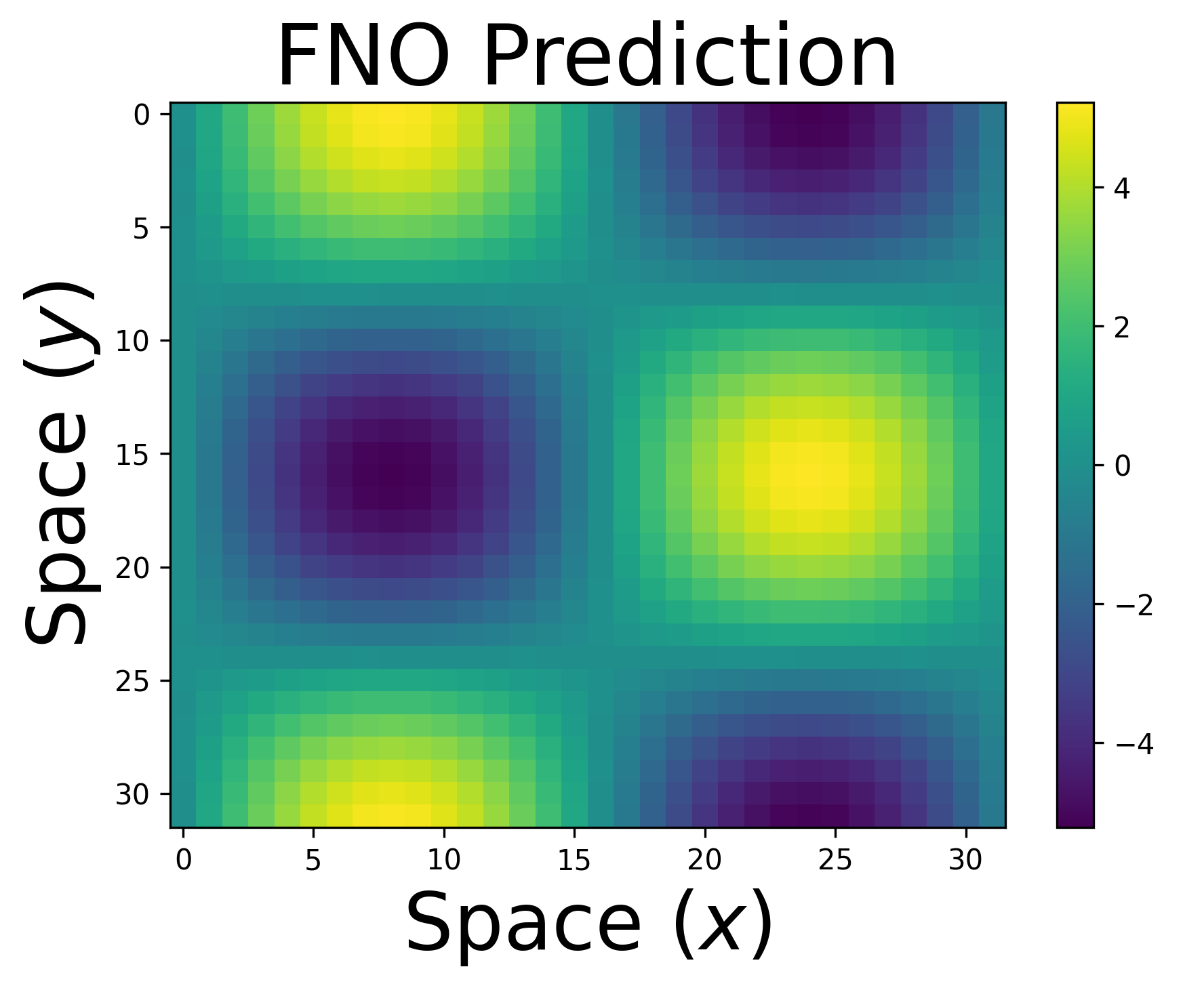} &
        \includegraphics[width=0.17\textwidth]{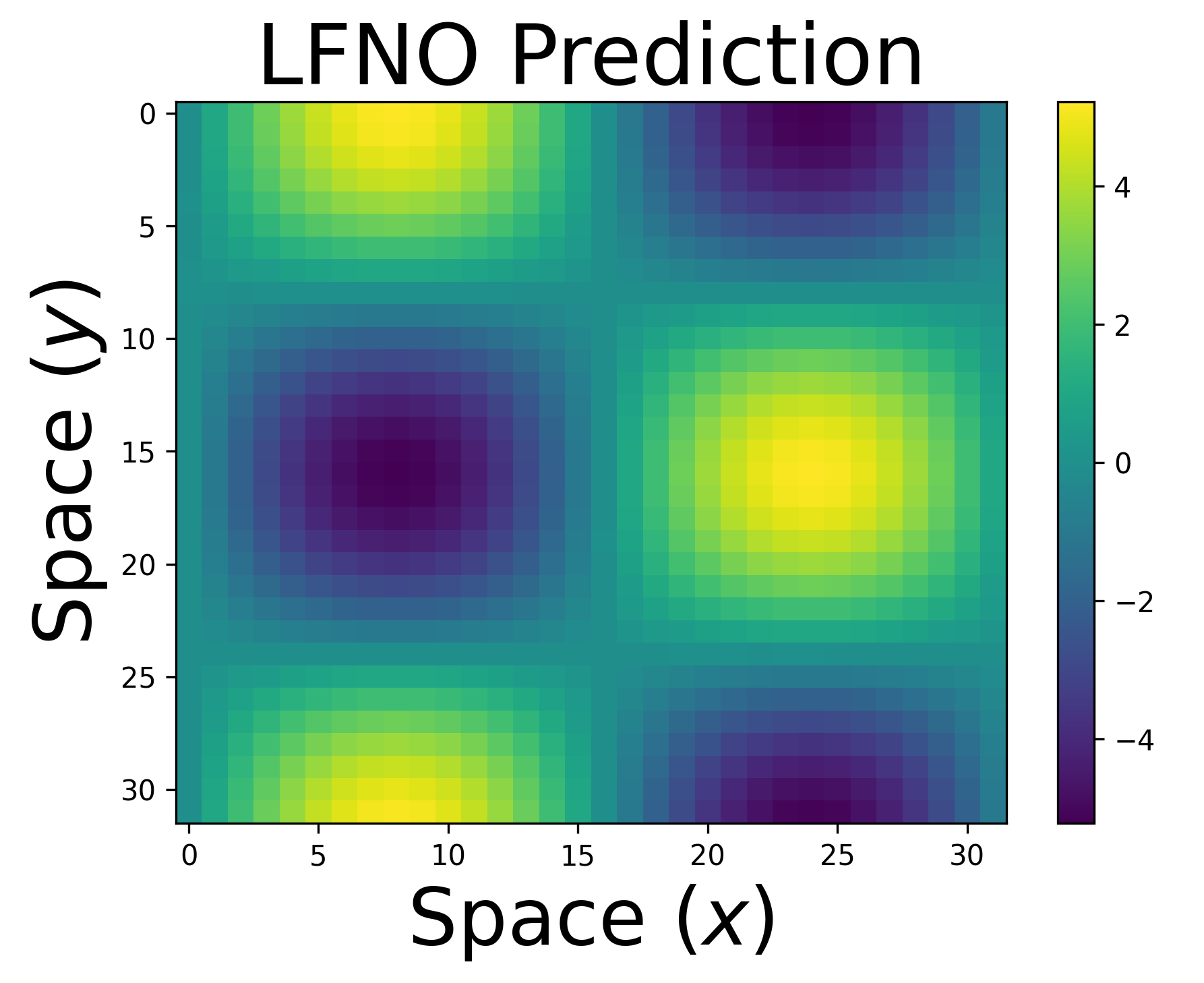} &
        \includegraphics[width=0.17\textwidth]{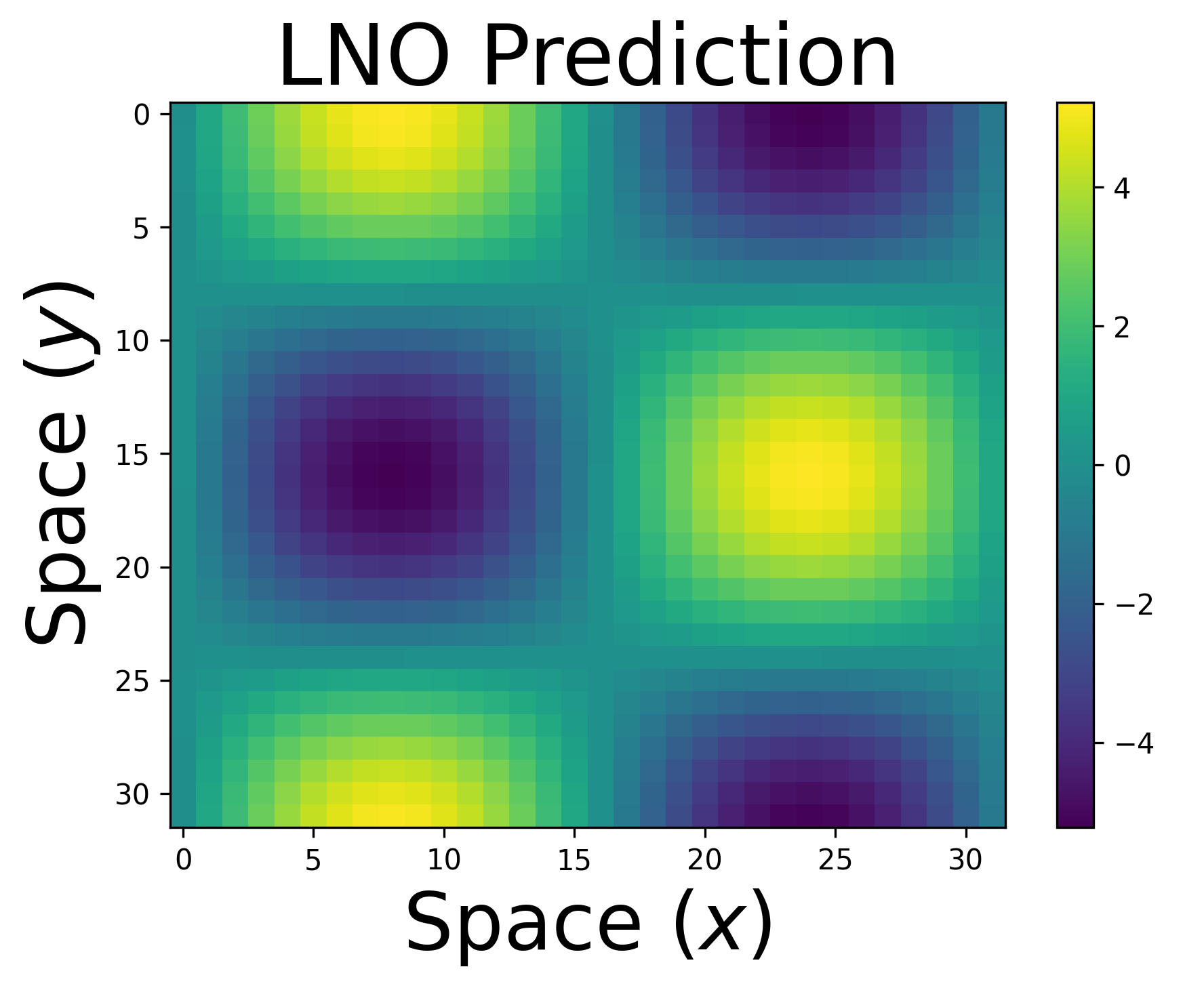}

    \end{tabular}

    \caption{
    Qualitative comparison on six PDE benchmarks. Columns correspond to ground truth (GT), FNO, LFNO, and LNO predictions. Full-resolution results are provided in \cref{sec: figure_PDEs}.
    }

    \label{fig: PDE_qualitative_comparison}
\end{figure*}

\subsection{PDE Experiments}\label{subsec: 2d_pdes_results}

As summarized in \cref{tab: l2_error_table_pde} and \cref{tab: l_infty_error_table_pde}, LFNO demonstrates performance highly competitive with, and in several cases superior to, FNO, while significantly surpassing LNO across all PDE benchmarks. Most notably, in the Burgers' equation, LFNO yields an $\mathcal{L}_{\infty}$ error of 0.0125, representing nearly a four-fold reduction compared to FNO’s 0.0471. This substantial suppression of the maximum point-wise error proves that LFNO’s dual-branch architecture effectively mitigates spectral ringing artifacts and local instabilities near shock fronts—regions where pure Fourier-based methods typically incur high worst-case errors.

The superiority of LFNO becomes even more pronounced in high-complexity regimes, such as the Navier-Stokes equation with lower viscosity ($\nu=10^{-3}$). In this challenging task, LFNO achieves an $\mathcal{L}_2$ error of 0.0026, outperforming FNO (0.0286) by over an order of magnitude. While FNO’s performance degrades significantly as the complexity of the flow increases, LFNO maintains high-fidelity spectral representations with a remarkably low $\mathcal{L}_{\infty}$ error of 0.0071 (compared to FNO’s 0.0411). This demonstrates that LFNO’s dual-branch approach is uniquely capable of resolving intricate, fine-scale vortical structures that emerge in low-viscosity regimes, which single-branch operators often fail to capture.

Qualitative comparisons in \cref{fig: PDE_qualitative_comparison} further underscore this trend; while LNO completely fails to produce meaningful predictions in stiff systems like the Euler-Bernoulli beam, LFNO excels in recovering complex spatiotemporal patterns. These results indicate that by decomposing dynamics into transient and steady components, LFNO ensures a more stable and interpretable learning process, maintaining its worst-case robustness even in stiff dynamical systems characterized by multi-scale spatiotemporal features.

\subsection{Learning Curves}\label{subsec: learning_curve}

\begin{figure}[ht]
    \resizebox{\columnwidth}{!}{%
    \begin{tabular}{cc}
    \centering
    \subcaptionbox{Duffing $c = 0$\label{fig: loss_curve_duffingc0}}{
        \includegraphics[width=0.49\textwidth]{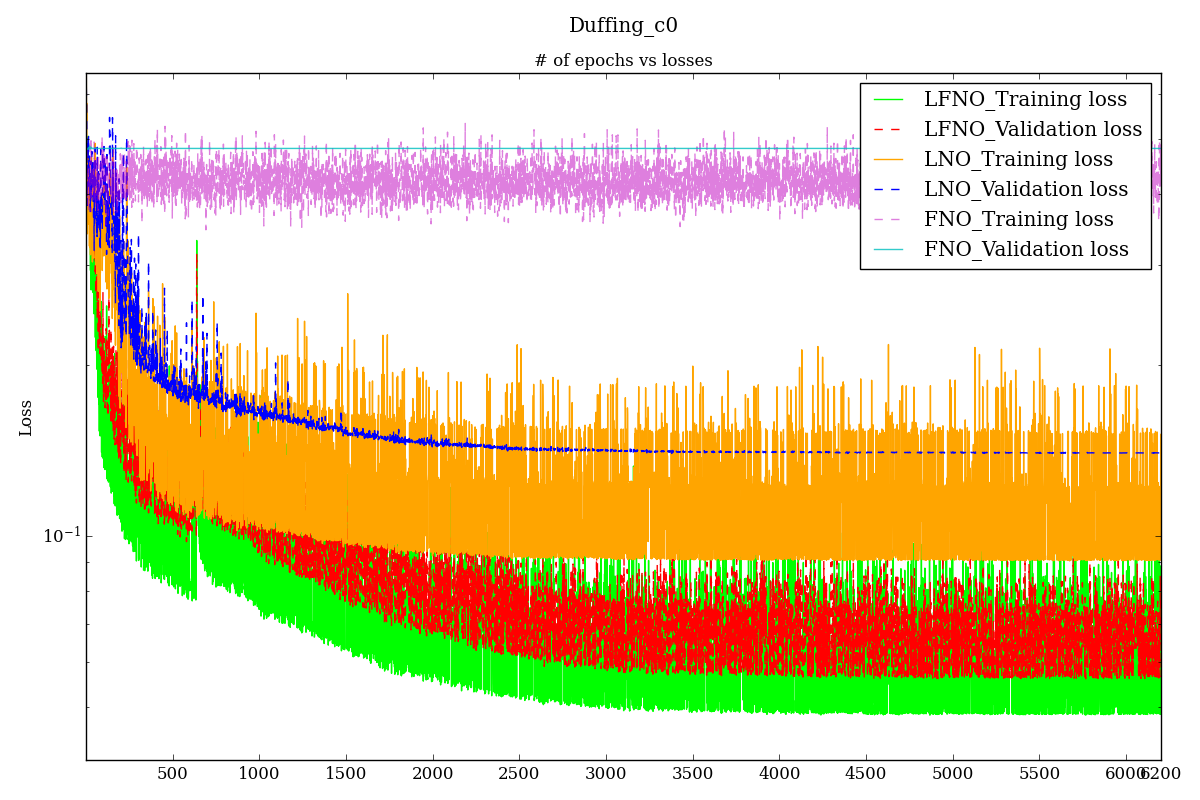}
    } & 
    \subcaptionbox{Duffing $c = 0.5$\label{fig: loss_curve_duffingc05}}{
        \includegraphics[width=0.49\textwidth]{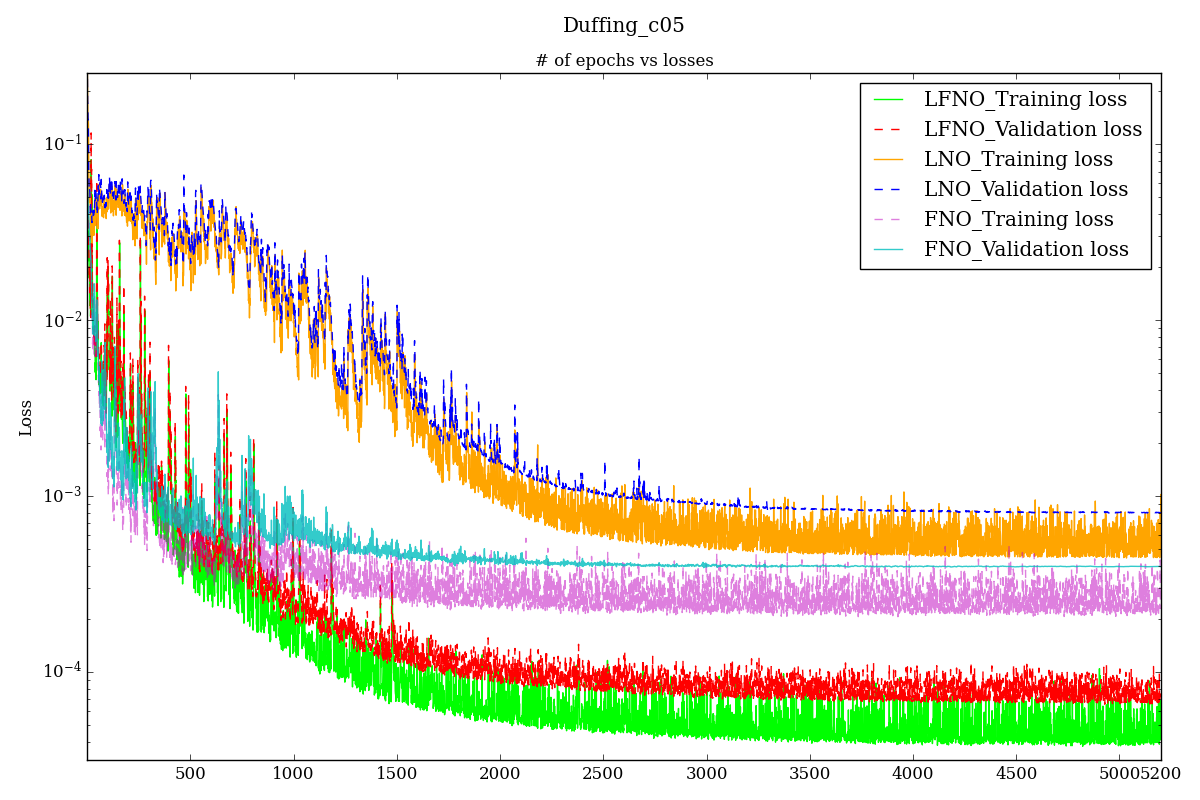}
    }\\[0.5em]
    \subcaptionbox{Lorenz $\rho = 5$\label{fig: loss_curve_lorez_rho5}}{
        \includegraphics[width=0.49\textwidth]{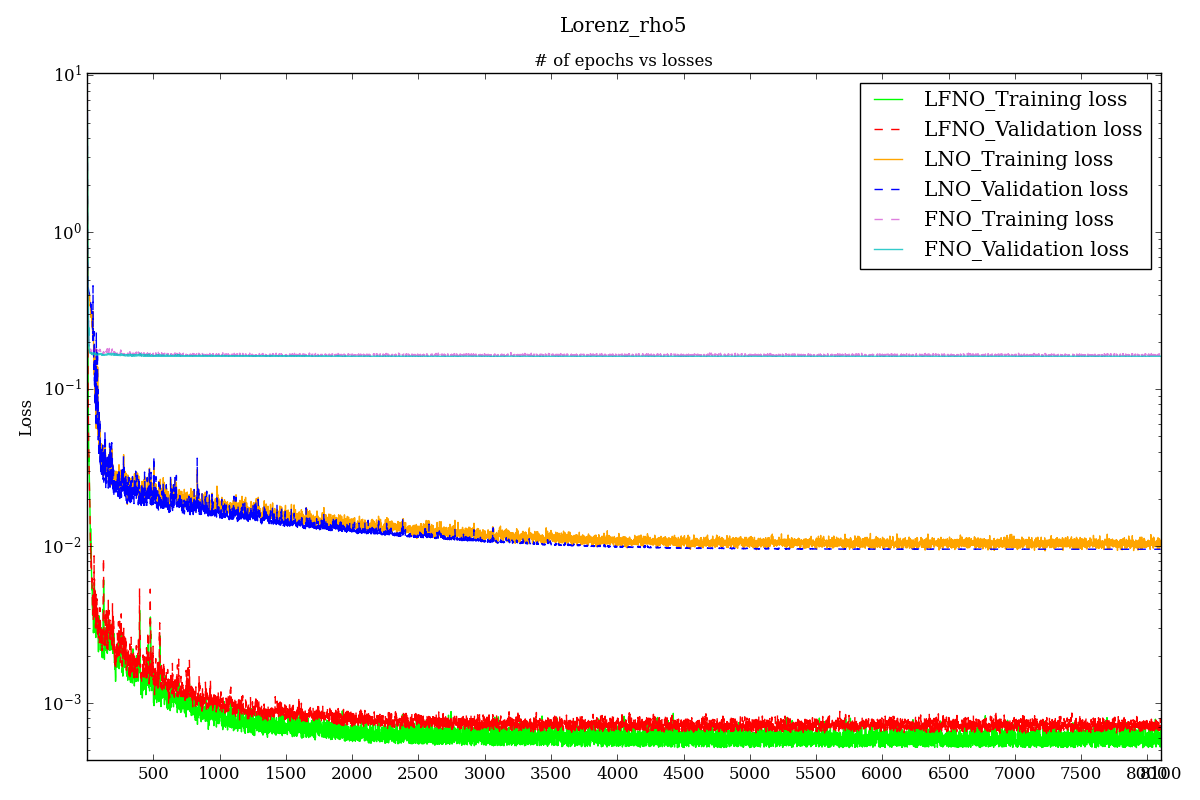}
    } &
    \subcaptionbox{Lorenz $\rho = 10$\label{fig: loss_curve_lorez_rho10}}{
        \includegraphics[width=0.49\textwidth]{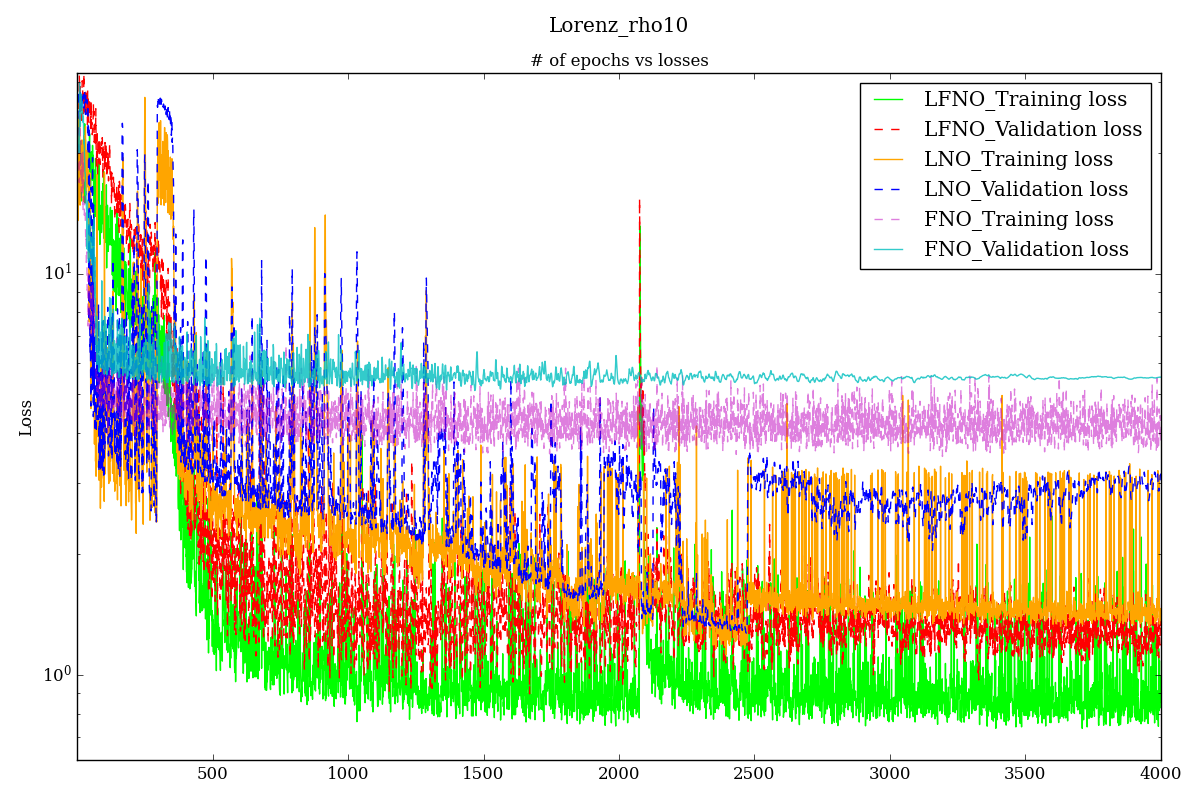}
    }\\[0.5em]
    \subcaptionbox{Pendulum $c = 0.5$\label{fig: loss_curve_pendulumc05}}{
        \includegraphics[width=0.49\textwidth]{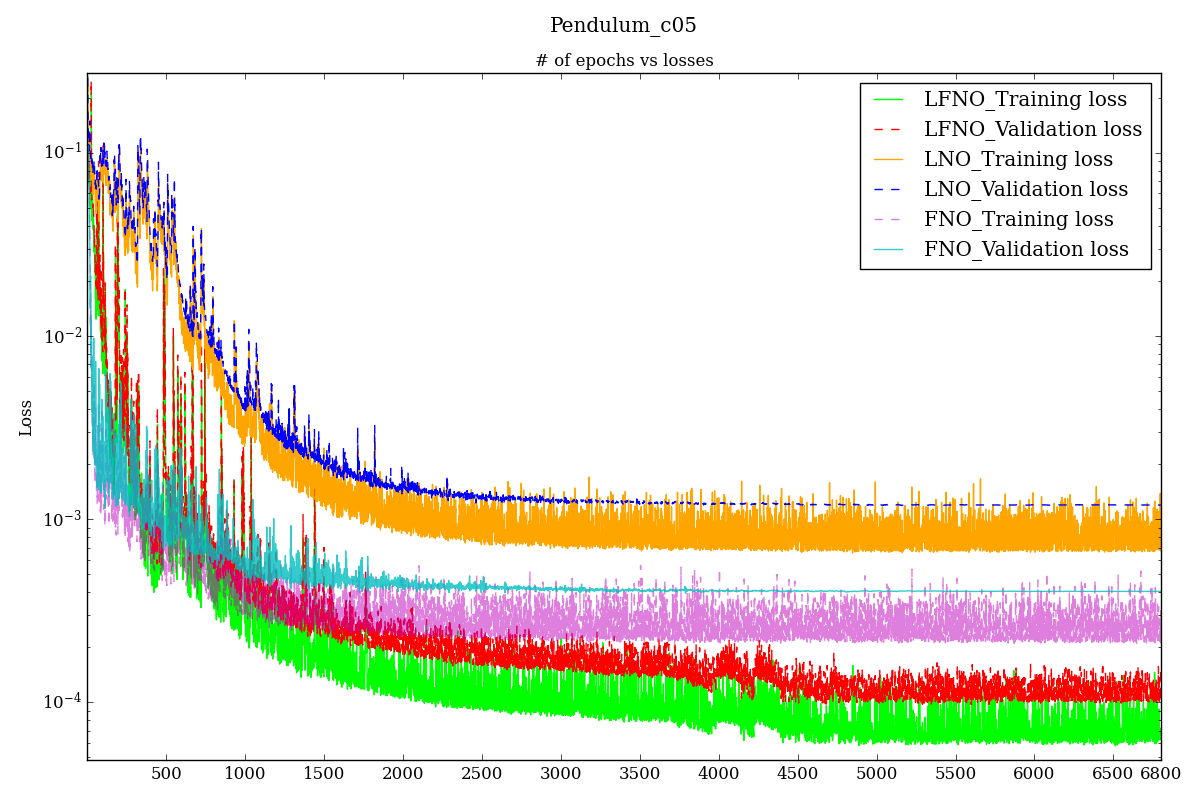}
    }
    \end{tabular}
    }
    \caption{ODE learning curves on different models. Training and validation sets are represented by lime and red for LFNO, orange and blue for LNO, and magenta and cyan for FNO.}
    \label{fig: ODEs_learning_curve}
\end{figure}

\begin{figure}[ht]
    \resizebox{\columnwidth}{!}{%
    \begin{tabular}{cc}
    \centering
    \subcaptionbox{Euler-Bernoulli beam\label{fig: loss_curve_Beam}}{
        \includegraphics[width=0.49\textwidth]{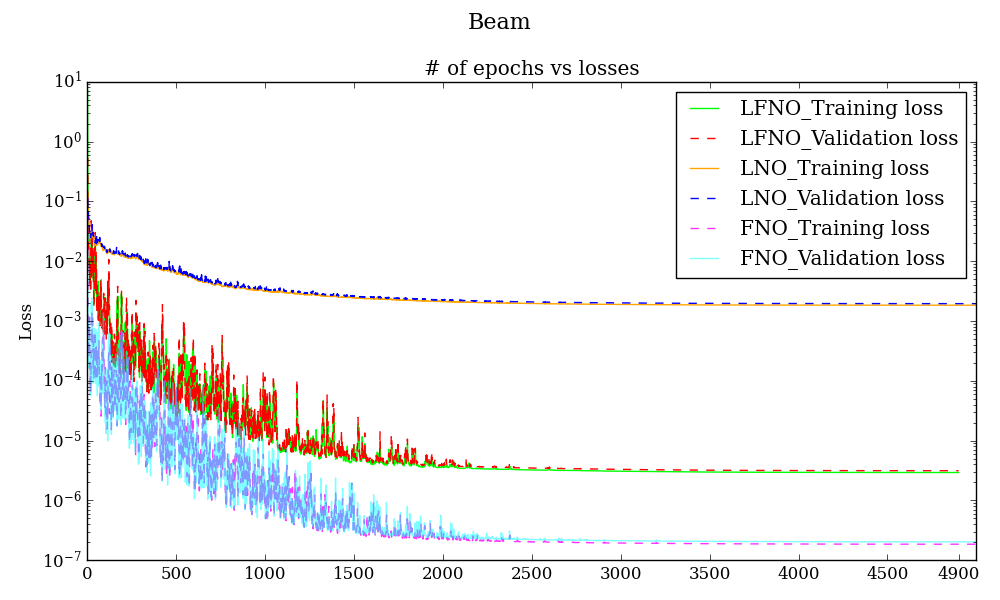}
    } & 
    \subcaptionbox{Heat\label{fig: loss_curve_Diffusion}}{
        \includegraphics[width=0.49\textwidth]{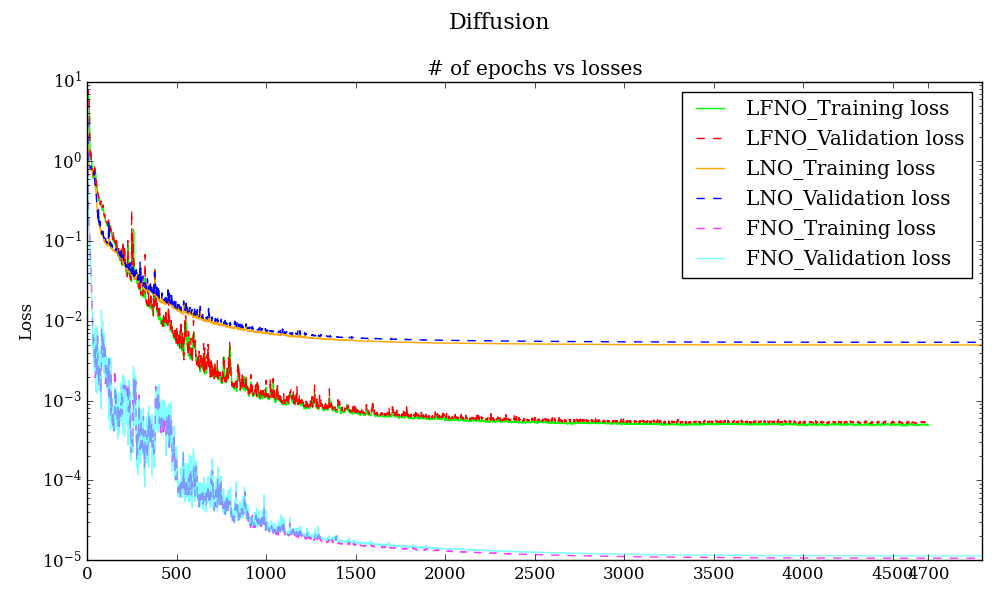}
    }\\[0.5em]
    \subcaptionbox{Reaction-diffusion\label{fig: loss_curve_ReactionDiffusion}}{
        \includegraphics[width=0.49\textwidth]{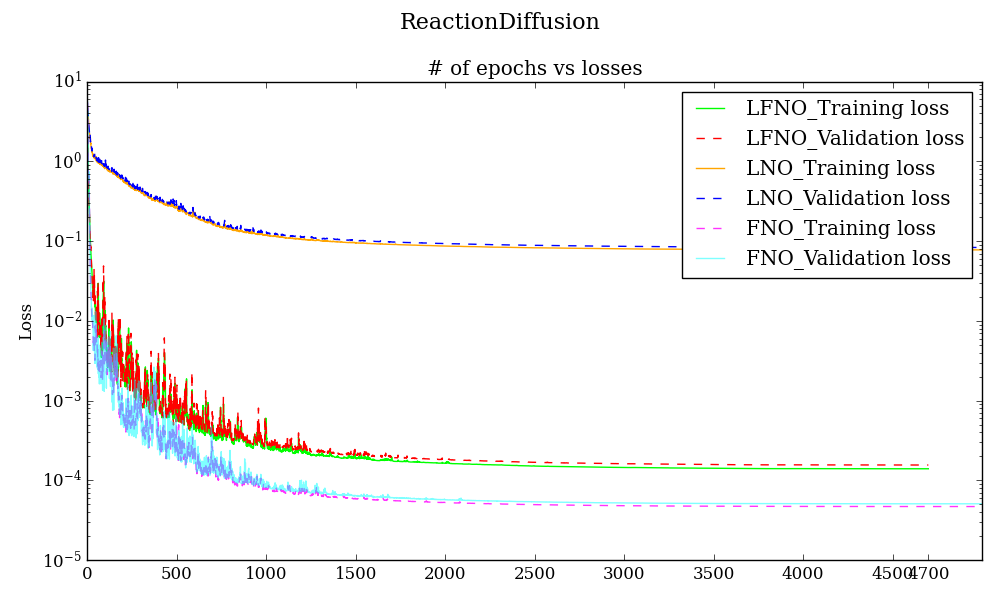}
    } &
    \subcaptionbox{Brusselator\label{fig: loss_curve_Brusselator}}{
        \includegraphics[width=0.49\textwidth]{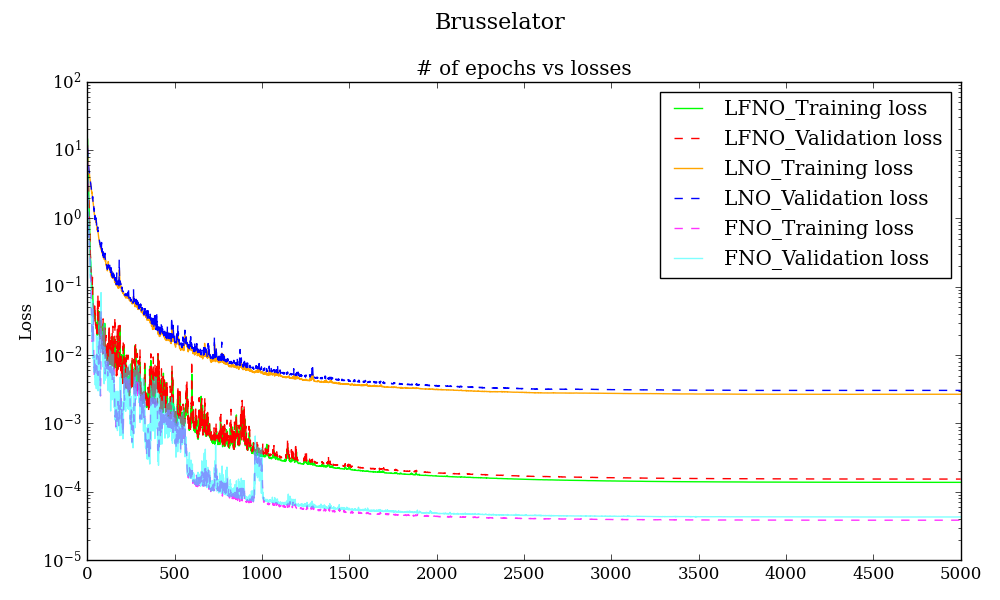}
    }\\[0.5em]
    \subcaptionbox{Burgers\label{fig: loss_curve_Burgers}}{
        \includegraphics[width=0.49\textwidth]{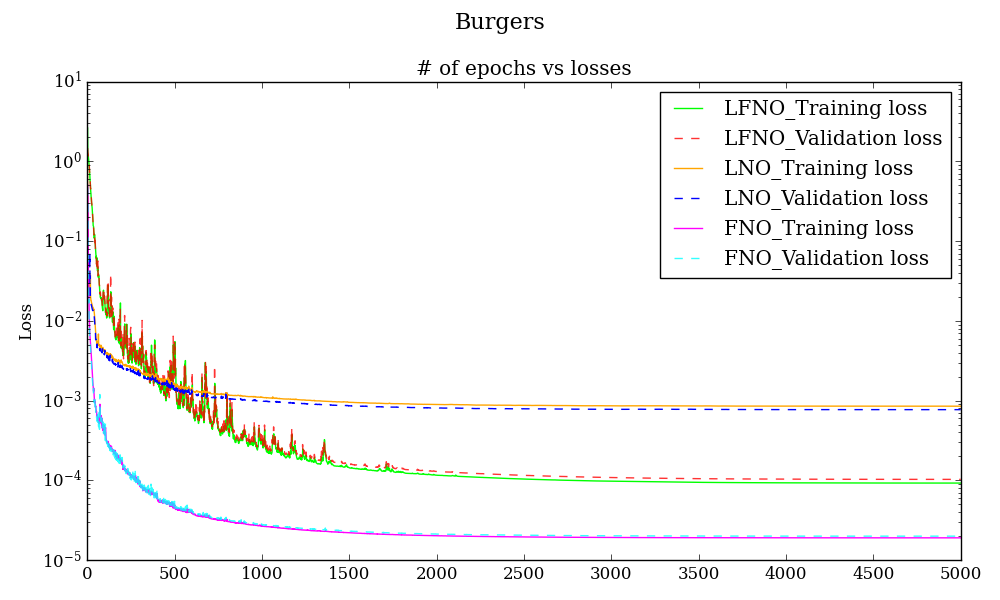}
    } &
    \subcaptionbox{Navier-Stokes\label{fig: loss_curve_Navier}}{
        \includegraphics[width=0.49\textwidth]{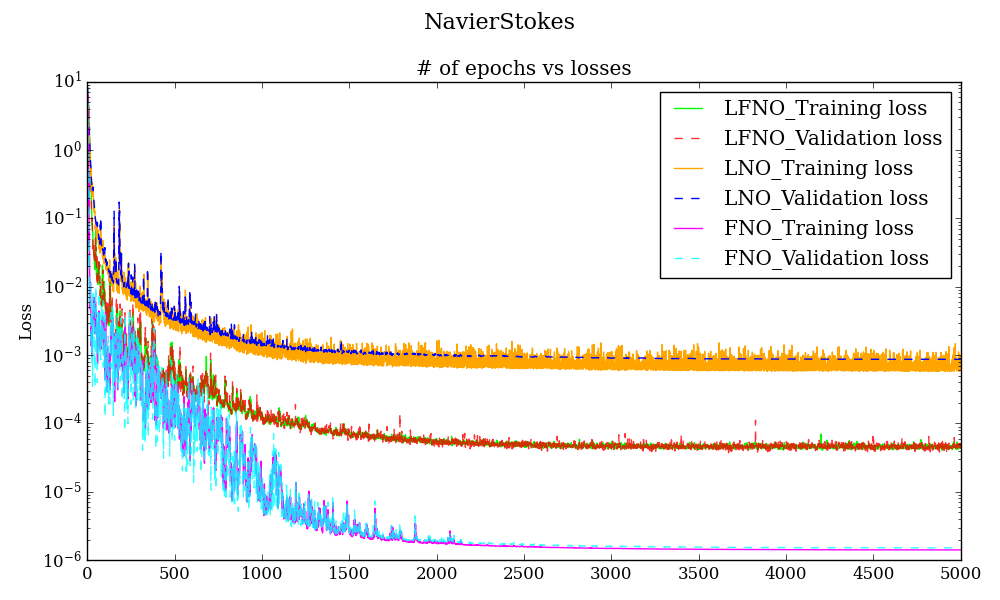}
    }
    \end{tabular}
    }
    \caption{PDE learning curves on different models. Training and validation sets are represented by lime and red for LFNO, orange and blue for LNO, and magenta and cyan for FNO.}
    \label{fig: PDEs_learning_curve}
\end{figure}

We analyze the training dynamics of FNO, LNO, and LFNO by comparing their loss trajectories over the designated training epochs in \cref{fig: ODEs_learning_curve}, \cref{fig: PDEs_learning_curve}. Across all experiments, the training and validation losses are visualized for LFNO (lime, red), LNO (orange, blue), and FNO (magenta, cyan).

In ODE tasks, LFNO demonstrates superior optimization stability compared to the baselines. As illustrated in \cref{fig: loss_curve_duffingc0}, \cref{fig: loss_curve_lorez_rho5}, and \cref{fig: loss_curve_lorez_rho10}, FNO frequently exhibits optimization failures or stagnation, particularly in chaotic regimes. While LNO manages to reduce loss, it displays significant fluctuations in the Lorenz systems, suggesting difficulty in stabilizing the pole-residue approximation. In contrast, LFNO maintains a monotonic and stable decay, indicating that the transient-steady decomposition effectively regularizes the loss landscape, preventing the model from being trapped in suboptimal local minima.

For the PDE experiments, the convergence behavior reveals an intriguing decoupling between training loss and empirical accuracy. In tasks such as the Burgers' and Navier-Stokes equations, FNO reaches the lowest training-loss floor, settling to a decimal order of $10^{-5}$. However, when comparing the final test $\mathcal{L}_2$ errors, LFNO achieves superior performance in \cref{tab: l2_error_table_pde}, despite maintaining a slightly higher validation loss plateau of $10^{-4}$.

This phenomenon suggests that LFNO’s explicit separation of transient and steady-state responses acts as an implicit regularizer. By isolating transient gradients, LFNO prevents the model from overfitting to high-frequency spectral noise, which FNO may capture in its pursuit of a lower loss floor, thereby preserving a more physically consistent representation of the underlying field. Meanwhile, LFNO maintains a convergence plateau consistently one order of magnitude lower than that of LNO, which fails to achieve competitive results in stiff systems such as the Euler-Bernoulli beam in \cref{fig: loss_curve_Beam}.

\section{Discussion}\label{sec: discussion}

While \citet{cao2024laplace} claimed that LNO produces lower validation errors than FNO, our empirical results consistently demonstrate otherwise under a unified benchmarking framework. Interestingly, despite its stated capability to capture transient behavior, LNO exhibited a disproportionately large performance gap in the Euler-Bernoulli beam experiments—a system characterized by high-order spatial derivatives and stiff dynamics—compared to its performance in ODE tasks. We hypothesize that while the pole-residue formulation is theoretically sound, the classical LNO framework encounters significant representation bottlenecks when the system's complexity increases, failing to stabilize the optimization of complex residues.

To address these challenges, LFNO utilizes a dual-branch architecture that isolates transient non-stationarities from steady-state oscillations. While reviewers may note a marginal accuracy gap compared to FNO in certain standard PDEs, we argue that this is not a fundamental performance regression but rather a result of preventing the steady-state module from overfitting to high-frequency spectral noise. By shielding the Fourier branch from transient gradients, LFNO ensures a more stable loss landscape and physically consistent representations, providing a robust compromise between spectral precision and temporal interpretability.

A critical factor contributing to this discrepancy lies in the target mapping consistency within the datasets. In the original LNO datasets, we observed that the forcing functions used to generate training samples differed from those used for validation. From a functional analysis perspective, this implies that the model is tasked with learning a transformation where the target operator shifts between the training and evaluation phases. Such a mismatch forces the neural operator to compensate for inconsistent physical mappings rather than learning the underlying differential operator. To eliminate this confounding factor, we constructed datasets where the source functions and their corresponding target responses are strictly aligned across all splits. Under this rigorous setup, the Euler-Bernoulli beam equation reveals substantially stronger transient effects than those found in conventional benchmarks. Our findings suggest that the perceived advantages of previous models might have been, in part, an artifact of inconsistent data configurations that failed to fully challenge the models' ability to resolve intricate transient transitions.

Looking forward, the success of bridging Laplace and Fourier domains suggests a broader potential for hybrid basis expansion in neural operator learning. While LFNO effectively resolves the decay-oscillation trade-off, future work could focus on integrating additional orthogonal bases, such as Wavelets for localized multiscale features or Chebyshev polynomials for non-periodic boundary conditions. By developing a general methodology to hybridize multiple spectral and temporal bases, we can create operators that are tailored to even more complex, non-homogeneous spatiotemporal dynamics. Such an approach would extend the utility of LFNO from a specific solution for transient-steady separation to a universal framework for adaptive basis learning in high-dimensional physical systems.

\section{Conclusion}\label{sec: conclusion}

In this work, we propose LFNO, a unified framework that integrates the Laplace and Fourier Neural Operators to resolve the inherent trade-offs between transient and steady-state modeling. By leveraging pole-residue dynamics within the Laplace domain and the periodic expressivity of the Fourier integral operator, LFNO provides a comprehensive representation of complex physical signals. The strategic allocation of model capacity toward the transient module, combined with nonlinear steady-state representations, enables LFNO to capture a broader range of spatiotemporal dynamics than conventional operators.

Our experimental results demonstrate that LFNO not only achieves a principled theoretical integration but also delivers empirically robust performance across diverse ODE and PDE benchmarks. By constructing datasets with consistent target mappings, we isolate and expose LFNO’s advantages in modeling intricate transitions that were previously obscured in inconsistent data configurations. Overall, these results establish LFNO as a highly competitive operator learning framework, demonstrating superior generalization over existing baselines in complex, non-periodic spatiotemporal tasks. LFNO highlights a versatile foundation for future research on hybrid spectral methods and adaptive basis expansion in high-dimensional physical dynamics by explicitly bridging the Laplace and Fourier domains.

\section*{Impact Statement}

This paper presents a methodological advancement in neural operator architectures for modeling complex dynamical systems. By explicitly decomposing transient and steady-state responses, the proposed LFNO framework provides enhanced accuracy and robustness in solving PDEs. We anticipate that this work will have a positive impact on scientific computing, particularly in domains requiring high-fidelity simulations such as climate modeling, aerospace engineering, and fluid dynamics. Furthermore, the computational efficiency of our approach could contribute to reducing the energy footprint associated with large-scale numerical simulations on high-performance computing resources.

While we do not foresee direct negative societal impacts, we acknowledge that reliance on purely data-driven or hybrid models requires careful validation. Improper deployment in safety-critical systems without domain-specific oversight could lead to unreliable predictions. We encourage practitioners to use this method as a complementary tool alongside traditional numerical solvers to ensure physical consistency and reliability.


\bibliography{references}
\bibliographystyle{icml2026}

\newpage
\appendix
\onecolumn

\section{Implementation Details}\label{sec: implementation_details}
In this section, we introduce the implementation details. For all models, the Adam optimizer with step learning rate scheduling is used to improve stability and better convergence. We trained the model on an NVIDIA RTX 3090 GPU. The model hyperparameters are listed in \cref{tab: Hyperparameter_table}. We particularly employed the early stopping method for all experiments and compared the corresponding LNOs with those from the same number of epochs iterated in the LNO experiment. We also present the computational cost for each training in \cref{tab: computational_cost}. To ensure precise measurement of pure GPU kernel execution time, the average inference latency was recorded over 100 iterations after 10 warm-up runs, with explicit synchronization using torch.cuda.synchronize(). We replaced Model Usage with Model VRAM footprint to provide a more precise measure of the peak hardware memory required during the model's actual deployment and inference process.

\begin{table}[ht]
    \centering
    \resizebox{\textwidth}{!}{
    \begin{tabular}{c|c|c|c|c|c|c}
    \toprule
        Application & FNO ODEs & LNO ODEs & LFNO ODEs & FNO PDEs & LNO PDEs & LFNO PDEs\\
    \midrule
         Layer & 4 & 1 & 6 & 4 & 1 & 6 \\
    \midrule
         Width & 4 & 4 & 4 & 16 & 16 & 16 \\
    \midrule
         Mode & 16 & 16 & 16 & 4 & 4 & 4 \\
    \midrule
         Train batch size & 64 & 64 & 64 & 50 & 50 & 50 \\
    \midrule         
         Learning rate & 0.0025 & 0.0025 & 0.0025 & 0.002 & 0.002 & 0.002 \\
    \midrule
         weight decay & 0.02 & 0.02 & 0.02 & 0.0001 & 0.0001 & 0.0001 \\
    \midrule     
         scheduler step size & 100 & 100 & 100 & 100 & 100 & 100 \\
    \midrule     
         gamma & 0.85 & 0.85 & 0.85 & 0.8 & 0.8 & 0.8 \\
    \midrule
         Activation function & relu & sin & relu & relu & sin & relu \\
    
    \bottomrule
    \end{tabular}
    }
    \vskip 0.1in
    \caption{Hyperparameters used in the FNO, LNO, and LFNO for training.}
    \label{tab: Hyperparameter_table}
\end{table}

\begin{table}[ht]
    \centering
    \begin{tabular}{c|c|c|c}
    \toprule
        Computational cost & FNO & LNO & LFNO \\
    \midrule
        Parameter count & 7,537 & 1,309 & 3,417 \\
    \midrule
        Inference time (ms) & 0.9195 & 0.6664 & 1.8522 \\
    \midrule
        Multiply-Accumulate (MACs) & 146.95 & 2.80 & 16.48\\
    \midrule
        FLOPs ($10^7$) & 0.812 & 0.636 & 1.592 \\
    \midrule
        Model VRAM footprint (MB) & 46.0 & 24.0 & 48.0 \\
    \bottomrule
    \end{tabular}
    \vskip 0.1in
    \caption{Computational cost for FNO, LNO, and LFNO training.}
    \label{tab: computational_cost}
\end{table}

\section{Dataset Configuration}\label{sec: dataset_configuration}

\subsection{ODEs}\label{subsec: 1d_odes_configuration}

\paragraph{Duffing equation}\label{para: Duffing_equation}
The Duffing equation is a nonlinear second-order differential equation describing the driven oscillator. The Duffing equation is written as follows.
\begin{equation}\label{eq: duffing_equation}
    \ddot{x} + c\dot{x} + \alpha x + \beta x^3 = f(t),
\end{equation}
where $c$ is the damping coefficient, $\alpha$ is the linear stiffness, $\beta$ is the non-linear coefficient, and $x, \dot{x},$ and $\ddot{x}$ are the displacement, velocity, and acceleration variables of the dynamic response. We set the parameters as $\alpha=1$ and $\beta=1$. The initial condition is set to $x(0)=0$ and $\dot{x}(0)=0$. To evaluate LFNO's ability to resolve both undamped persistent oscillations and transient decay dynamics, we conduct experiments with damping coefficients of $c=0$ and $c=0.5$.

\paragraph{Pendulum equation}\label{para: Pendulum_equation}
The Pendulum equation describes the angular displacement $x(t)$ of a pendulum relative to its rest position under an external driving force $f(t)$. The equation can be expressed as follows.
\begin{equation}\label{eq: pendulum_dquation}
    \ddot x + c\dot x + \frac{g}{l}\sin(x)=f(t).
\end{equation}
Here, $g$ is the acceleration due to gravity, $l$ is the length of the pendulum, and $c$ is the damping coefficient. For brevity, we use $g/l=1$ and the initial condition of $x(0)=0$ and $\dot{x}(0) =0$. We experiment with the damping coefficient $c = 0.5$.

\paragraph{Forced Lorenz system}\label{para: Lorenz_system}
The forced Lorenz system is the set of three coupled nonlinear differential equations representing simplified atmospheric convection.
\begin{equation}\label{eq: lorenz_system}
    \begin{aligned}
        \dot x &= \gamma (y-x) \\ 
        \dot y&= x(\rho -z)-y \\ 
        \dot z &= xy-\beta z - f(t),
    \end{aligned}
\end{equation}
where $\gamma$, $\rho$, and $\beta$ are physical parameters. To evaluate the framework's robustness in chaotic regimes and its sensitivity to parameter variations, we fix $\gamma=10$ and $\beta=8/3$ while setting the Rayleigh number $\rho$ to $5$ and $10$.  Both experiments are initialized from the state $x(0)=1, y(0)=0, z(0)=0$.

\subsection{PDEs}\label{subsec: 2d_pdes_configuration}

\paragraph{Euler-Bernoulli beam equation}\label{para: Beam_equation}
The deflection $y(x, t)$ of a homogeneous Euler-Bernoulli beam of 1D with forcing function $f(x,t)$ is given as follows:
\begin{equation}\label{eq: beam_equation}
    EI\frac{\partial^4 y}{\partial x^4} + \rho A \frac{\partial^2 y}{\partial t^2} = f(x, t),
\end{equation}
where $EI, \rho,$ and $A$ denote bending stiffness, density, and cross-sectional area, respectively. The forcing function is defined as $f(x, t)=-99Ae^{-x}\sin(10t)$, which represents a periodic spatially decaying excitation. We use the parameters of $EI=1.334$, $\rho=7850$, and $A=0.01$.

\paragraph{Heat equation}\label{para: Diffusion_equation}
The heat equation describes the spatiotemporal evolution of a temperature field $y(x,t)$ driven by a source term $f(x, t).$ The governing equation is given as follows:
\begin{equation}\label{eq: diffusion_equation}
    \frac{\partial^2y}{\partial x^2}-\frac{\partial y}{\partial t}=f(x, t).
\end{equation}
Here, we use the source term of $f(x,t)=Ae^{-t}(1-\pi^2)\sin(\pi x)$.

\paragraph{Reaction-diffusion equation}\label{para: Reaction_diffusion_equation}
Reaction-diffusion systems model the chemical concentration $y(x, t)$ under the influence of both diffusion and nonlinear reaction kinetics. In this experiment, we use the following governing equation:
\begin{equation}\label{eq: reaction_diffusion_equation}
    D\frac{\partial^2y}{\partial x^2}+ky^2-\frac{\partial y}{\partial t}=f(x,t),
\end{equation}
where $y(x,t)$ is the concentration of the chemical, $D=1-0.95\pi^2$ is the diffusion coefficient, and $k=1$ is the rate of reaction. The source term $f(x,t)=Ae^{-t}(1-\pi^2)\sin(\pi x)+A^2e^{-2t}\sin(\pi x)^2$ is configured to produce complex nonlinear interactions between the decaying forcing and the term $y^2$, challenging the model's spectral expressivity.

\paragraph{Brusselator equation}\label{para: Brusselator}
The Brusselator is a prototypical model for autocatalytic oscillating reactions, describing the concentrations of two interacting chemicals, $u(x, t)$ and $v(x, t)$. The system is governed by:
\begin{equation}\label{eq: brusselator}
    \begin{aligned}
        \frac{\partial u}{\partial t}&=D_0 \frac{\partial^2u}{\partial x^2} + a + f(x, t) - (1+b)u+vu^2 \\
        \frac{\partial v}{\partial t}&=D_1\frac{\partial^2v}{\partial x^2} + bu - vu^2,
    \end{aligned}
\end{equation}
where $D_0=1$ and $D_1=0.5$ are diffusion coefficients and $a=1, b=3$ are the parameters of the system. The forcing function is $f(x, t)=Ae^{-t}(1-\pi^2)\sin(\pi x).$

\paragraph{Burgers' equation} \label{para: burgers}
Burgers' equation serves as a fundamental model capturing the interplay between nonlinear advection and diffusion. Unlike the previously described benchmarks, this is an unforced system where the field evolves solely from its initial state:
\begin{equation}\label{eq: burgers}
    \frac{\partial u}{\partial t} + u \frac{\partial u}{\partial x} = \nu \frac{\partial^2u}{\partial x^2}
\end{equation}
where $\nu=0.01/\pi$ is the diffusion coefficient. We initialize the system with $u(x, 0)=Ux$, where $U \sim \mathcal{U}[1.6, 2.4)$.

\paragraph{Navier-Stokes equation} \label{para: navier-stokes}
The Navier-Stokes equation governs the motion of incompressible viscous fluids. In our experiment, the system representing the vorticity $\omega$ is given by:
\begin{equation}\label{eq: navier-stokes}
\begin{aligned}
    &\frac{\partial \omega}{\partial t} + (\mathbf u \cdot \nabla )\omega = \nu \nabla ^2\omega \\[6px]
    &\nabla^2\psi =-\omega\\
    &\mathbf u = \left\langle \frac{\partial \psi}{\partial y}, -\frac{\partial \psi}{\partial x}  \right\rangle.
\end{aligned}
\end{equation}
Here, $\nu = 0.001$ denotes a kinematic viscosity. The initial condition is $\omega(x, y, 0)=V\sin(x)\cos(y),$ where $V$ is sampled uniformly from $[0.8, 1.2).$

\begin{table}[t]
  \centering 
  \footnotesize 
  \setlength{\tabcolsep}{4pt} 
  
  \begin{tabular}{lcccc} 
    \toprule
    Task & Boundary conditions & \makecell{Initial values \\ (for all $x$)} & \makecell{Spatial \\ Res.} & $\Delta t$ \\
    \midrule
    Euler-Bernoulli beam & \multirow{3}{*}{\makecell{Fixed ends \\ ($y=0$)}}
      & \makecell{$y(x,0)=0$ \\ $\dot{y}(x,0)=0$} & $\Delta x=0.03$ & 0.02\\
    \cmidrule{1-1}\cmidrule{3-5}
    Heat equation & & \multirow{2}{*}{$y(x,0)=0$} & $\Delta x=0.08$ & 0.01\\
    \cmidrule{1-1}\cmidrule{4-5}
    Reaction-diffusion & & & $\Delta x=0.08$ & 0.02\\
    \midrule
    Brusselator & Neumann
      & \makecell{$u(x,0)=0.5$ \\ $v(x,0)=1$} & $\Delta x=0.03$&  0.02 \\
    \midrule
    Burgers & Periodic & Vary & $\Delta x=\pi/25$ & 0.02 \\
    \midrule
    Navier-Stokes & Periodic & Vary & $\Delta x=\pi/25$ & 0.01 \\
    \bottomrule
  \end{tabular}
  \caption{Detailed configurations of PDEs}
  \label{tab: pde_configurations}
\end{table}

\section{Analysis of Transient and Steady-state Error Dynamics}

In this section, we present a comparative analysis of the empirical properties of LNO and LFNO, focusing on their error stability across different dynamical regimes. To evaluate long-term predictive performance, the time horizon is partitioned into the transient state ($t < 500$) and the steady state ($t \geq 500$). This threshold of $t=500$ is determined based on the physical settling time of the governing equations and the observed stabilization of error profiles. In the following figures in \cref{fig: mae_analysis}, results for LNO and LFNO are denoted in green and magenta, respectively. To clarify long-term trends, a moving average with a window size of $100$ steps is applied. 

Across all test cases, LNO exhibits significant fluctuations in error as time progresses. In contrast, LFNO maintains a remarkably narrow error range with minimal variance, demonstrating superior statistical stability during long-term integration. As the coefficients of the Duffing equation $c$ and the Lorenz system $\rho$ increase, the systems exhibit stronger chaotic characteristics, which typically hinders the learning process. However, LFNO consistently maintains a stable convergence pattern regardless of these variations, whereas the performance of LNO tends to degrade under increased complexity. These visual and numerical evidences confirm that LFNO provides substantially higher steady-state stability and predictive precision than the conventional LNO, particularly in chaotic dynamics.

\begin{figure}[ht]
    \resizebox{\columnwidth}{!}{%
    \begin{tabular}{cc}
    \centering
    \subcaptionbox{Duffing $c = 0$\label{fig: mae_duffingc0}}{
        \includegraphics[width=0.49\textwidth]{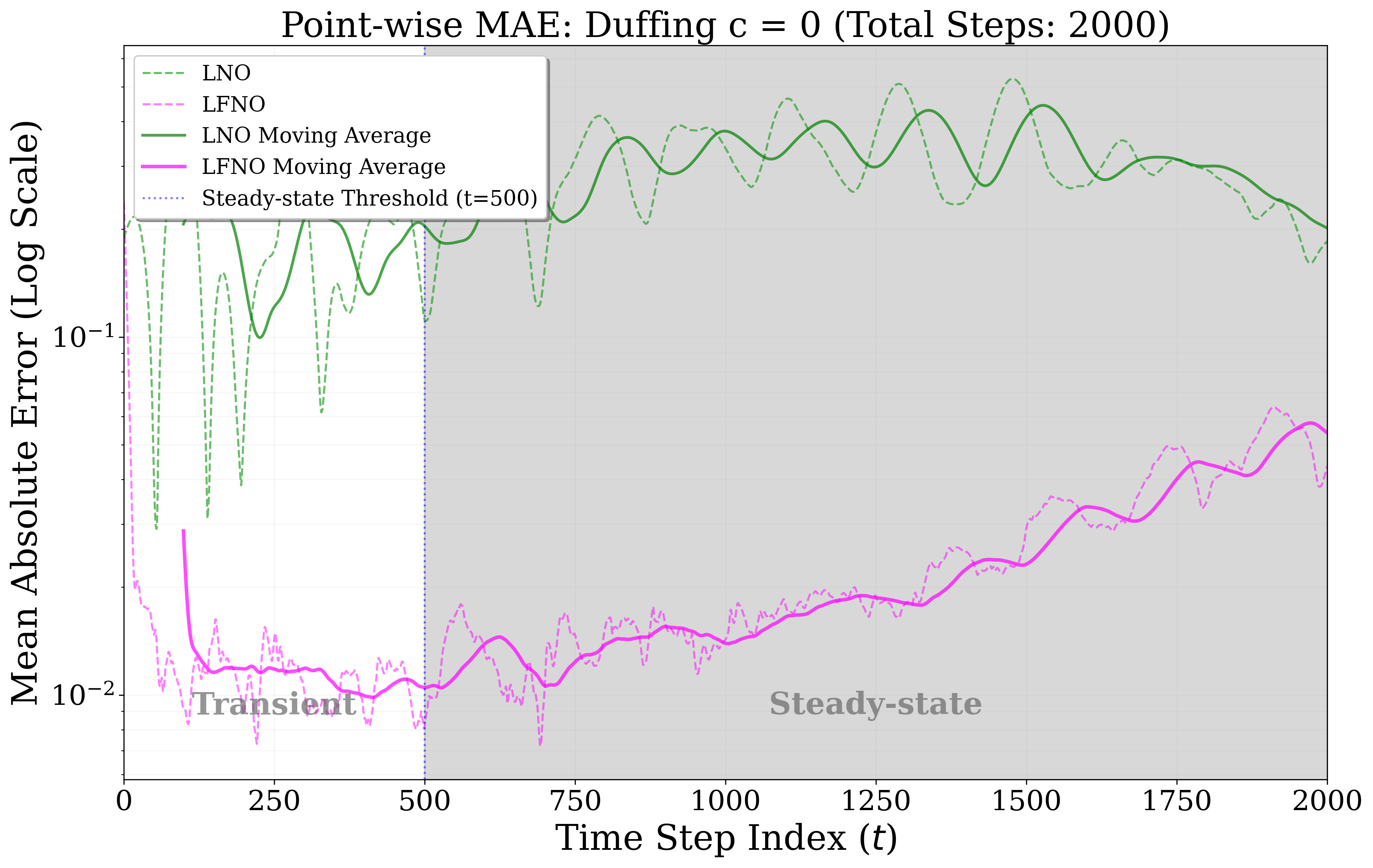}
    } & 
    \subcaptionbox{Duffing $c = 0.5$\label{fig: mae_duffingc05}}{
        \includegraphics[width=0.49\textwidth]{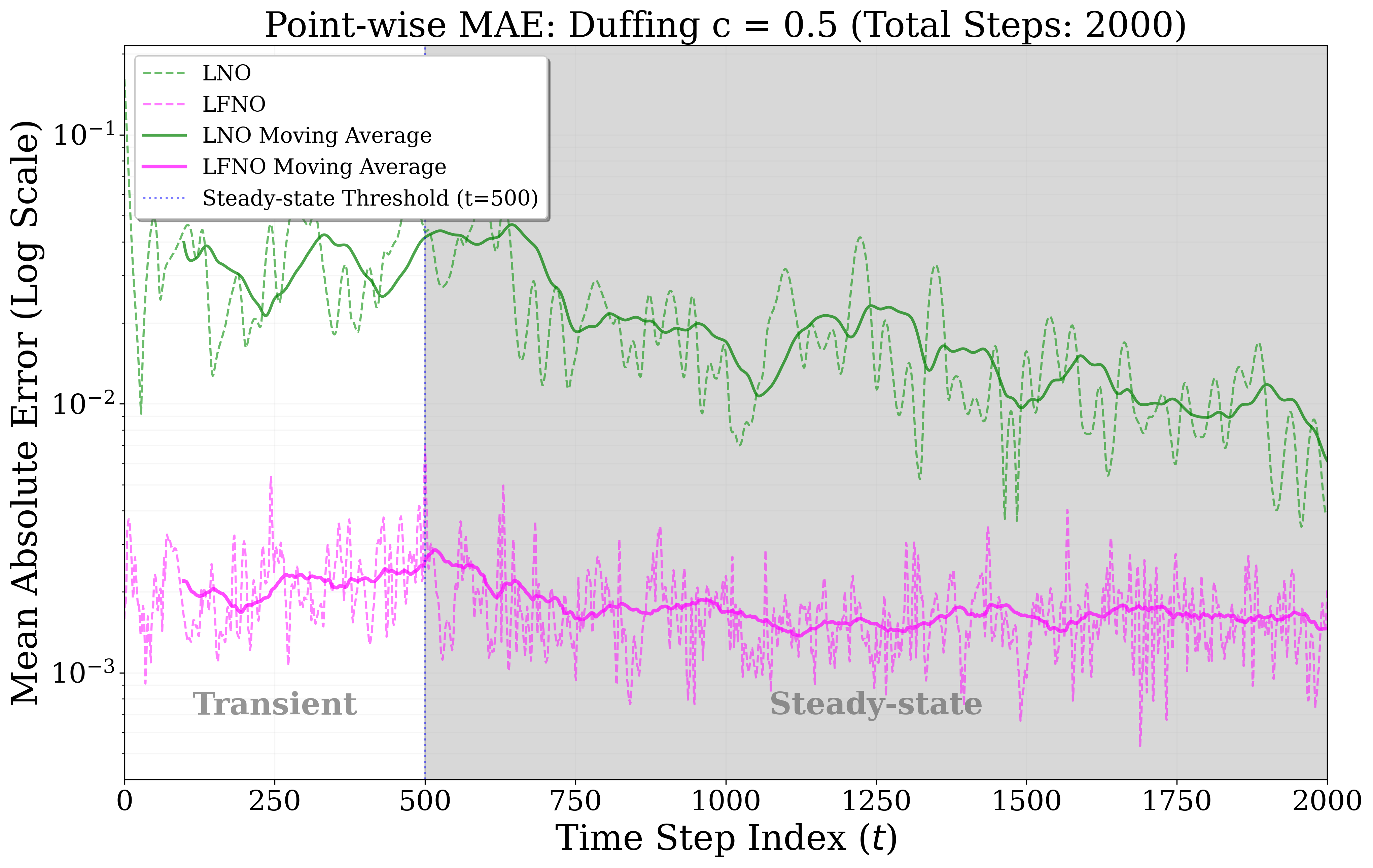}
    }\\[0.5em]
    \subcaptionbox{Lorenz $\rho = 5$\label{fig: mae_lorenz_rho5}}{
        \includegraphics[width=0.49\textwidth]{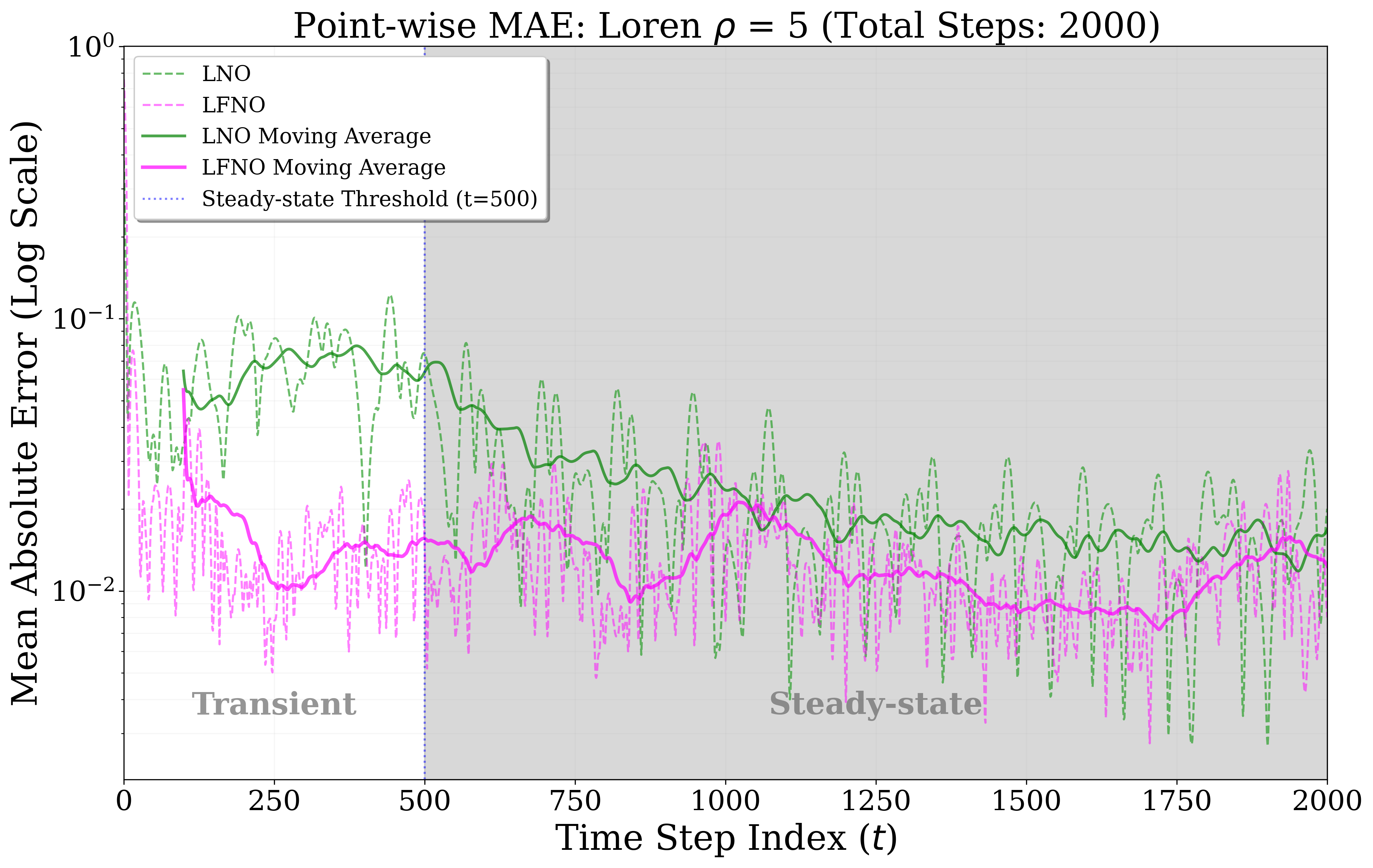}
    } &
    \subcaptionbox{Lorenz $\rho = 10$\label{fig: mae_lorenz_rho10}}{
        \includegraphics[width=0.49\textwidth]{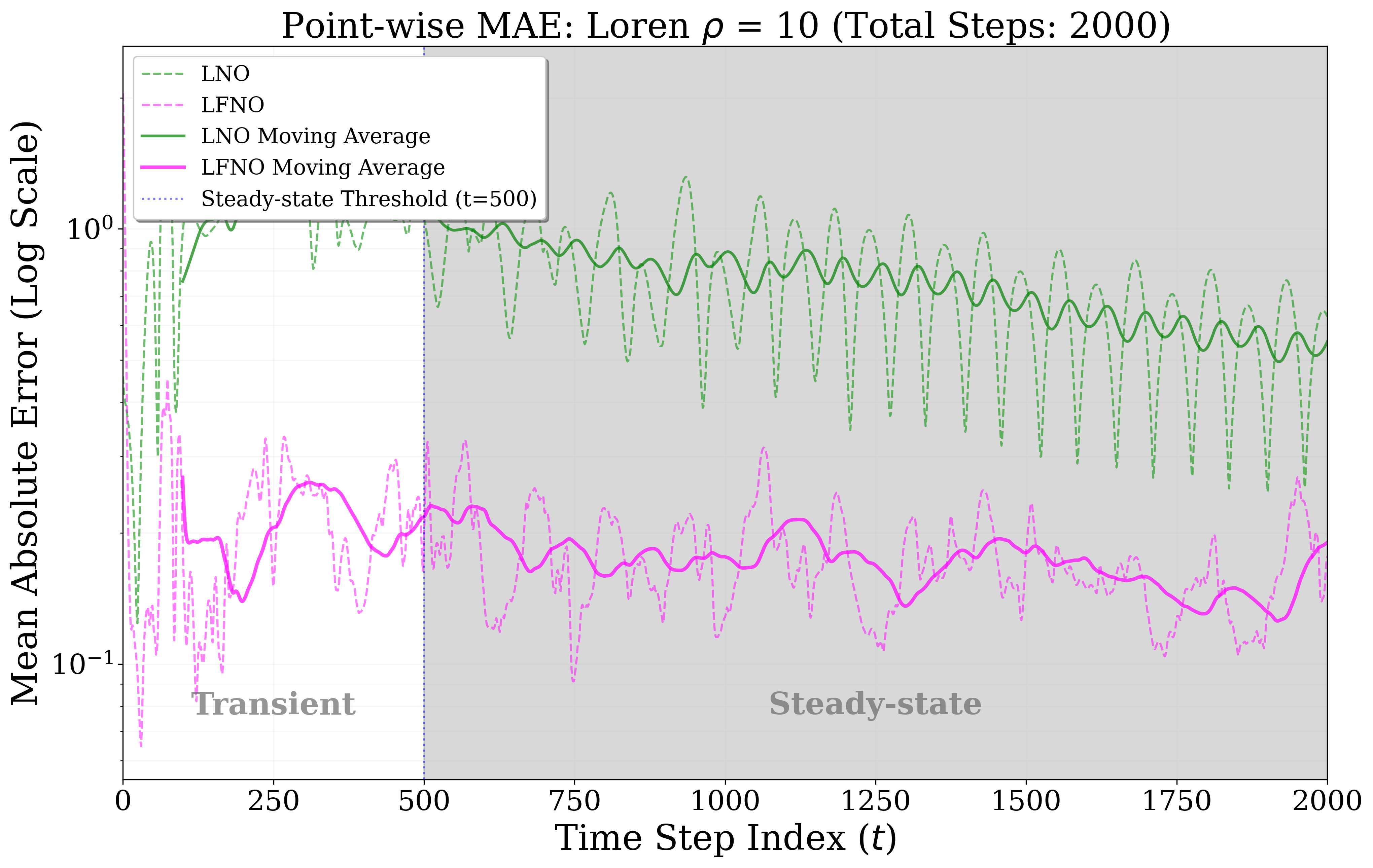}
    }\\[0.5em]
    \subcaptionbox{Pendulum $c = 0.5$\label{fig: mae_pendulumc05}}{
        \includegraphics[width=0.49\textwidth]{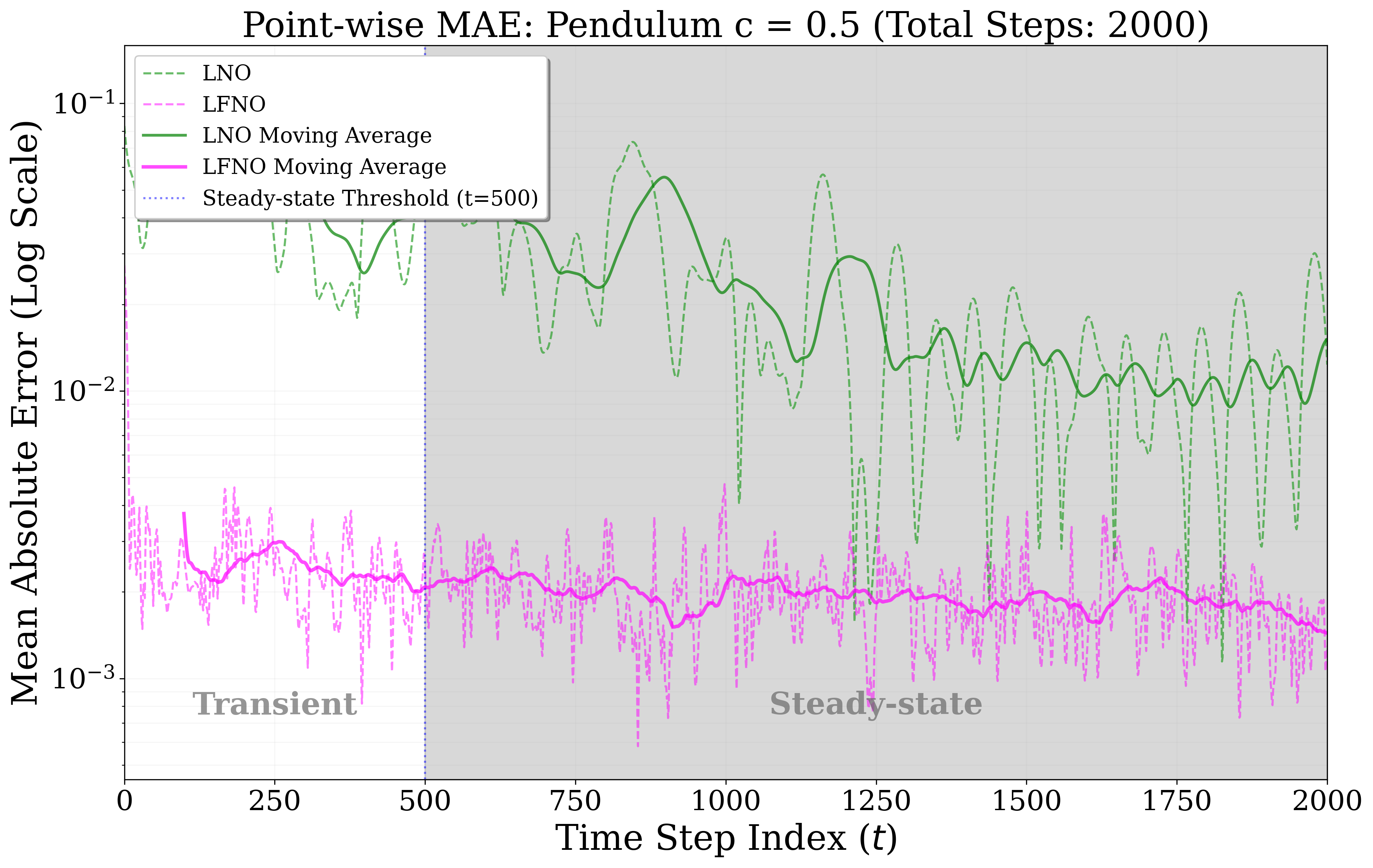}
    }
    \end{tabular}
    }
    \caption{Comparison of point-wise Mean Absolute Error (MAE) between LNO and LFNO across ODE systems. The time horizon is partitioned into the transient state ($t < 500$), and the steady-state ($t > 500$). (a), (b) Duffing ($c = 0, 0.5$), (c), (d) Lorenz ($\rho = 5, 10$), and (e) Pendulum ($c = 0.5$). The green lines represent the LNO, while the magenta lines denote the LFNO.}
    \label{fig: mae_analysis}
\end{figure}

\section{Dataset Samples}\label{sec: dataset_samples}

\cref{tab: dataset_samples_ODEs} and \cref{tab: dataset_samples_PDEs} show the visualizations of the datasets for each task. The datasets themselves do not have a split, as we used a split of K-folds or a random split on the fly to shuffle and split the test and validation samples.

{
    \newcommand{\imhere}[1]{
        \includegraphics[width=40mm]{images/dataset/#1.png}
    }
    \newcommand{\imheret}[1]{
        \includegraphics[width=35mm]{images/dataset/#1.png}
    }
    \newcolumntype{C}{>{\centering\arraybackslash}m{40mm}}
    \begin{table}[t]
        \centering
        \begin{tabular}{cCC}
            \toprule
            Task & Signal & Grount Truth (GT) \\
            \midrule
            \makecell{Duffing \\ c0} & 
            \imhere{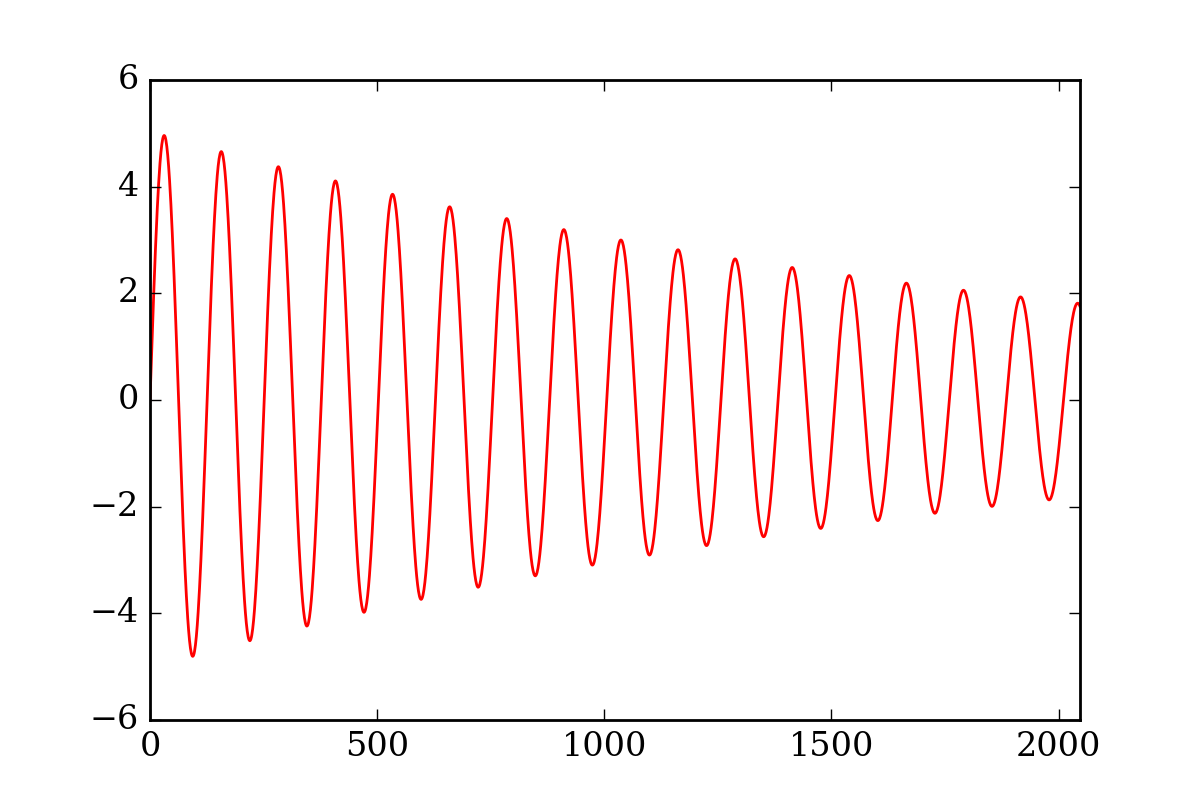} &
            \imhere{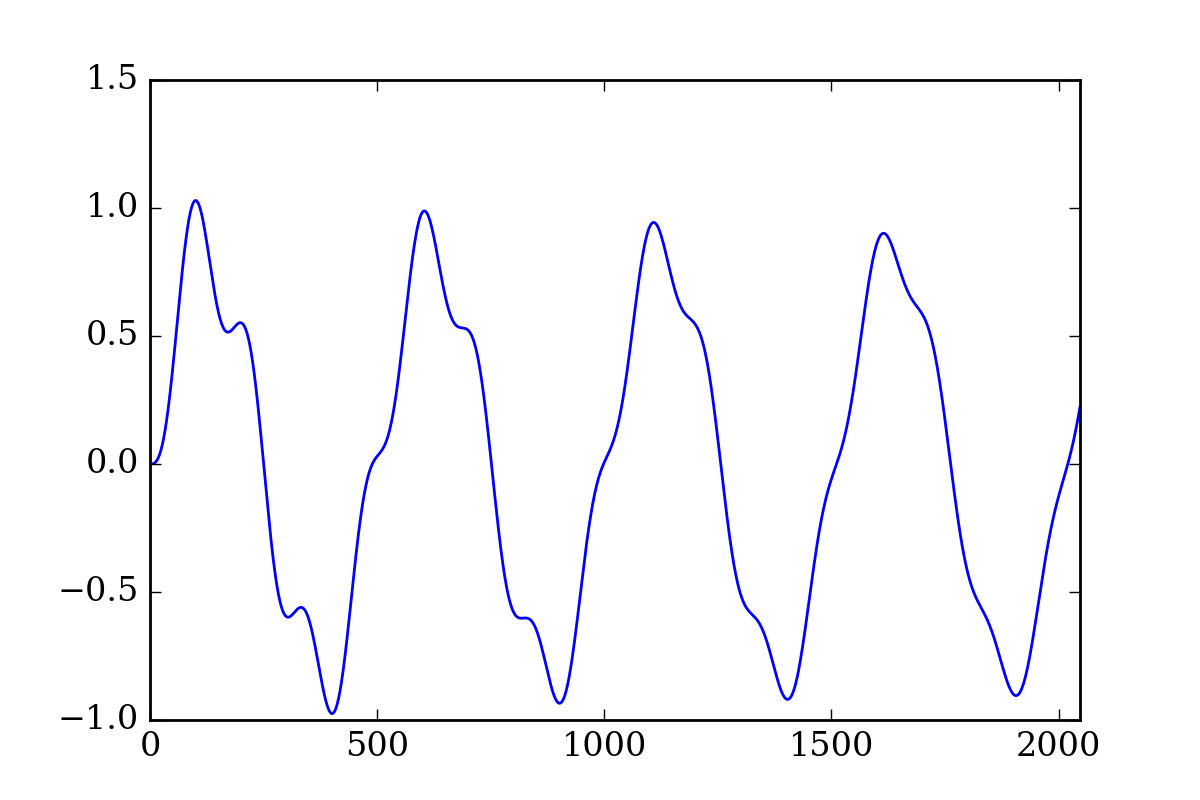} \\
            \makecell{Duffing \\ c05} & 
            \imhere{1d_signal} & 
            \imhere{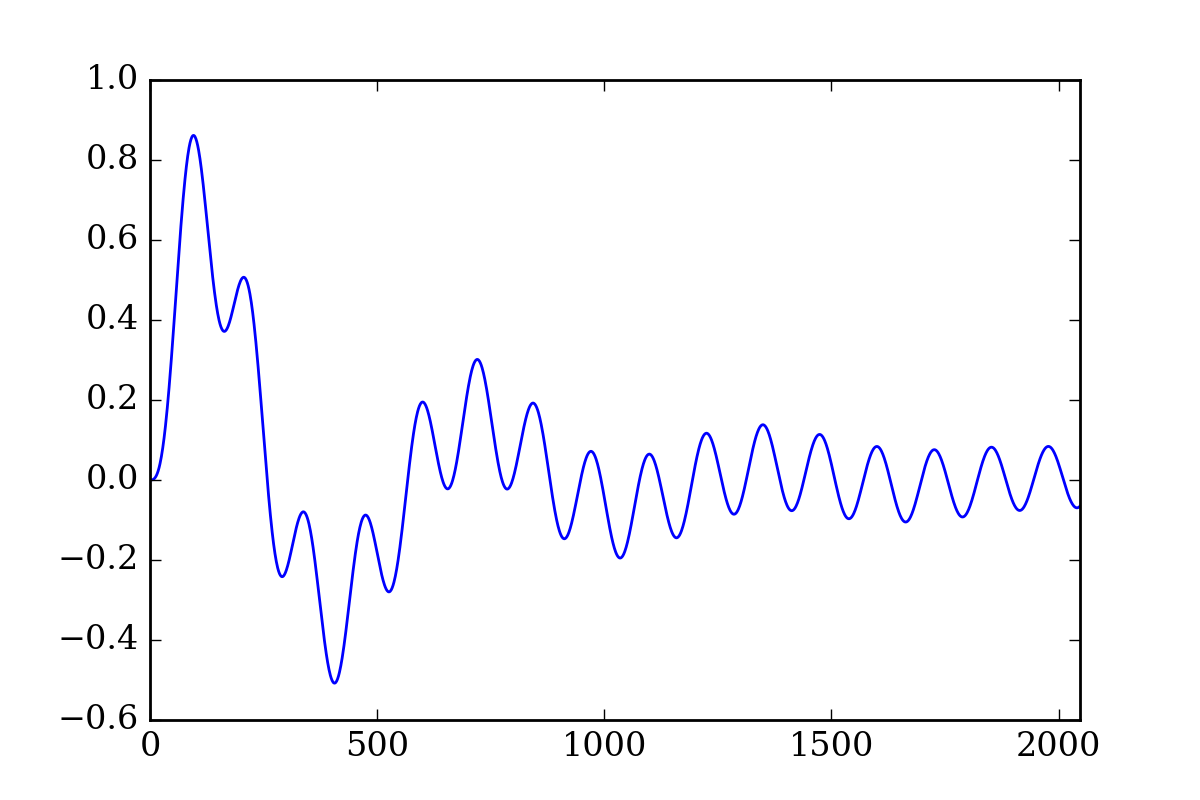} \\
            \makecell{Lorenz \\ rho5} &
            \imhere{1d_signal} &
            \imhere{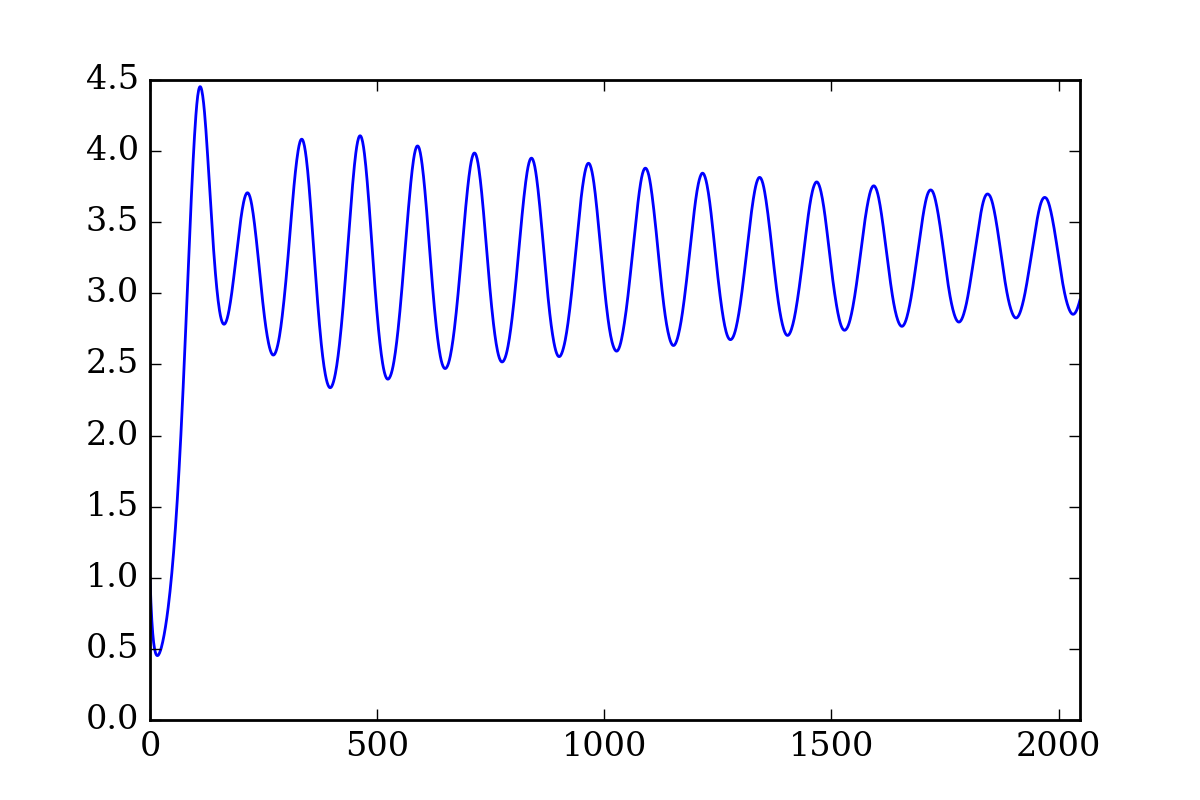} \\
            \makecell{Lorenz \\ rho10} &
            \imhere{1d_signal} &
            \imhere{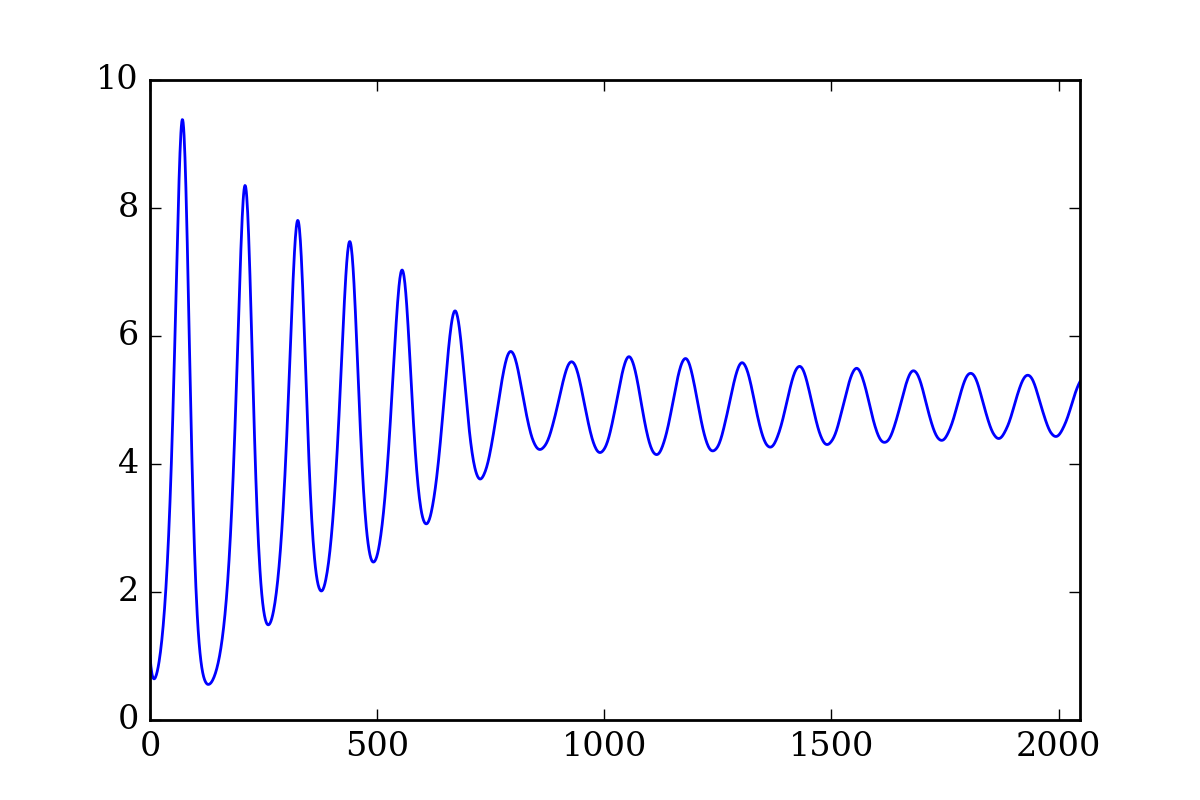} \\
            \makecell{Pendulum \\ c05} &
            \imhere{1d_signal} & 
            \imhere{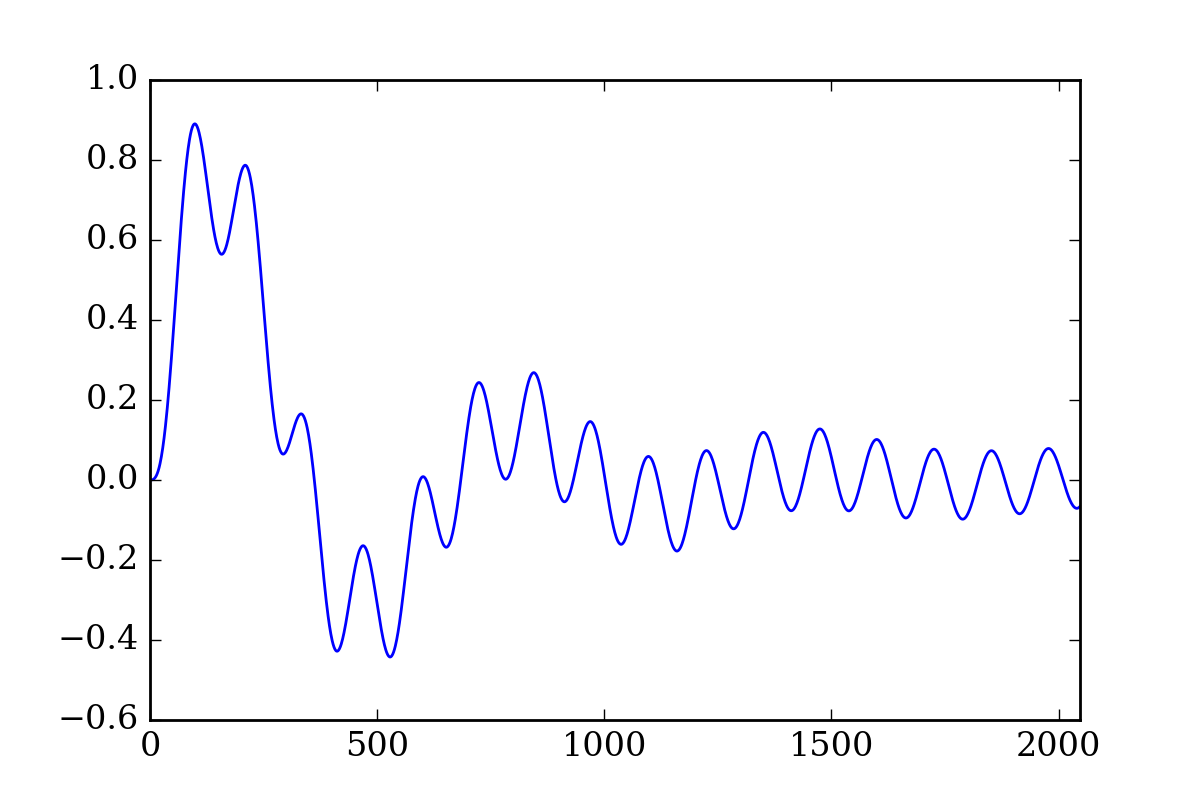}\\
            \bottomrule
        \end{tabular}
        \vskip 0.1in
        \caption{Visualization of ODE samples in our dataset}
        \label{tab: dataset_samples_ODEs}
    \end{table}
}

{
    \newcommand{\imhere}[1]{
        \includegraphics[width=40mm]{images/dataset/#1.png}
    }
    \newcommand{\imheret}[1]{
        \includegraphics[width=35mm]{images/dataset/#1.png}
    }
    \newcolumntype{C}{>{\centering\arraybackslash}m{40mm}}
    \begin{table}[t]
        \centering
        \begin{tabular}{cCC}
            \toprule
            Task & Signal & Grount Truth (GT) \\
            \midrule
            Beam & 
            \imhere{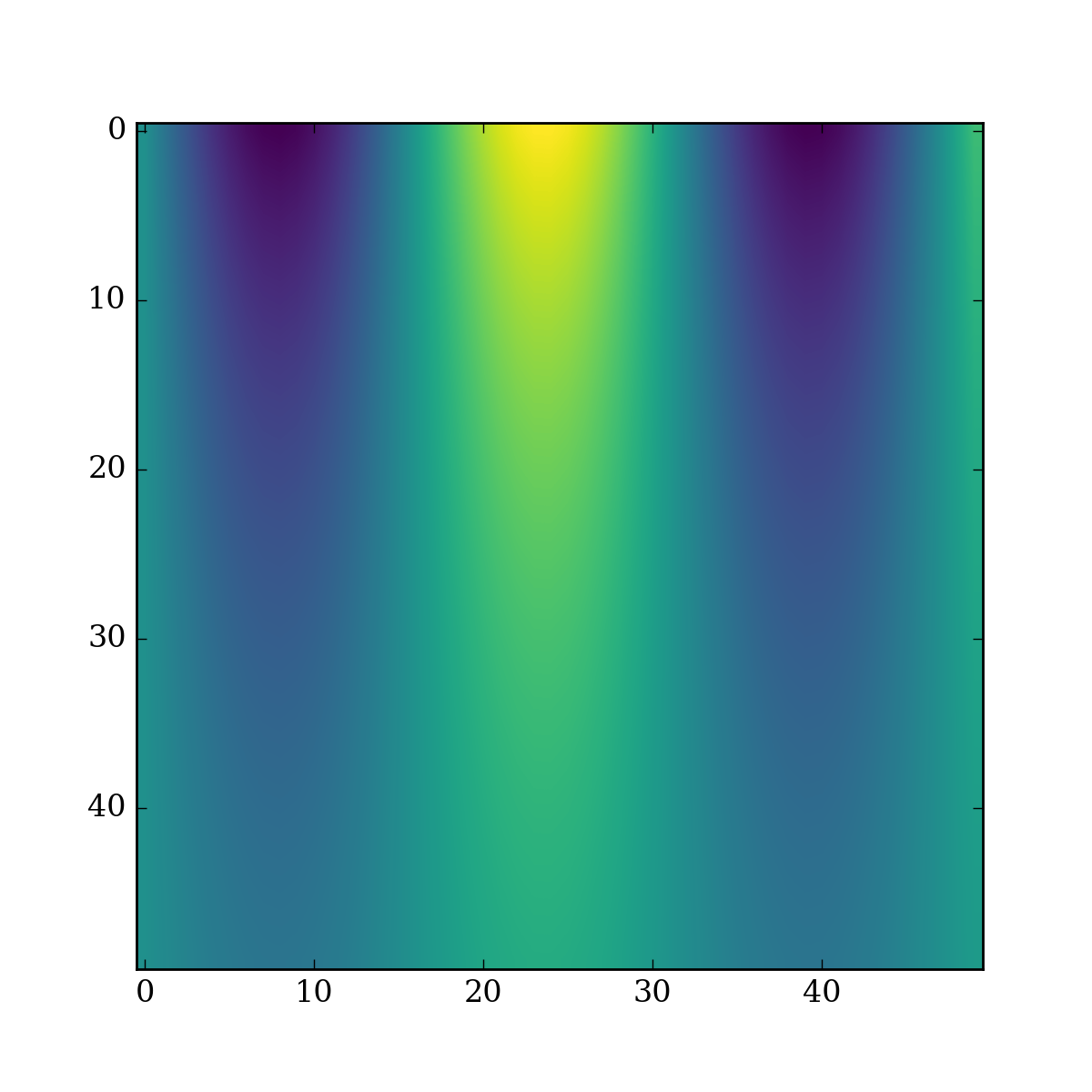} & 
            \imhere{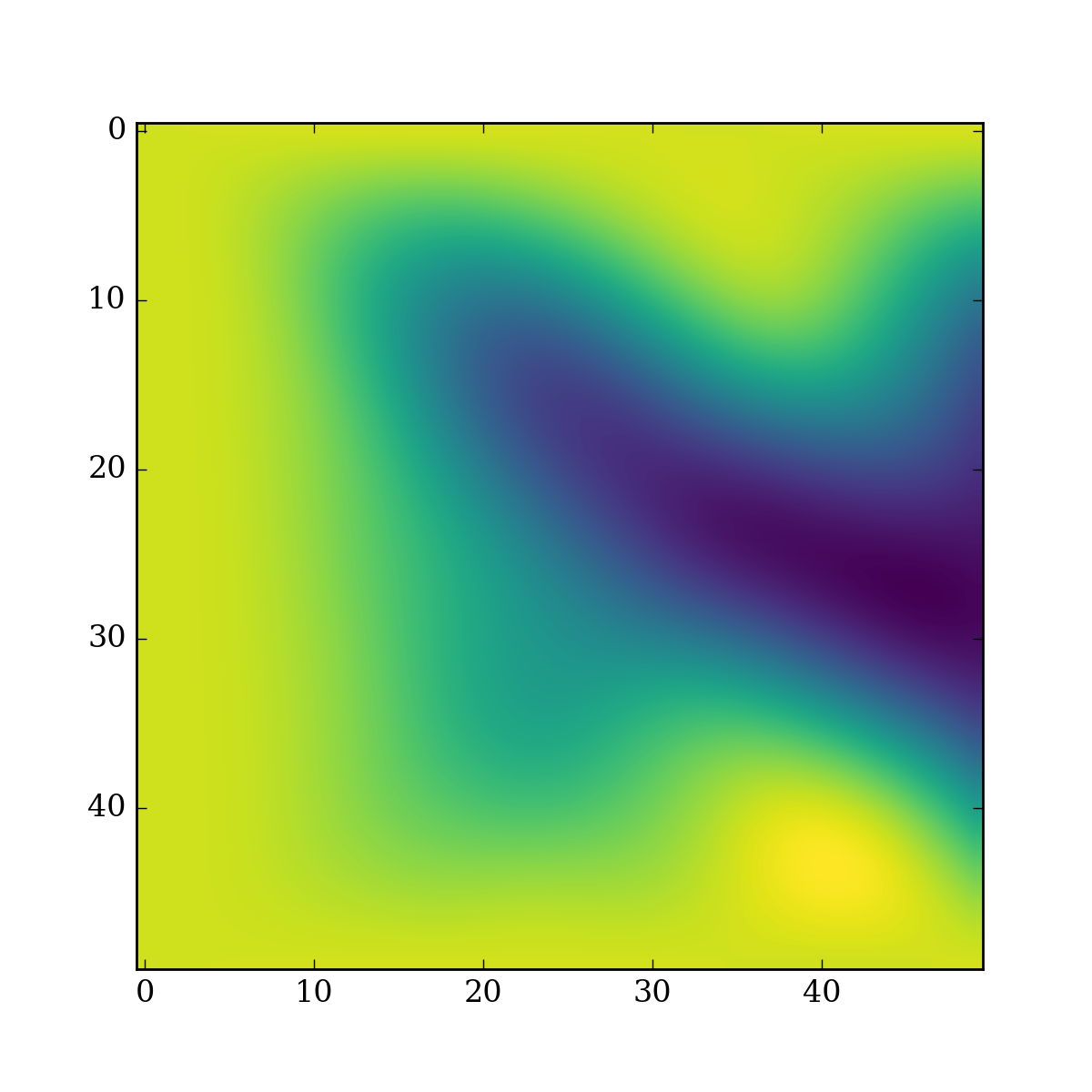} \\
            Heat & 
            \imhere{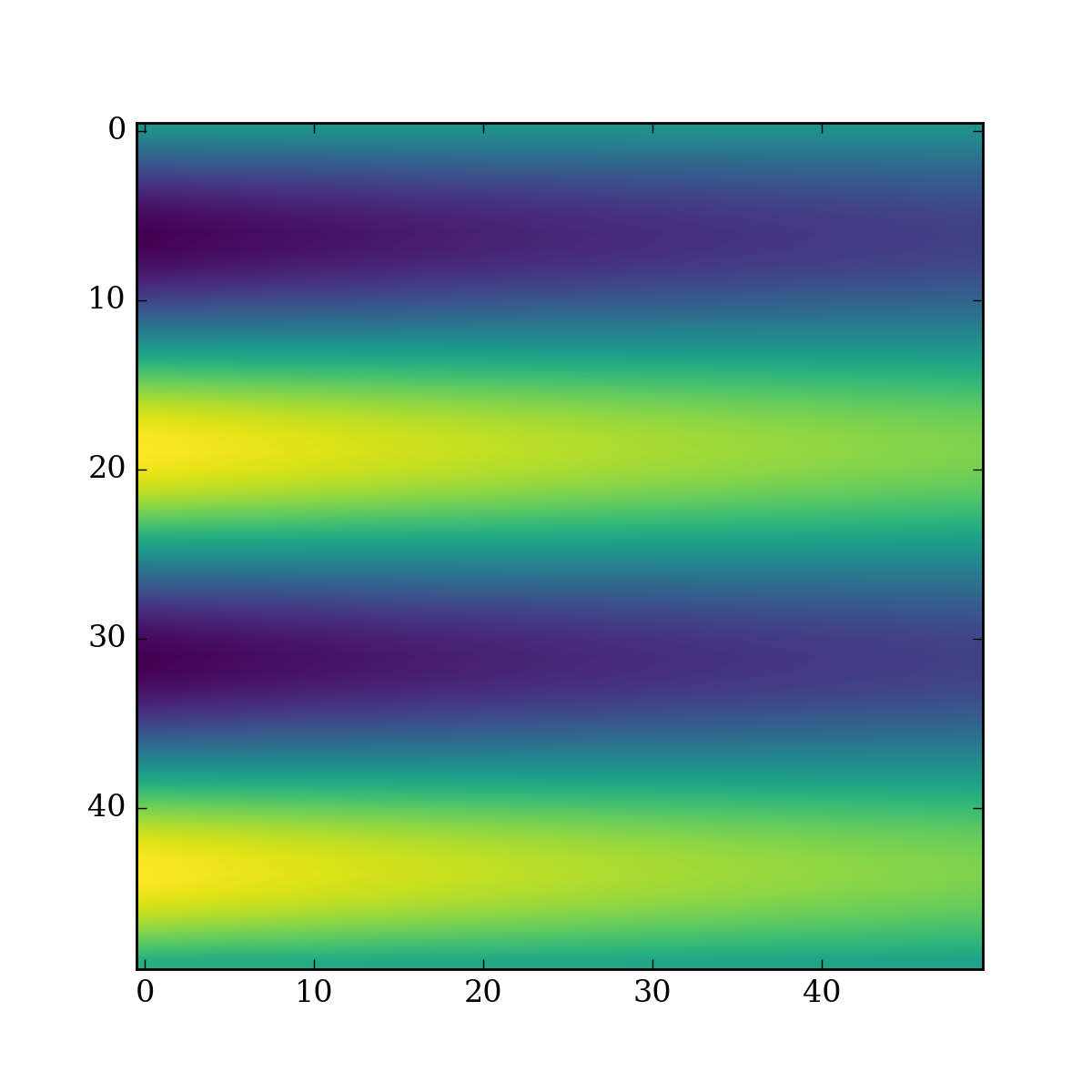} &
            \imhere{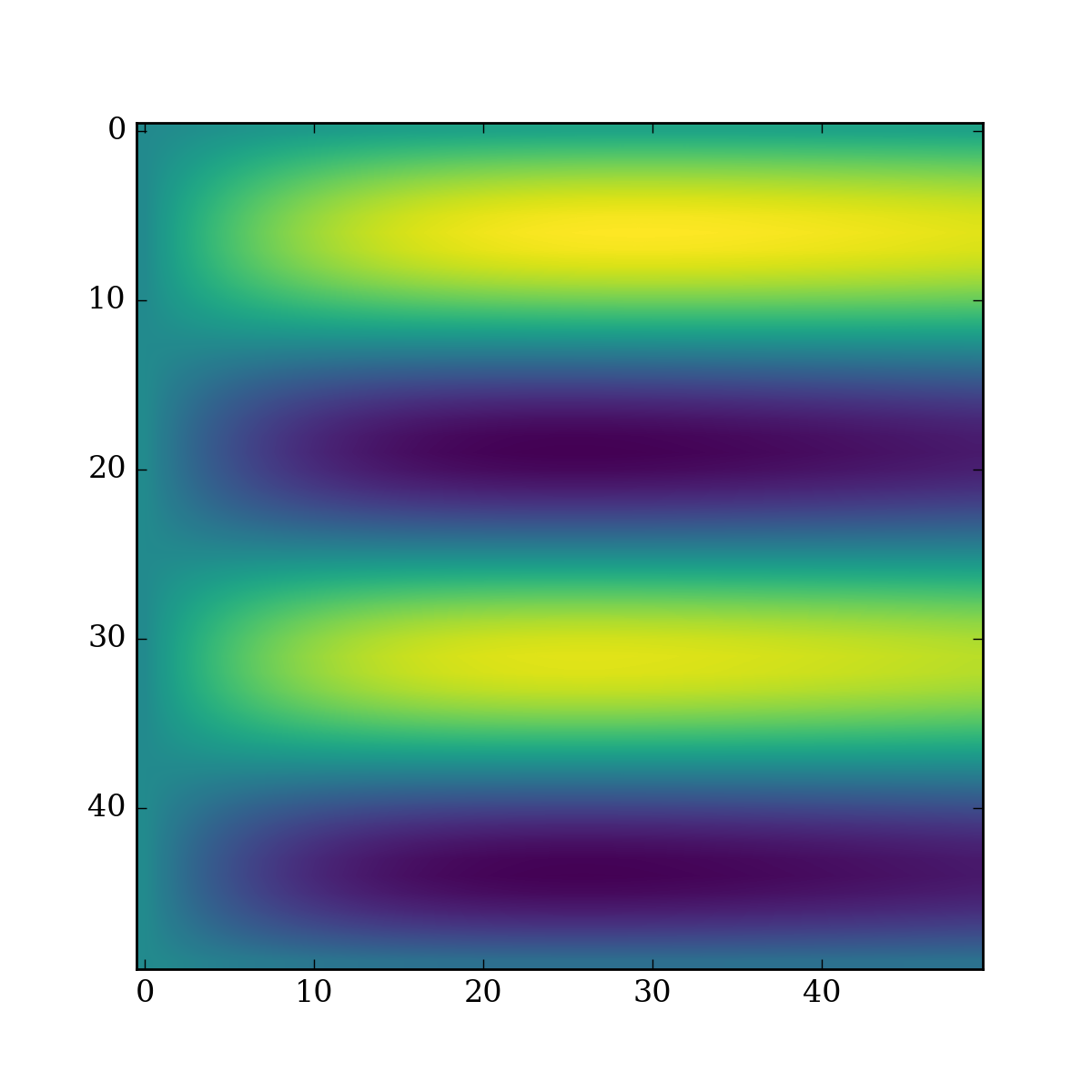} \\
            Reaction-diffusion &
            \imhere{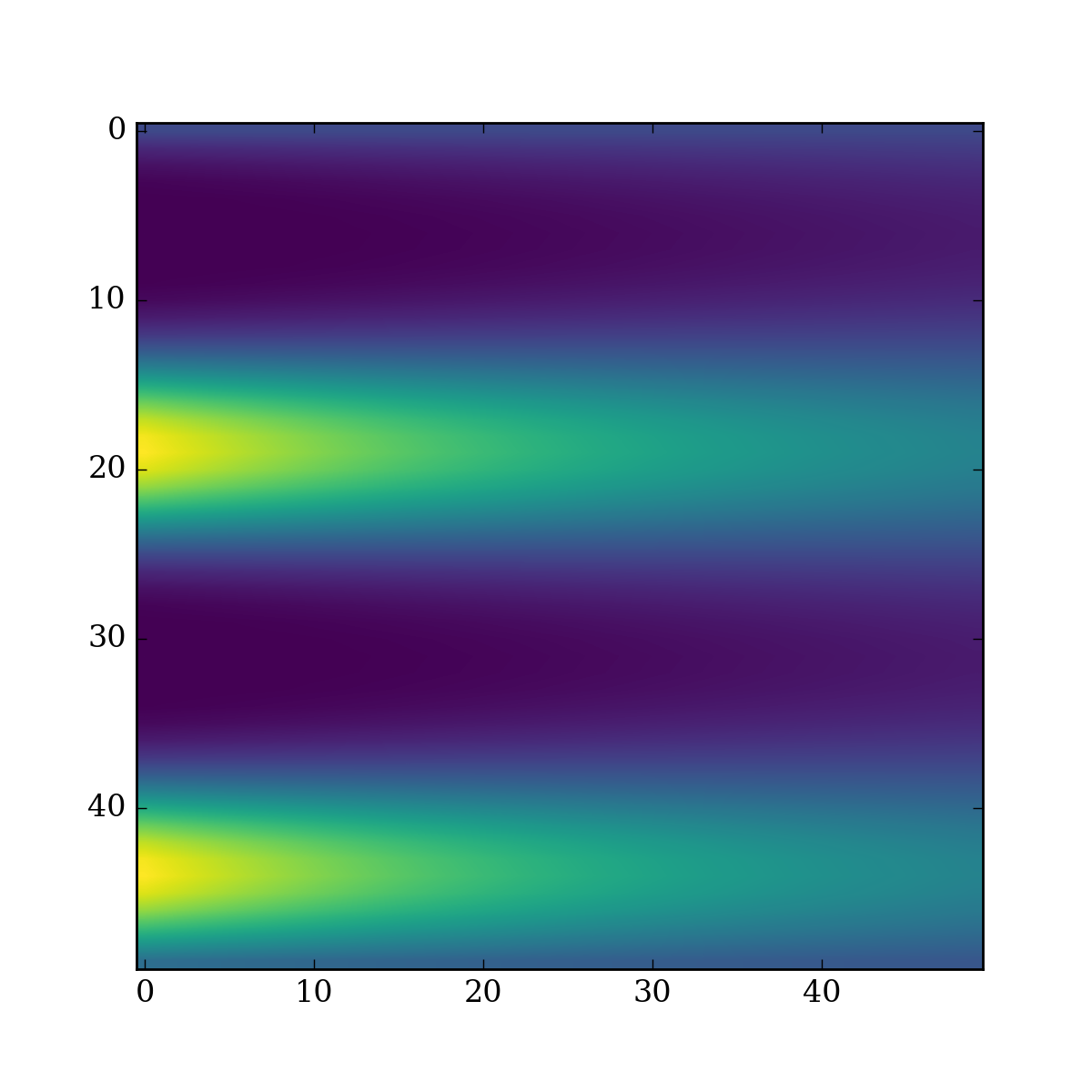} &
            \imhere{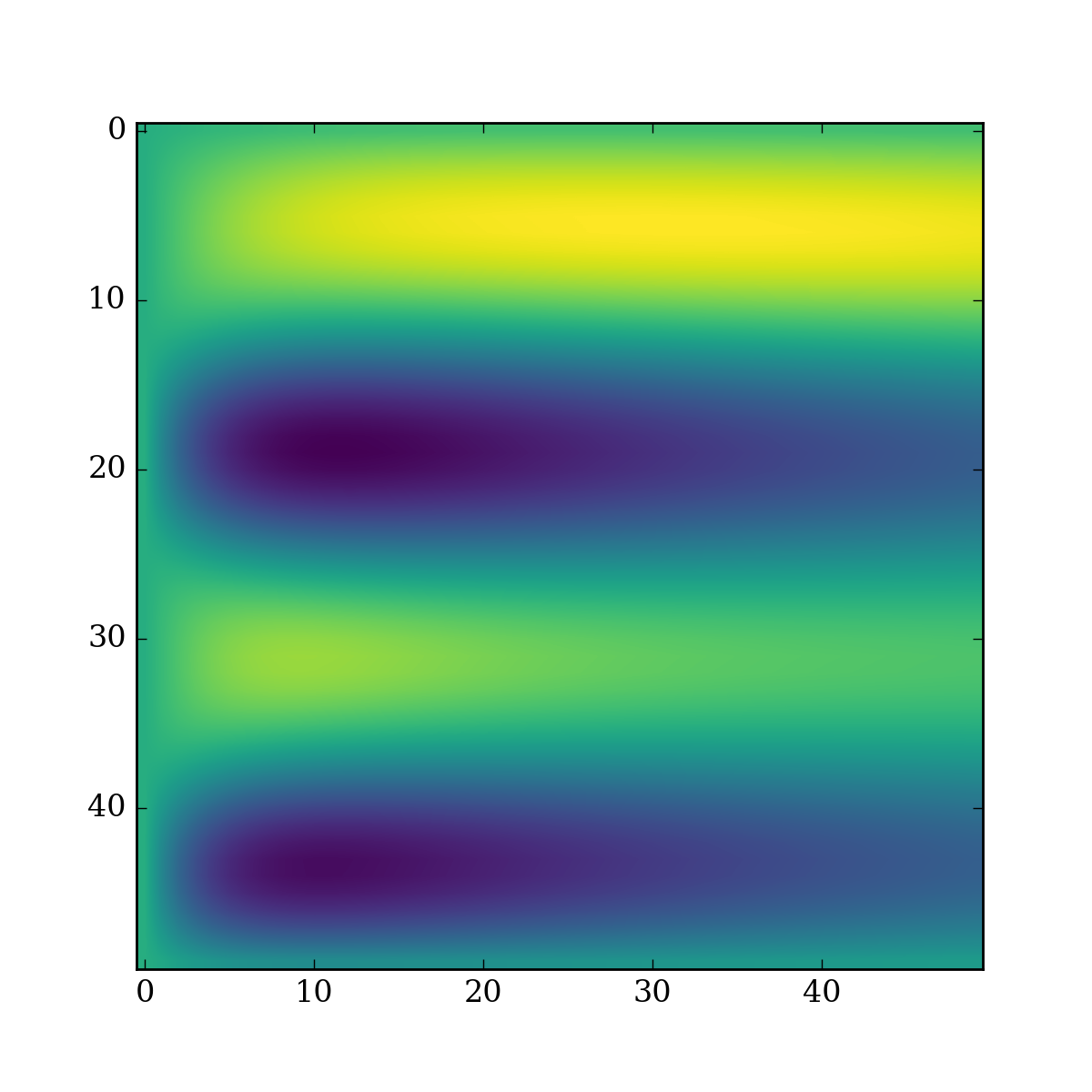} \\
            Brusselator & 
            \imheret{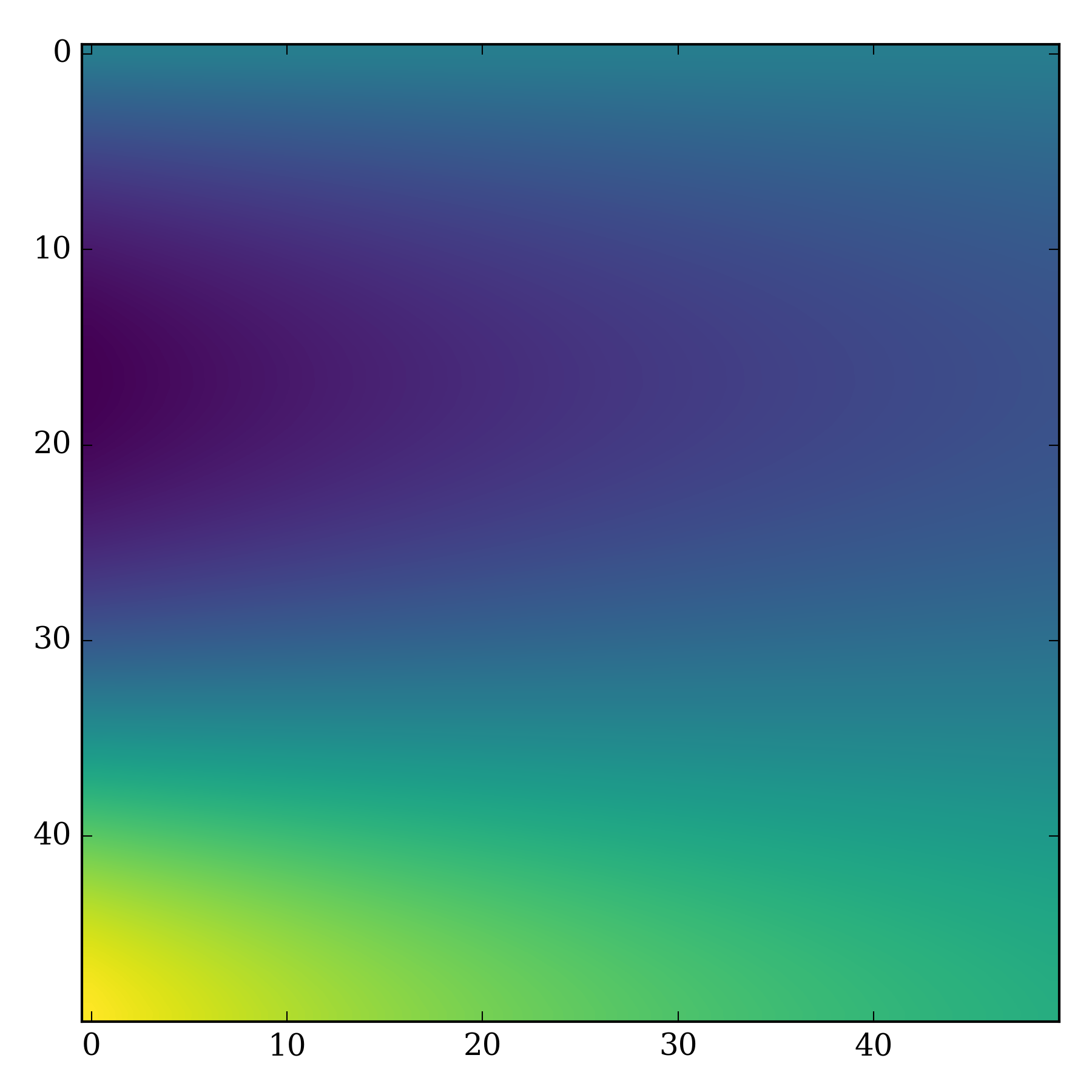} &  
            \imheret{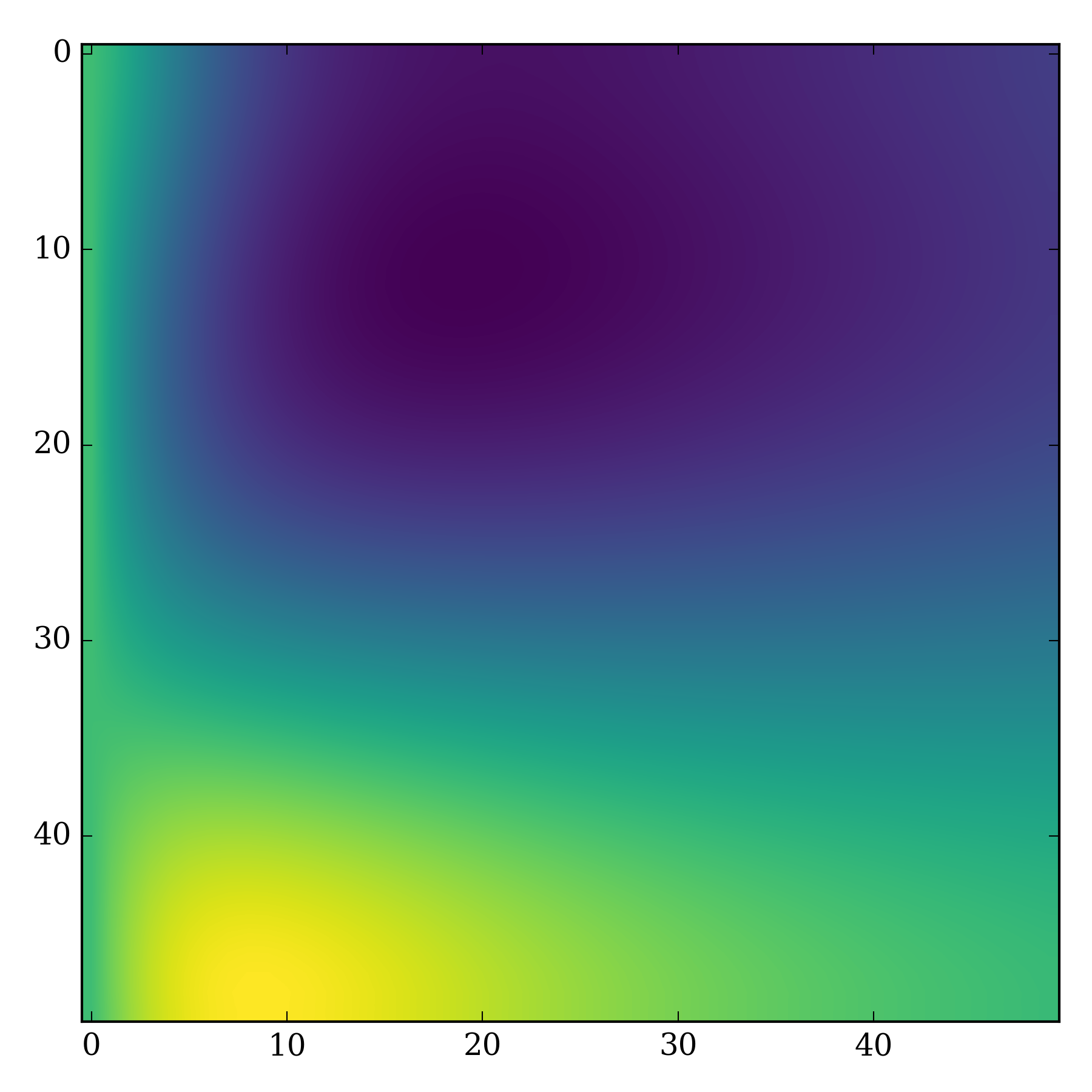} \\
            Burgers & 
            \imheret{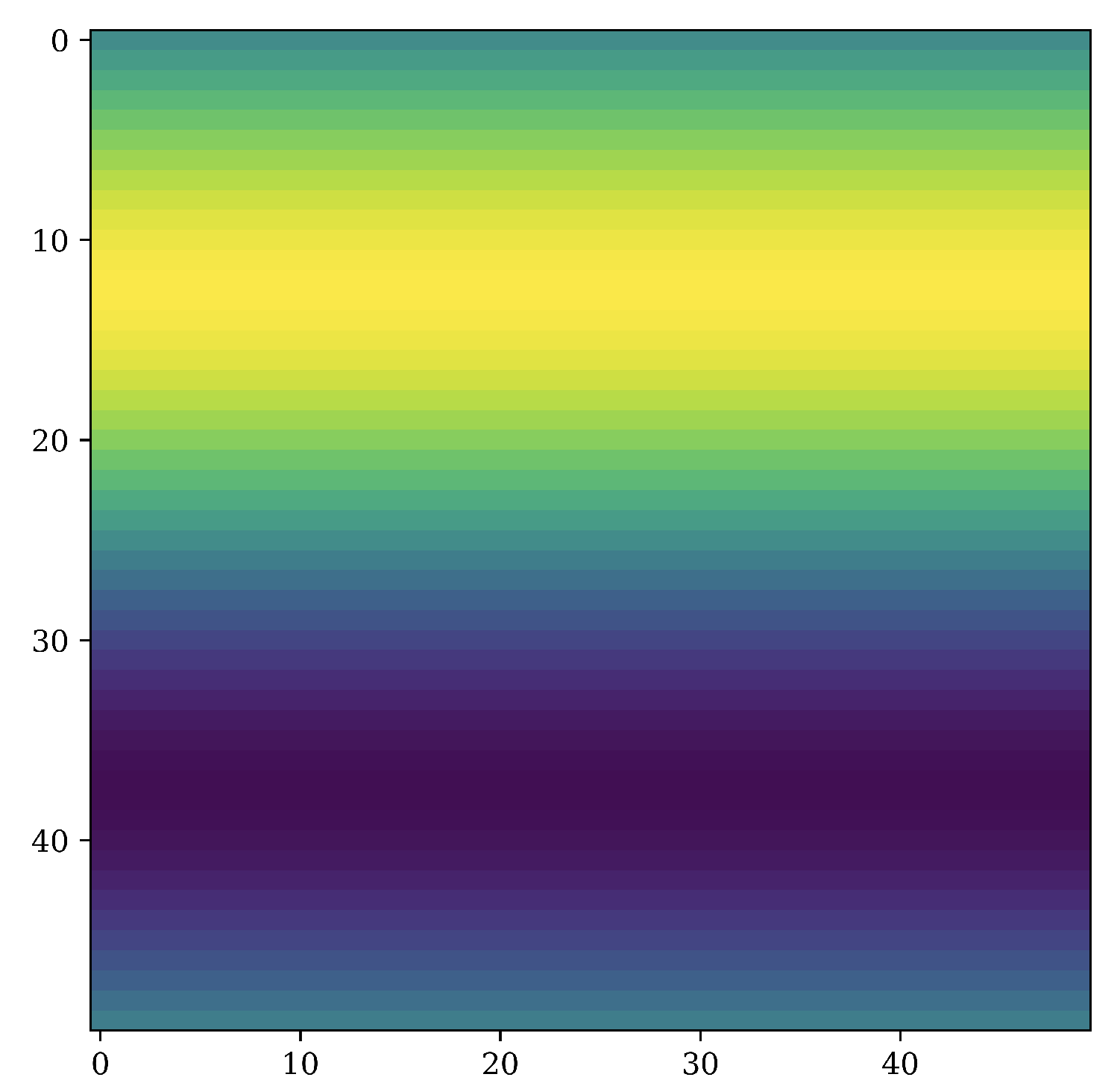} &
            \imheret{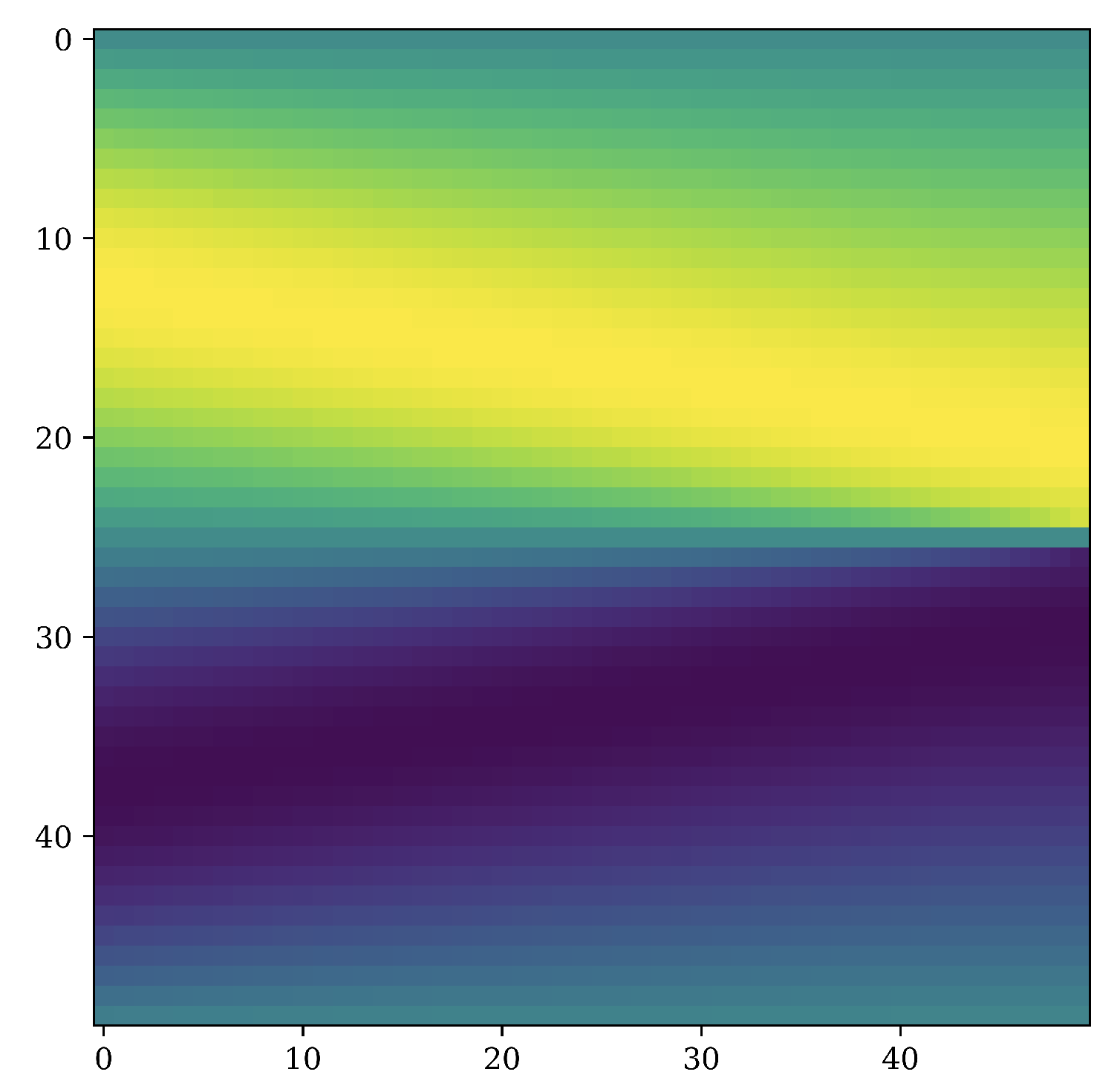} \\
            \bottomrule
        \end{tabular}
        \vskip 0.1in
        \caption{Visualization of PDE samples in our dataset}
        \label{tab: dataset_samples_PDEs}
    \end{table}
}

\section{PDEs Qualitative Figures}\label{sec: figure_PDEs}

\begin{figure}[ht]
    \centering
    \begin{subfigure}[b]{0.8\textwidth}
        \centering
        \includegraphics[height=4cm]{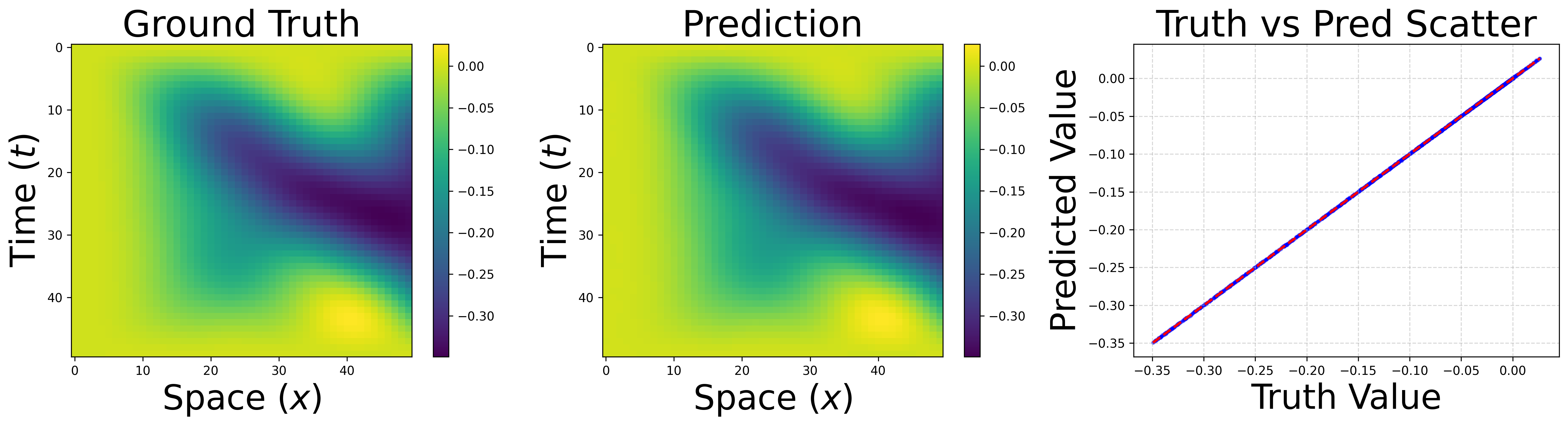}
        \caption{Qualitative Results and Scatter of the Euler-Bernoulli beam in FNO}
        \label{fig: FNO_Beam_qualitative}
    \end{subfigure}
    \vspace{0.5em}
    \begin{subfigure}[b]{0.8\textwidth}
        \centering
        \includegraphics[height=4cm]{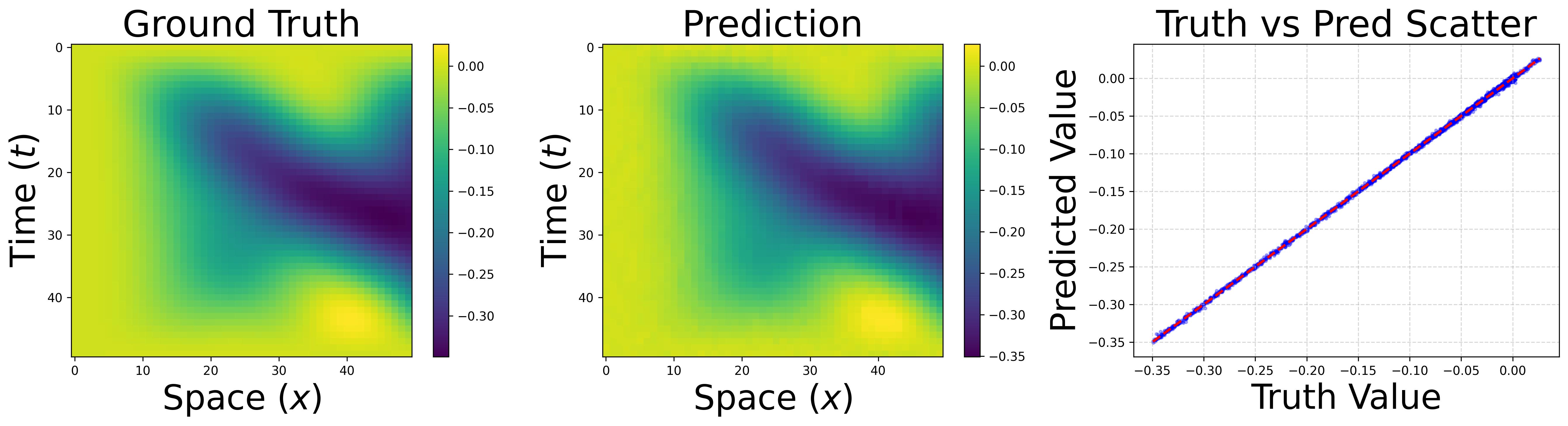}
        \caption{Qualitative Results and Scatter of the Euler-Bernoulli beam in LFNO}
        \label{fig: LFNO_Beam_qualitative}
    \end{subfigure}
    \vspace{0.5em}
    \begin{subfigure}{0.8\textwidth}
        \centering
        \includegraphics[height=4cm]{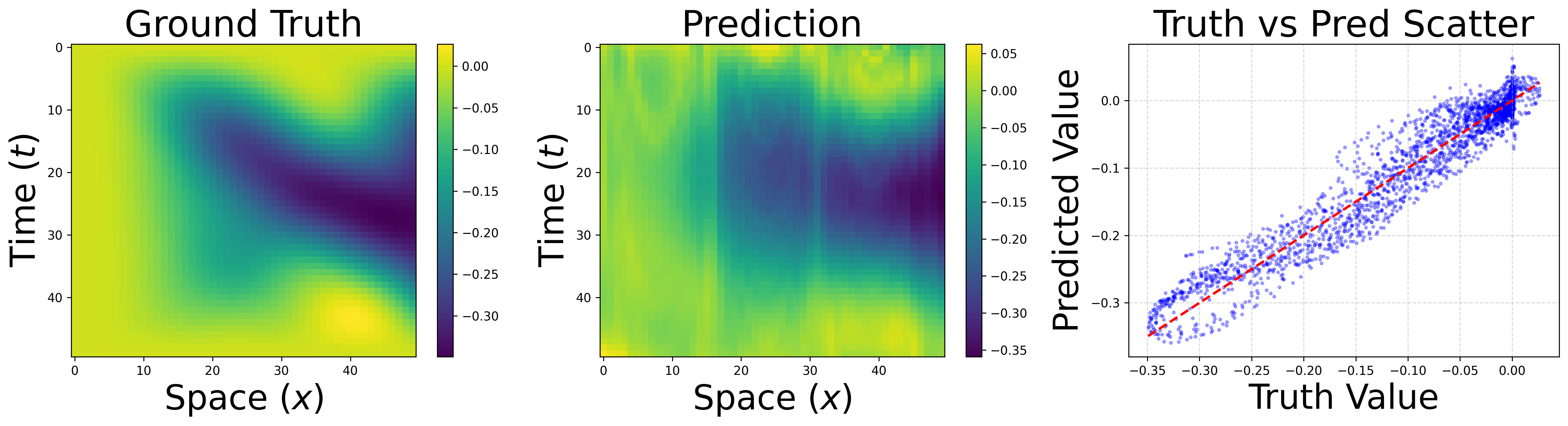}
        \caption{Qualitative Results and Scatter of the Euler-Bernoulli beam in LNO}
        \label{fig: LNO_Beam_qualitative}
    \end{subfigure}

    \caption{Euler-Bernoulli beam equation: Comparison of qualitative results computed by FNO, LFNO, and LNO. It represents the ground truth shown in the left plot, the prediction of each model in the center plot, and the (truth, pred) scatters are shown in the right plot.}
    \label{fig: Beam_qualitative_comparison}
\end{figure}

\begin{figure}[ht]
    \centering
    \begin{subfigure}[b]{0.8\textwidth}
        \centering
        \includegraphics[height=4cm]{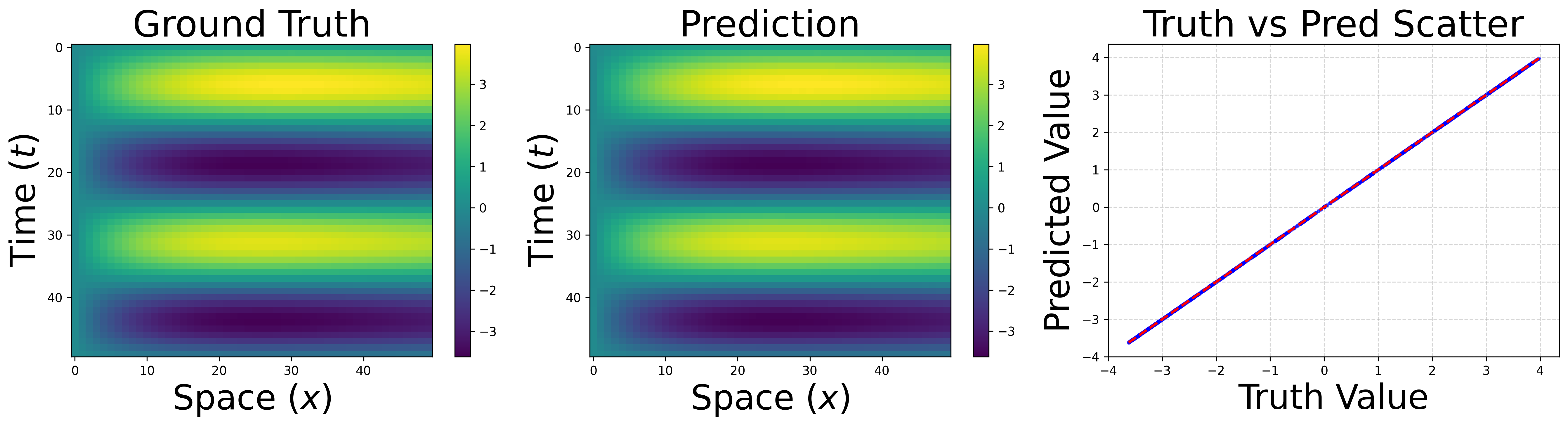}
        \caption{Qualitative Results and Scatter of the heat equation in FNO}
        \label{fig: FNO_Diffusion_qualitative}
    \end{subfigure}
    \vspace{0.5em}
    \begin{subfigure}[b]{0.8\textwidth}
        \centering
        \includegraphics[height=4cm]{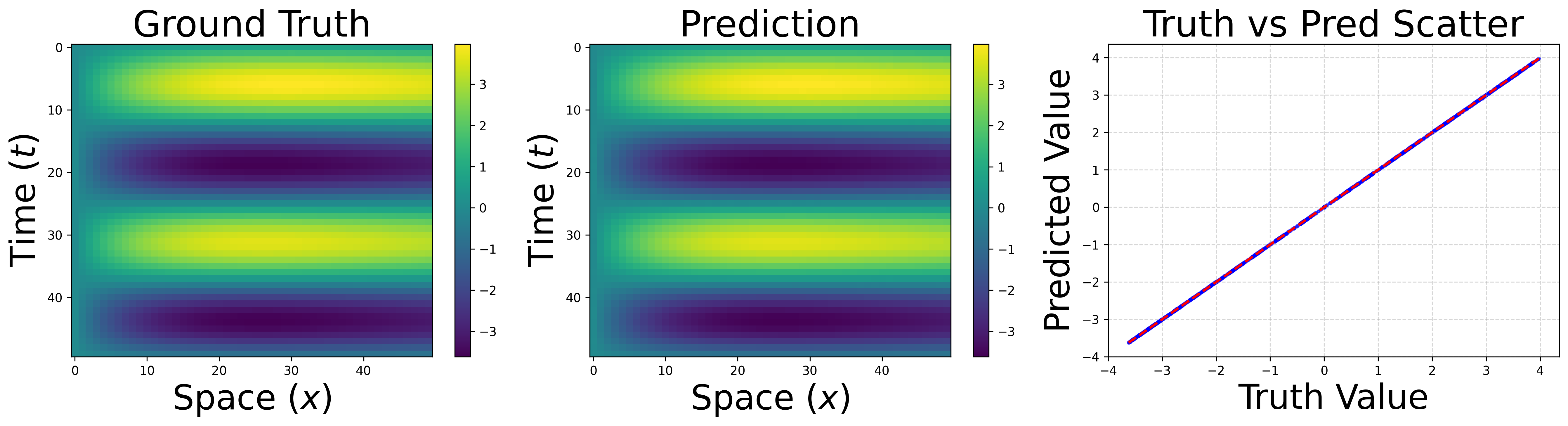}
        \caption{Qualitative Results and Scatter of the heat equation in LFNO}
        \label{fig: LFNO_Diffusion_qualitative}
    \end{subfigure}
    \vspace{0.5em}
    \begin{subfigure}{0.8\textwidth}
        \centering
        \includegraphics[height=4cm]{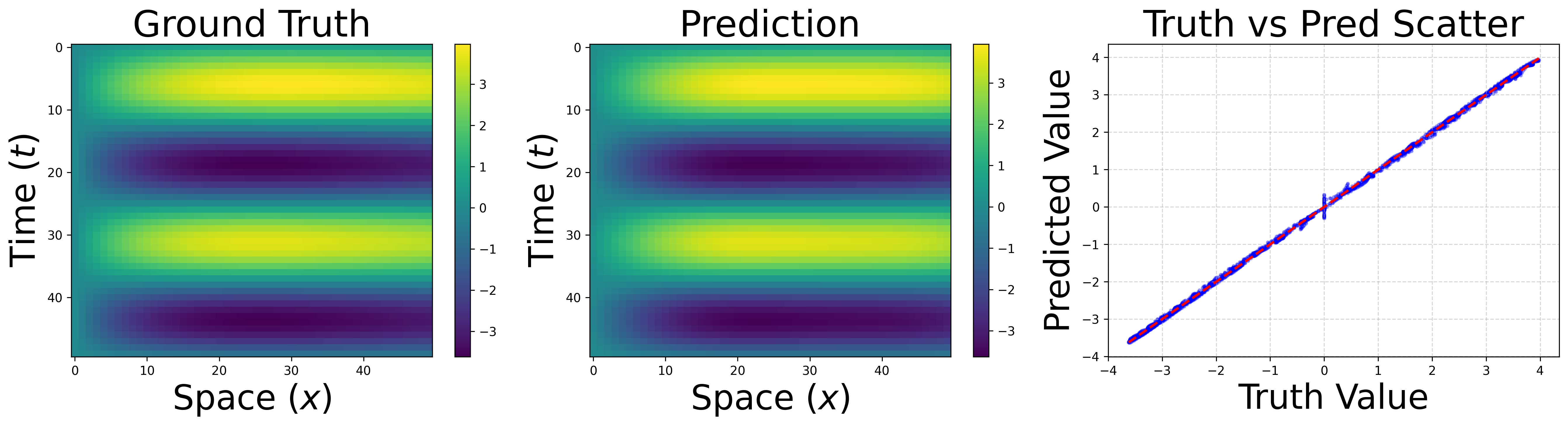}
        \caption{Qualitative Results and Scatter of the heat equation in LNO}
        \label{fig: LNO_Diffusion_qualitative}
    \end{subfigure}

    \caption{Heat equation: Comparison of qualitative results computed by FNO, LFNO, and LNO. It represents the ground truth shown in the left plot, the prediction of each model in the center plot, and the (truth, pred) scatters are shown in the right plot.}
    \label{fig: Diffusion_qualitative_comparison}
\end{figure}

\begin{figure}[ht]
    \centering
    \begin{subfigure}[b]{0.8\textwidth}
        \centering
        \includegraphics[height=4cm]{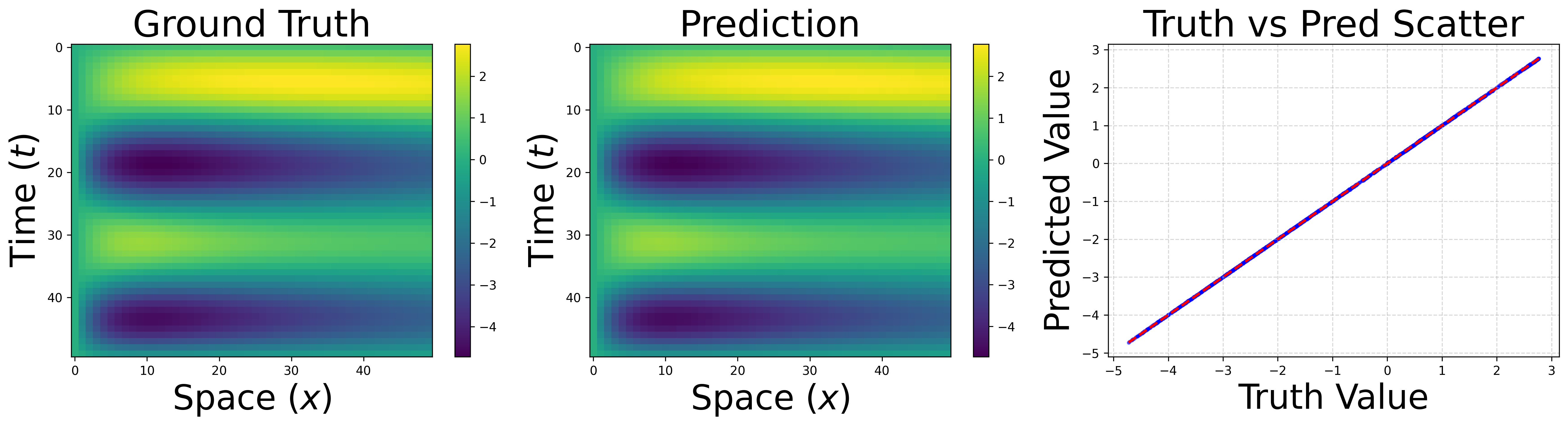}
        \caption{Qualitative Results and Scatter of the reaction-diffusion equation in FNO}
        \label{fig: FNO_ReactionDiffusion_qualitative}
    \end{subfigure}
    \vspace{0.5em}
    \begin{subfigure}[b]{0.8\textwidth}
        \centering
        \includegraphics[height=4cm]{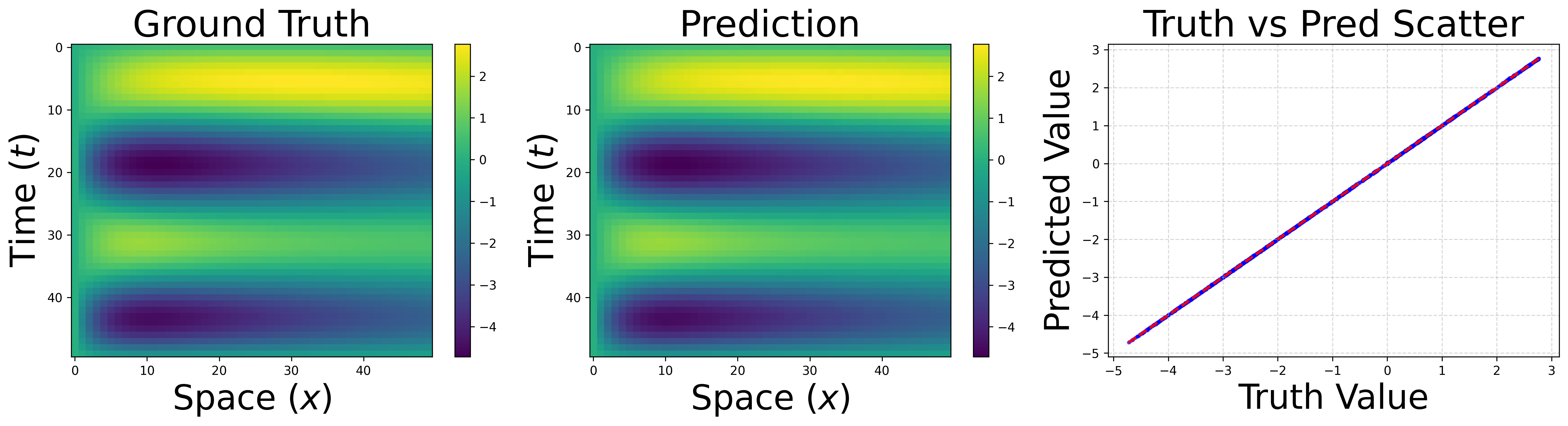}
        \caption{Qualitative Results and Scatter of the reaction-diffusion equation in LFNO}
        \label{fig: LFNO_ReactionDiffusion_qualitative}
    \end{subfigure}
    \vspace{0.5em}
    \begin{subfigure}{0.8\textwidth}
        \centering
        \includegraphics[height=4cm]{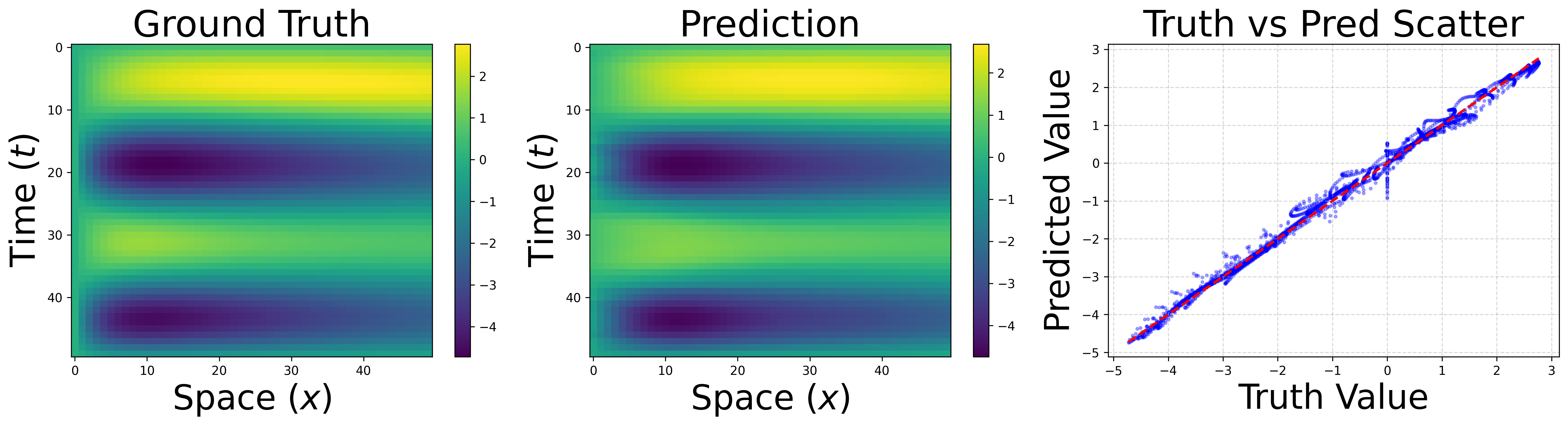}
        \caption{Qualitative Results and Scatter of the reaction-diffusion equation in LNO}
        \label{fig: LNO_ReactionDiffusion_qualitative}
    \end{subfigure}

    \caption{Reaction-diffusion equation: Comparison of qualitative results computed by FNO, LFNO, and LNO. It represents the ground truth shown in the left plot, the prediction of each model in the center plot, and the (truth, pred) scatters are shown in the right plot.}
    \label{fig: ReactionDiffusion_qualitative_comparison}
\end{figure}

\begin{figure}[ht]
    \centering
    \begin{subfigure}[b]{0.8\textwidth}
        \centering
        \includegraphics[height=4cm]{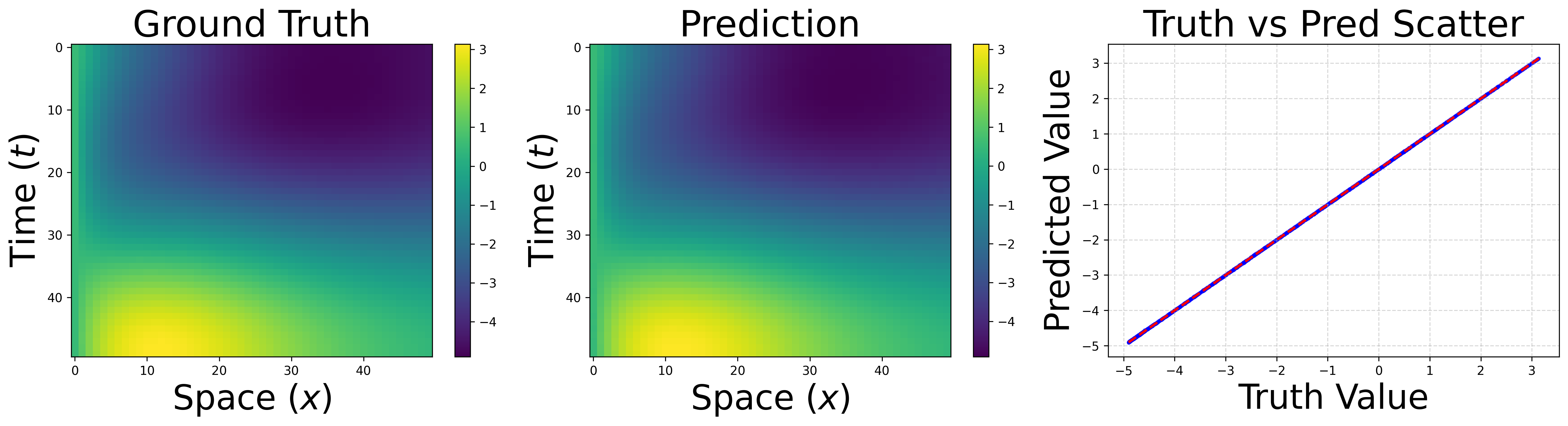}
        \caption{Qualitative Results and Scatter of the Brusselator equation in FNO}
        \label{fig: FNO_Brusselator_qualitative}
    \end{subfigure}
    \vspace{0.5em}
    \begin{subfigure}[b]{0.8\textwidth}
        \centering
        \includegraphics[height=4cm]{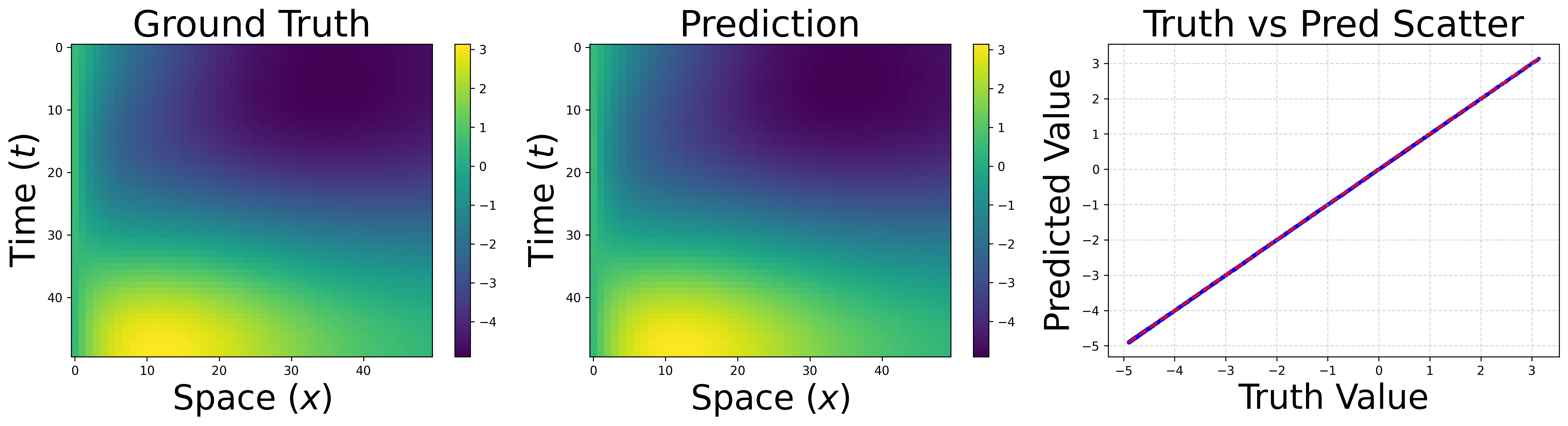}
        \caption{Qualitative Results and Scatter of the Brusselator equation in LFNO}
        \label{fig: LFNO_Brusselator_qualitative}
    \end{subfigure}
    \vspace{0.5em}
    \begin{subfigure}{0.8\textwidth}
        \centering
        \includegraphics[height=4cm]{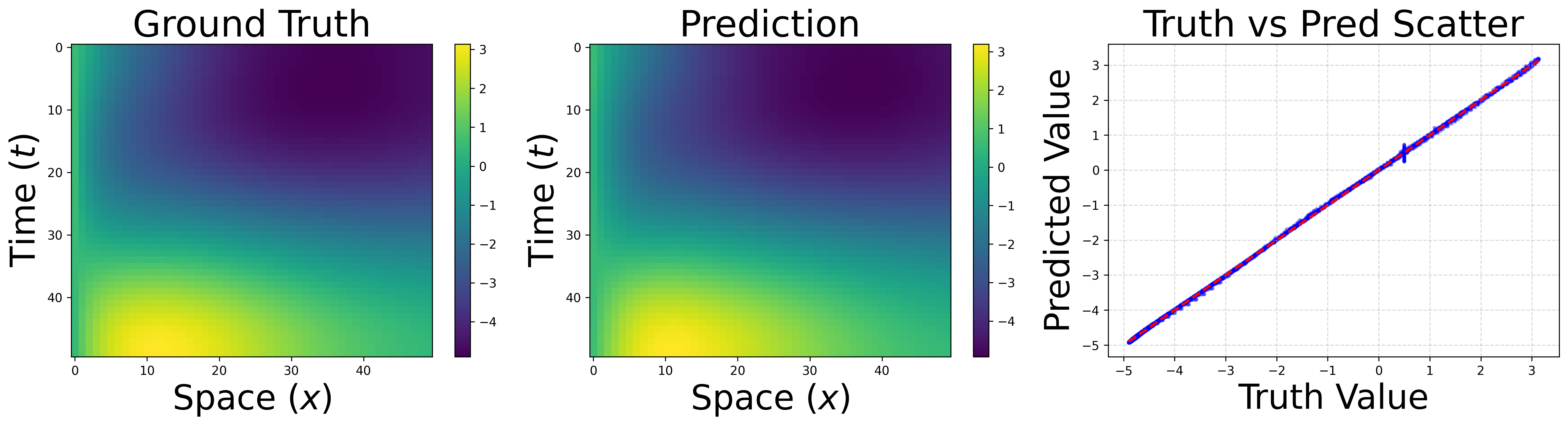}
        \caption{Qualitative Results and Scatter of the Brusselator equation in LNO}
        \label{fig: LNO_Brusselator_qualitative}
    \end{subfigure}

    \caption{Brusselator equation: Comparison of qualitative results computed by FNO, LFNO, and LNO. It represents the ground truth shown in the left plot, the prediction of each model in the center plot, and the (truth, pred) scatters are shown in the right plot.}
    \label{fig: Brusselator_qualitative_comparison}
\end{figure}

\begin{figure}[ht]
    \centering
    \begin{subfigure}[b]{0.8\textwidth}
        \centering
        \includegraphics[height=4cm]{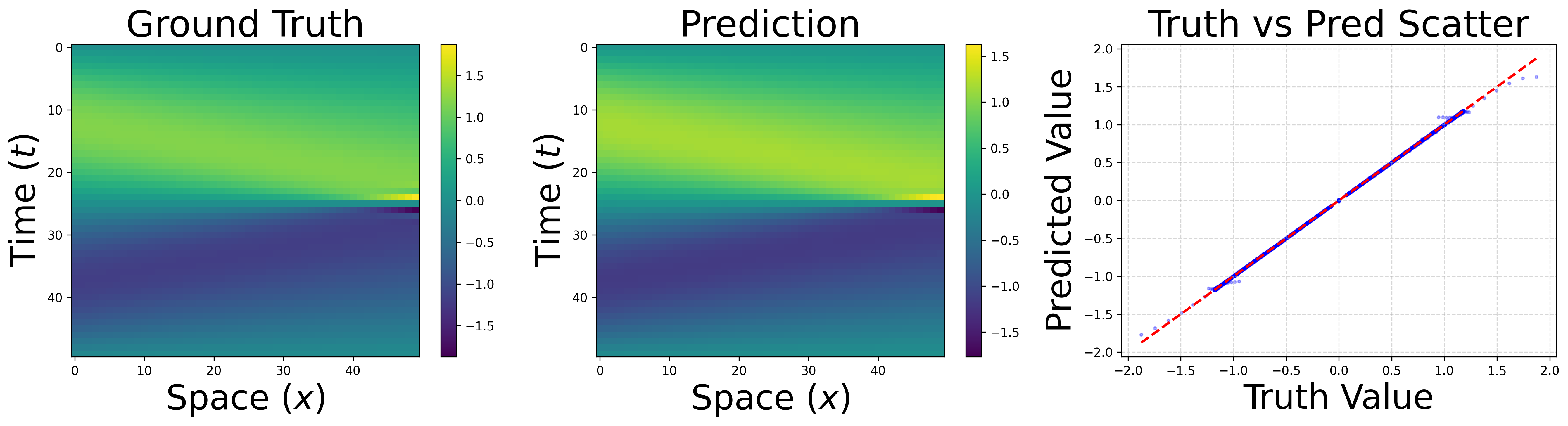}
        \caption{Qualitative Results and Scatter of the Burgers' equation in FNO}
        \label{fig: FNO_Burgers_qualitative}
    \end{subfigure}
    \vspace{0.5em}
    \begin{subfigure}[b]{0.8\textwidth}
        \centering
        \includegraphics[height=4cm]{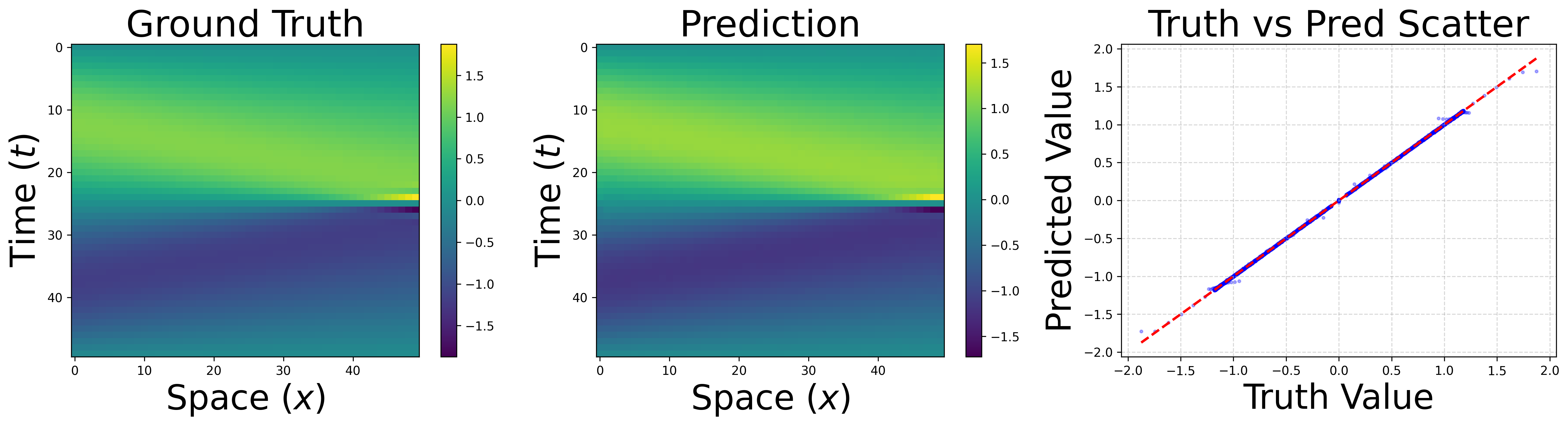}
        \caption{Qualitative Results and Scatter of the Burgers' equation in LFNO}
        \label{fig: LFNO_Burgers_qualitative}
    \end{subfigure}
    \vspace{0.5em}
    \begin{subfigure}{0.8\textwidth}
        \centering
        \includegraphics[height=4cm]{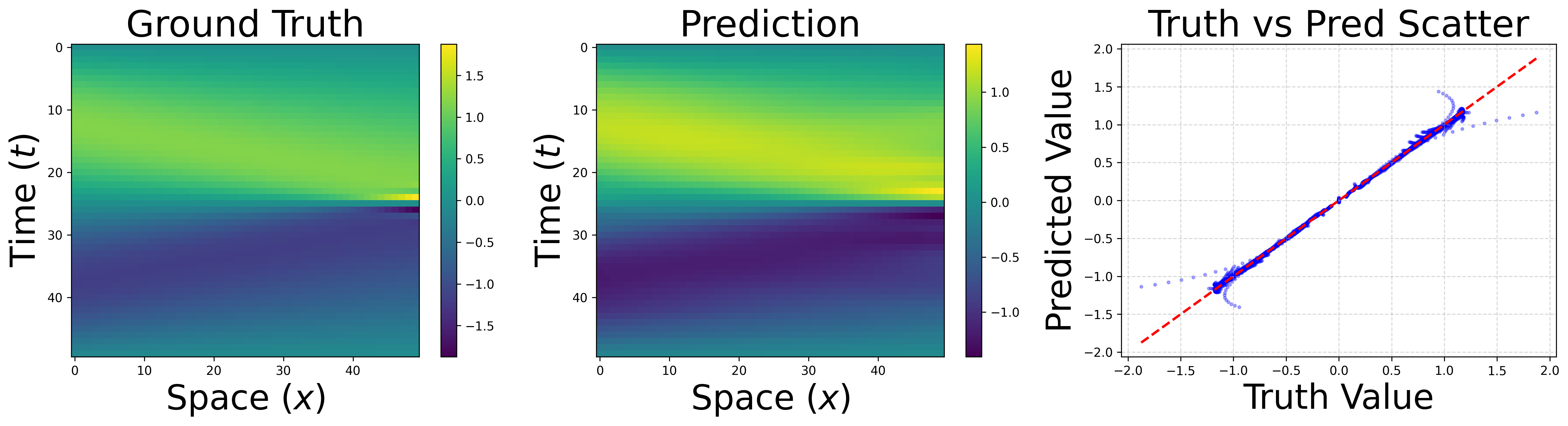}
        \caption{Qualitative Results and Scatter of the Burgers' equation in LNO}
        \label{fig: LNO_Burgers_qualitative}
    \end{subfigure}
    
    \caption{Burgers' equation: Comparison of qualitative results computed by FNO, LFNO, and LNO. It represents the ground truth shown in the left plot, the prediction of each model in the center plot, and the (truth, pred) scatters are shown in the right plot.}
    \label{fig: Burgers_qualitative_comparison}
\end{figure}

The qualitative results of PDEs are shown in \cref{fig: Beam_qualitative_comparison}, \cref{fig: Diffusion_qualitative_comparison}, \cref{fig: ReactionDiffusion_qualitative_comparison}, \cref{fig: Brusselator_qualitative_comparison}, and \cref{fig: Burgers_qualitative_comparison}. Our model's qualitative results are competitive with those of FNO.

\begin{figure}[ht]
    \centering
    \begin{subfigure}[b]{0.8\textwidth}
        \centering
        \includegraphics[height=4cm]{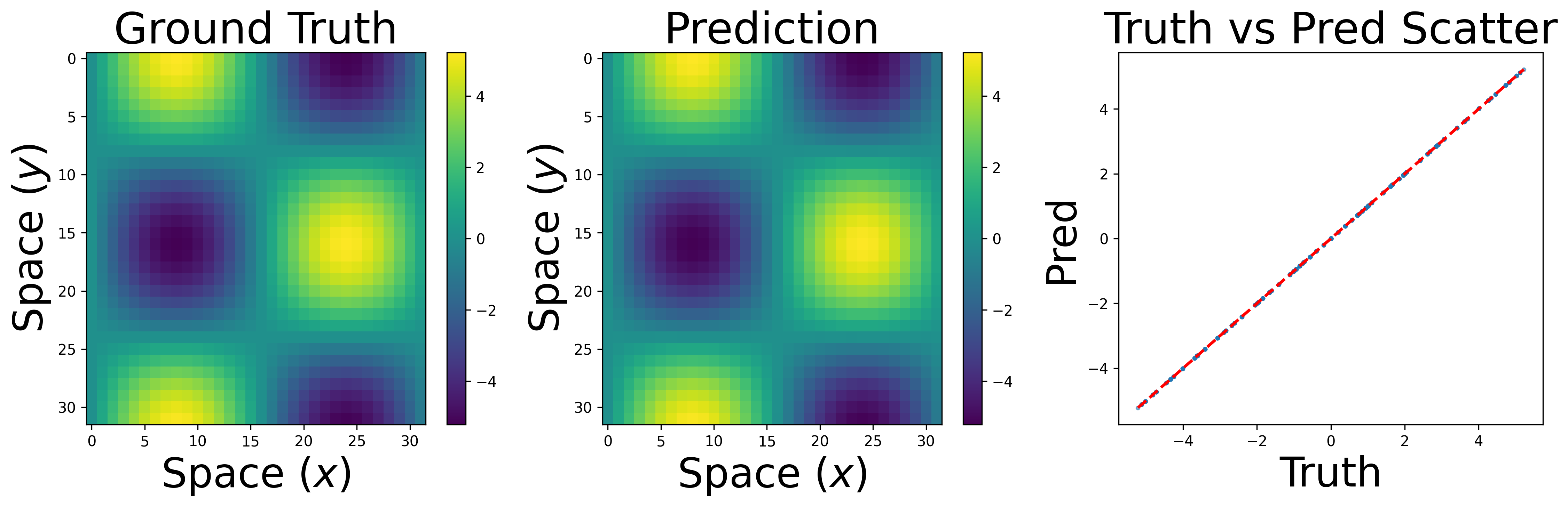}
        \caption{Qualitative Results and Scatter of the Navier-Stokes equation in FNO}
        \label{fig: FNO_NavierStokes_qualitative}
    \end{subfigure}
    \vspace{0.5em}
    \begin{subfigure}[b]{0.8\textwidth}
        \centering
        \includegraphics[height=4cm]{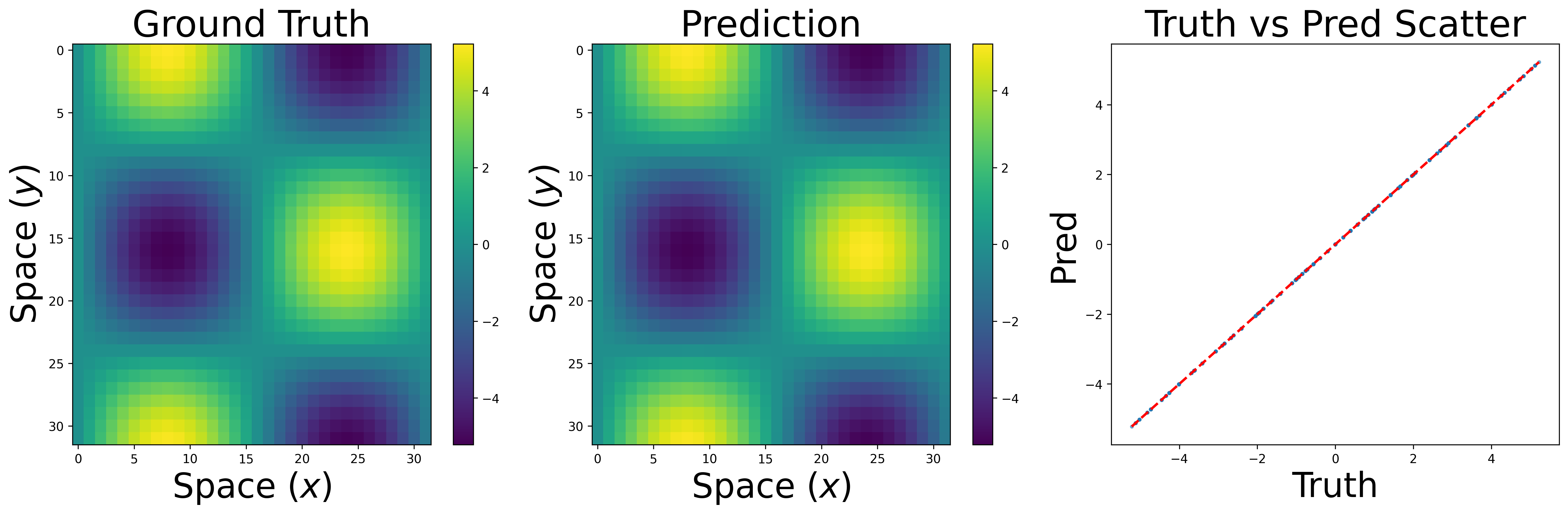}
        \caption{Qualitative Results and Scatter of the Navier-Stokes equation in LFNO}
        \label{fig: LFNO_NavierStokes_qualitative}
    \end{subfigure}
    \vspace{0.5em}
    \begin{subfigure}{0.8\textwidth}
        \centering
        \includegraphics[height=4cm]{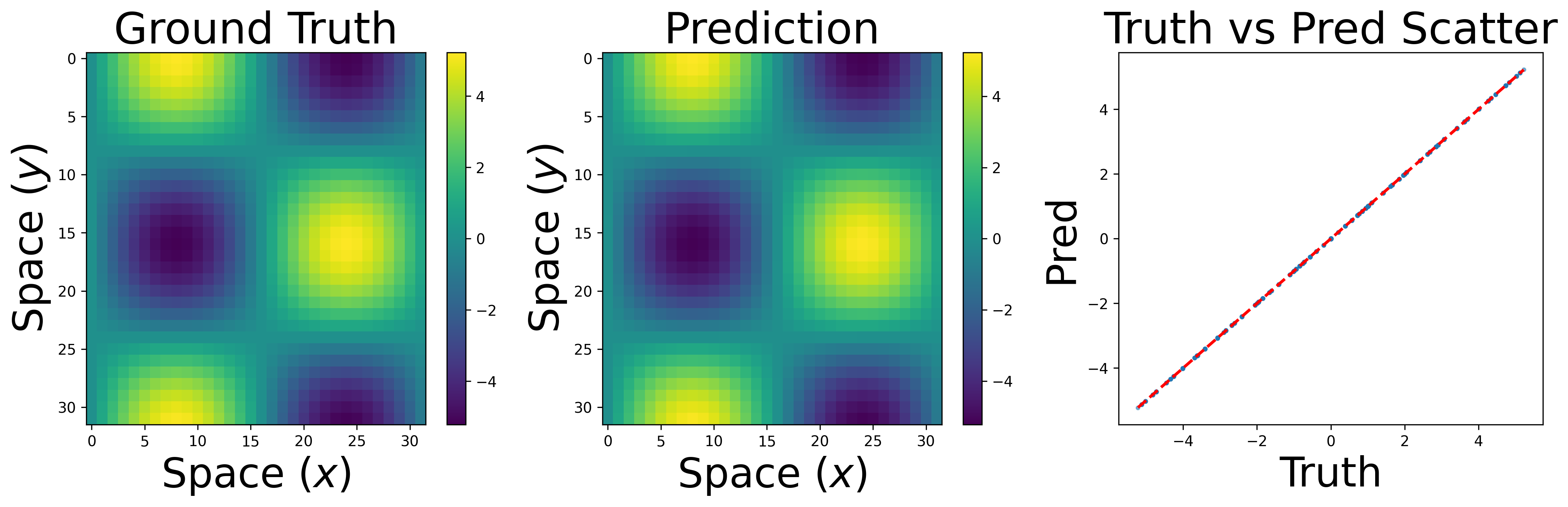}
        \caption{Qualitative Results and Scatter of the Navier-Stokes equation in LNO}
        \label{fig: LNO_NavierStokes_qualitative}
    \end{subfigure}
    
    \caption{Navier-Stokes equation: Comparison of qualitative results computed by FNO, LFNO, and LNO. It represents the ground truth shown in the left plot, the prediction of each model in the center plot, and the (truth, pred) scatters are shown in the right plot.}
    \label{fig: NavierStokes_qualitative_comparison}
\end{figure}


\end{document}